\documentclass[runningheads]{llncs}
\usepackage[T1]{fontenc}
\usepackage{graphicx}

\usepackage{amsmath}
\usepackage{amssymb}
\usepackage{amsfonts}

\newcommand{\N}{\mathcal{N}}
\newcommand{\mysubheading}[1]{\vspace{2mm}\noindent\textbf{#1}}
\newcommand{\E}[1]{\mathbb{E}\!\left[ #1 \right]}
\newcommand\intrange[3]{#1 \in \{#2, ..., #3\}}

\begin{document}
\title{Addressing degeneracies in latent \\ interpolation for diffusion models}

\author{Erik Landolsi\orcidID{0000-0002-6639-1257} \and
Fredrik Kahl\orcidID{0000-0001-9835-3020}}

\authorrunning{E. Landolsi and F. Kahl}

\institute{Chalmers University of Technology, Göteborg, Sweden\\
\email{\{erik.landolsi,fredrik.kahl\}@chalmers.se}}

\maketitle

\begin{abstract}
There is an increasing interest in using image-generating diffusion models for deep data augmentation and image morphing. In this context, it is useful to interpolate between latents produced by inverting a set of input images, in order to generate new images representing some mixture of the inputs. We observe that such interpolation can easily lead to degenerate results when the number of inputs is large. We analyze the cause of this effect theoretically and experimentally, and suggest a suitable remedy. The suggested approach is a relatively simple normalization scheme that is easy to use whenever interpolation between latents is needed. We measure image quality using FID and CLIP embedding distance and show experimentally that baseline interpolation methods lead to a drop in quality metrics long before the degeneration issue is clearly visible. In contrast, our method significantly reduces the degeneration effect and leads to improved quality metrics also in non-degenerate situations.

\end{abstract}

% \date{December 2024}

\section{Introduction}
Over the last few years, diffusion models have shown remarkable performance in image generation \cite{balaji2022ediffi,ramesh2022hierarchical,saharia2022photorealistic}. This has sparked an interest in using diffusion models for data augmentation or fully synthetic data generation in label-constrained tasks~\cite{azizi2023synthetic,sariyildiz2023fake,trabucco2024}. Such tasks often involve adapting a diffusion model to create images with a similar appearance as a set of input images (the scarce available training data). Approaches in this direction sometimes involve some sort of interpolation in latent space, e.g. computing the centroid of a set of input examples~\cite{samuel2024norm} or blending inverted inputs with noise~\cite{landolsi2024tiny}.

Another line of research around diffusion models involves image morphing, with the goal of producing visually pleasing smooth transitions between two images~\cite{yang2023impus,zhang2024diffmorpher}. Such methods could be extended to smooth interpolation between a larger set of images. The input images would then define a manifold in the latent space, from which new images could be generated from any choice of interpolation coefficients mixing the original images.

\begin{figure*}[t!]
\centering
  \includegraphics[width=0.15\linewidth]{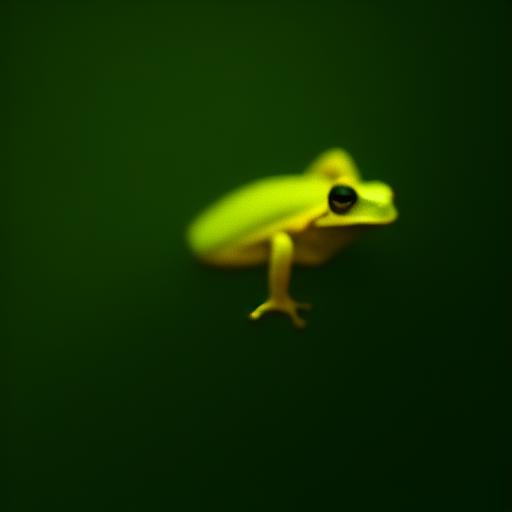}
  \includegraphics[width=0.15\linewidth]{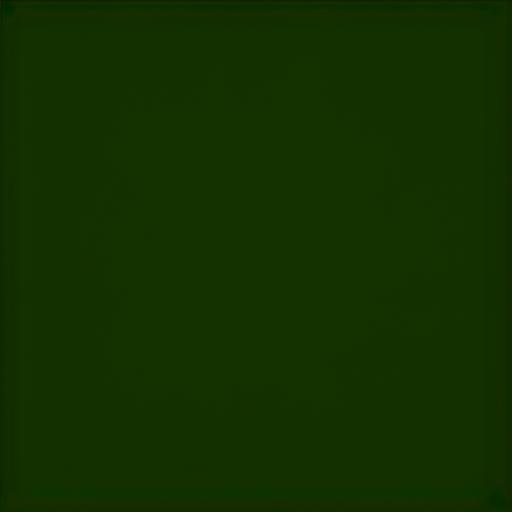}
  \includegraphics[width=0.15\linewidth]{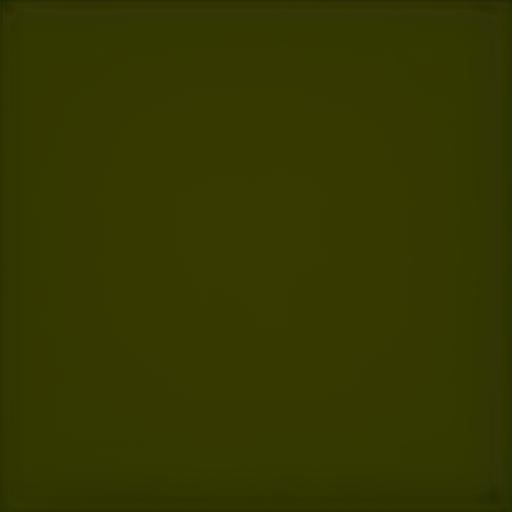}
  \includegraphics[width=0.15\linewidth]{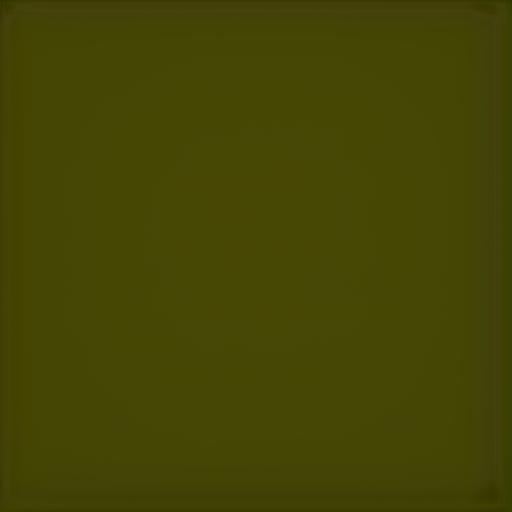} 
  \includegraphics[width=0.15\linewidth]{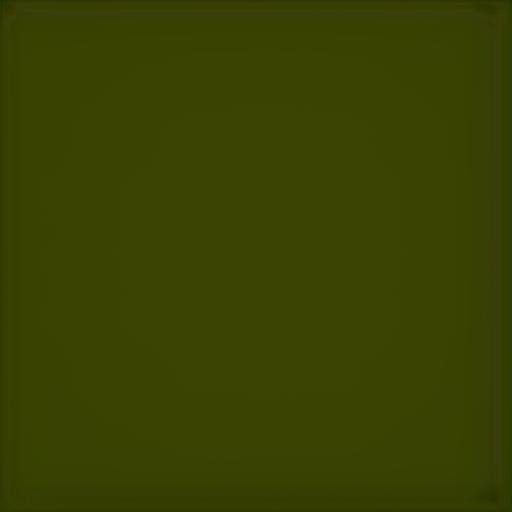}
  \includegraphics[width=0.15\linewidth]{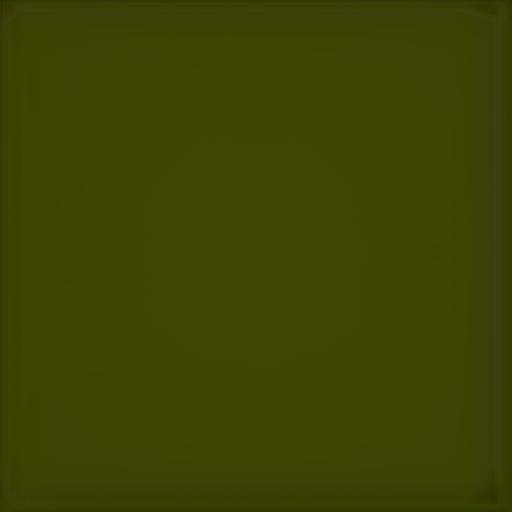} \\
  \vspace{0.2cm}
  \includegraphics[width=0.15\linewidth]{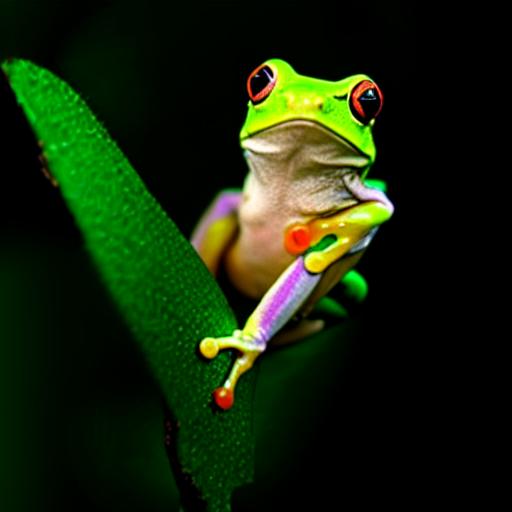}
  \includegraphics[width=0.15\linewidth]{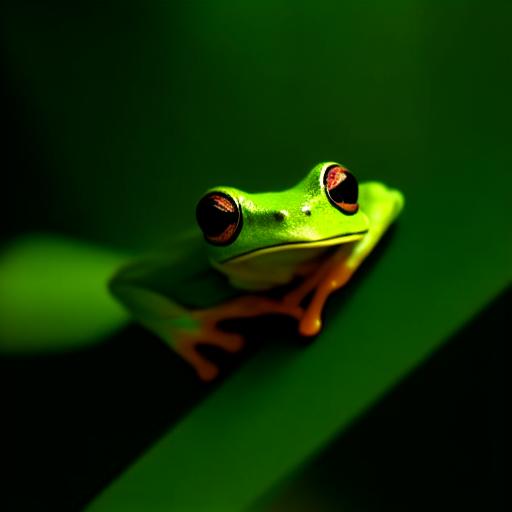}
  \includegraphics[width=0.15\linewidth]{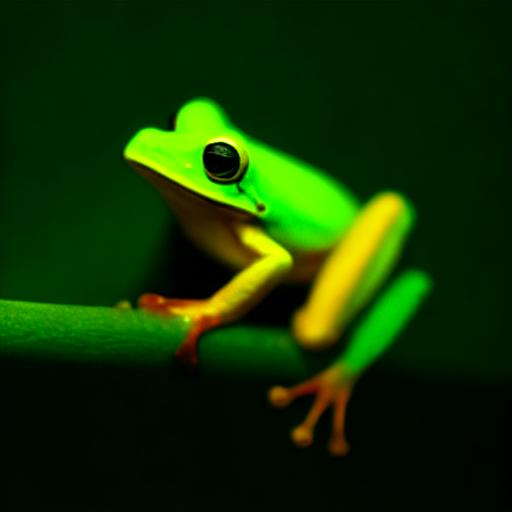}
  \includegraphics[width=0.15\linewidth]{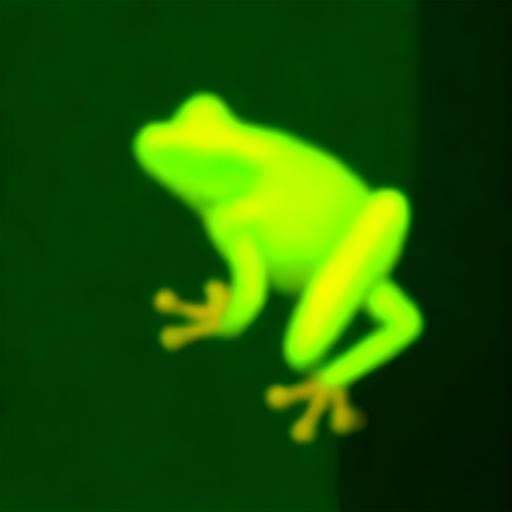}
  \includegraphics[width=0.15\linewidth]{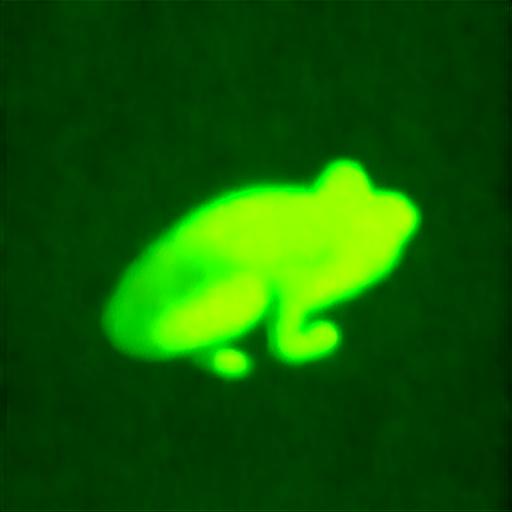}
  \includegraphics[width=0.15\linewidth]{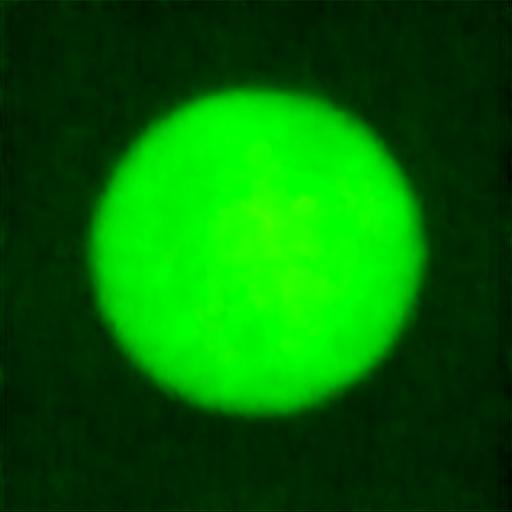} \\
  \vspace{0.2cm}
  \includegraphics[width=0.15\linewidth]{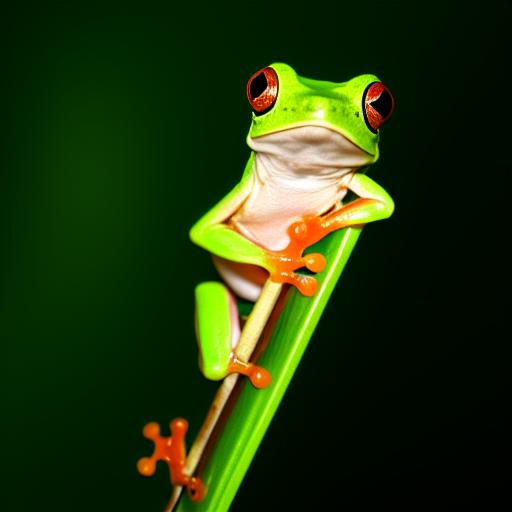}
  \includegraphics[width=0.15\linewidth]{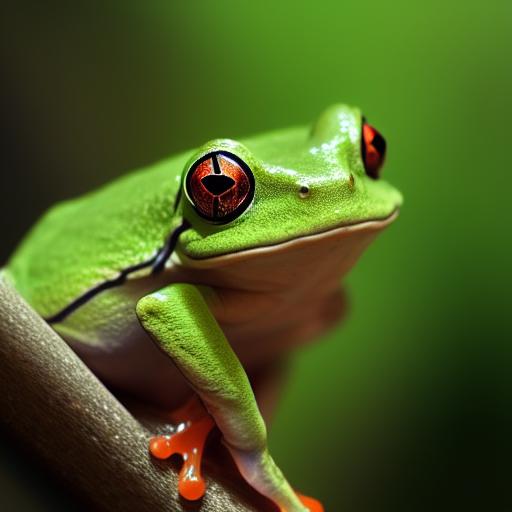}
  \includegraphics[width=0.15\linewidth]{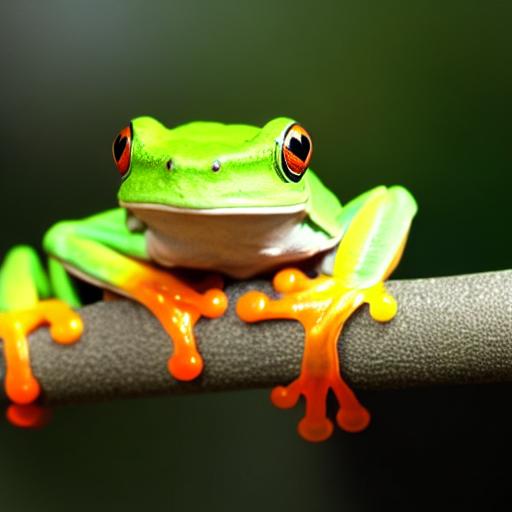}
  \includegraphics[width=0.15\linewidth]{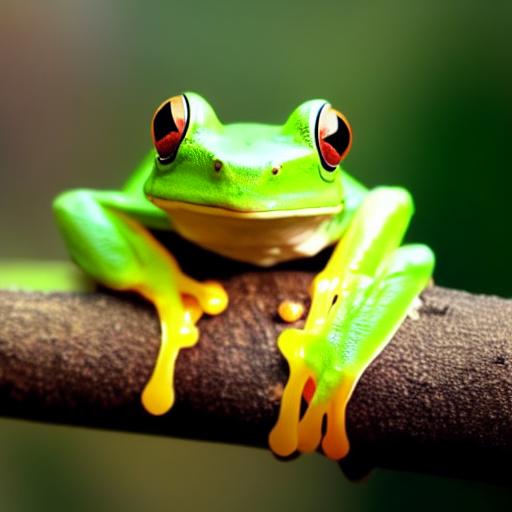} 
  \includegraphics[width=0.15\linewidth]{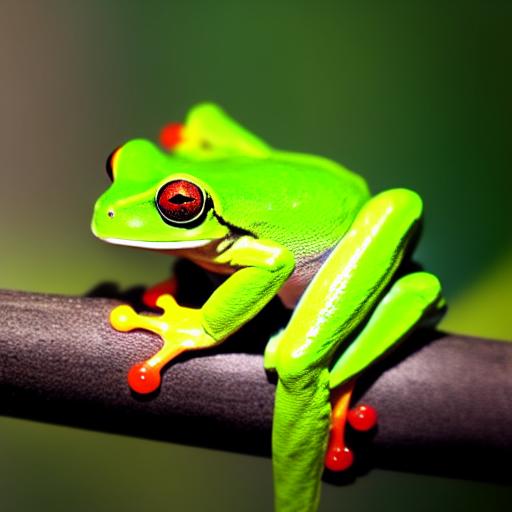}
  \includegraphics[width=0.15\linewidth]{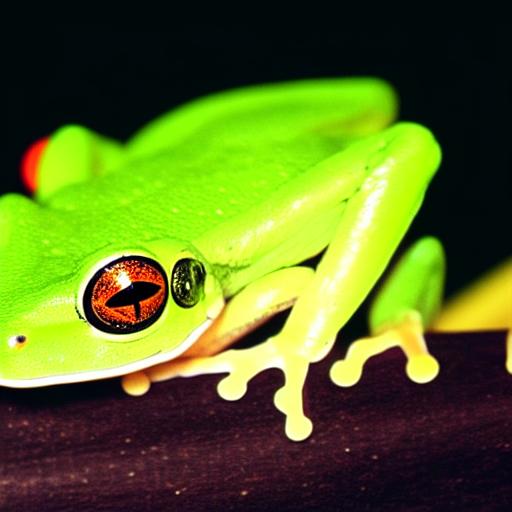} \\
 
  \caption{Images generated from \emph{centroids} of $N$ latents obtained from $N$ input images of ImageNet class ``tree frog'' for $N = 2, 8, 32, 48, 64, 96$. Top row: Linear interpolation. Middle row: Linear interpolation with fixed normalization. Bottom row: channel-wise mean adjustment (our suggested method).}

  \label{fig:intro_example}
\end{figure*}

A common operation in the mentioned methods is interpolation between diffusion model latents. That is, given a set of input images $\{\mathbf{x}_n\}$ with latents $\{\mathbf{z}_n\}$ and a corresponding set of mixing weights $\{w_n : w_n > 0, \sum{w_n} = 1\}$, finding a mixed latent representing a meaningful mixture of the input images. We note that naive methods for computing such mixtures can easily lead to degenerate results. As a motivating example, in Figure~\ref{fig:intro_example}, a set of latents was produced from $N$ images using DDIM inversion~\cite{song2020denoising}. The latent centroids were then computed and used to generate new images. The top row shows results using regular linear interpolation, which quickly breaks down due to the norm of the interpolated latent being too low~\cite{samuel2024norm}. A simple fix would be to normalize the latent centroid to a suitable norm level. As shown in the middle row, this works well when $N$ is small, but the results are still degenerate when $N$ is large. 

In this paper, we deep-dive into the issue of degenerate output from diffusion model latent interpolation. After presenting related work in Section~\ref{sec:related_work}, we review baseline interpolation methods in Section~\ref{sec:latent_space_interp}. In Section~\ref{sec:degenerate}, we diagnose the degeneracy illustrated in Figure~\ref{fig:intro_example}, showing when and why it appears. In Section~\ref{sec:remedies_normalized_ccs}, we examine alternative normalization schemes as a potential remedy. Finally, in Section~\ref{sec:results}, we show experimentally that such schemes can measurably improve the quality of generated images even for smaller $N$, where the issue is not as obvious as in the high $N$ examples in Figure~\ref{fig:intro_example}.

\section{Related work}
\label{sec:related_work}
Our work builds on text-to-image diffusion models~\cite{ho2020denoising,song2020denoising}, specifically latent diffusion models in the Stable Diffusion family~\cite{esser2024scaling,rombach2022high}. We are not aware of any prior work analyzing degenerate output from latent interpolation in such models. 

Various forms of latent space interpolation are most commonly found in morphing methods, and already the original DDPM paper~\cite{ho2020denoising} showed examples of this. A more recent example \cite{yang2023impus} first run textual inversion~\cite{gal2022image} on the inputs. The resulting text embeddings were interpolated linearly and the noisy latents using SLERP. The authors reached improved results by also performing a low-rank adaptation of the diffusion model. Finally, they also proposed a perceptually uniform sampling to ensure a smooth transition between the inputs. Another work \cite{zhang2024diffmorpher} used similar interpolation of text conditions and embeddings, but introduced an interpolation also on the attention maps. Both these works only considered two inputs, and the authors did not identify or address the degeneracy that is our focus. Note that the latter method \cite{zhang2024diffmorpher} includes an adaptive instance normalization (AdaIN) that is similar to our suggested channel-wise mean adjustment. However, they introduce it in an ad-hoc fashion, without any deeper motivation, whereas we provide a detailed analysis and show that such normalization is key to avoid degenerate results for large $N$.

Our work is also related to diffusion model inversion. The degeneration issue appears using inverted latents, and improved inversion techniques may be a competing way of fixing it. In this paper, we rely mostly on the DDIM inversion introduced in the original DDIM paper~\cite{song2020denoising}. Since then, null-text inversion~\cite{mokady2023null} has been suggested as a way of inverting diffusion models including classifier-free guidance by optimizing the unconditional embeddings. A later work~\cite{miyake2023negative} provided a related but faster method by avoiding the costly optimization. Another line of work studied fixed-point methods for inverting diffusion models~\cite{meiri2023fixed,pan2023effective}. The recent ReNoise method~\cite{garibi2024renoise} included a fixed-point inversion and additional loss terms aimed at better shaping the noise statistics. While improved inversion methods could potentially serve as competing remedies for degenerate centroids, none of the mentioned papers identified or studied this specific issue.

One related work~\cite{samuel2024norm} studied interpolation paths in the latent space. The authors first noted that it is critical to maintain a well-scaled latent norm. They then constructed interpolation paths between latents by minimizing a likelihood-related measure, preferring paths passing through areas where the latents have a typical norm value. Their interpolation method induces a metric in the latent space that they used also for computing centroids of up to 5 examples. These centroids were then used in a deep data augmentation pipeline for label-constrained recognition problems. However, they failed to notice that such centroids can become degenerate as the number of inputs increases, and they do not provide any analysis or remedy relating to this effect.

\section{Latent space interpolation}
\label{sec:latent_space_interp}

\subsection{Latent diffusion models}
\emph{Denoising diffusion probabilistic models} \cite{ho2020denoising} model the distribution of a random variable $\mathbf{x}_0$ by transforming it into a tractable distribution (noise) over timesteps $\intrange{t}{0}{T}$. A \emph{latent diffusion model} performs the diffusion process on a latent variable $\mathbf{z}_t$ of lower dimensionality than $\mathbf{x}_t$. 
We use the \emph{Stable Diffusion} (SD)~\cite{rombach2022high} family of models due to their public availability and wide use in prior work. To make sure that our findings are not just based on implementation curiosities in a single version, we use two distinct SD versions (1.5 and 3.5). In SD 1.5, the noise estimation is implemented using a U-net \cite{ronneberger2015u}, and the conditioning input is a CLIP embedding \cite{radford2021learning} of a text prompt. The forward noise model can be written as $\mathbf{z}_t = \sqrt{\alpha_t} \mathbf{z}_0 + \sqrt{1-\alpha_t} \mathbf{\epsilon}$ and the latent variable is a 4-channel feature map with a resolution 8 times lower than the images.

In SD 3.5~\cite{esser2024scaling}, the U-net is replaced with a transformer model, and the latent feature dimensionality is increased to 16. Furthermore, the noise model is changed into a rectified flow model~\cite{liu2022flow}, where the data is transformed to noise using a direct linear path according to $\mathbf{z}_t = (1-t) \mathbf{z}_0 + t \mathbf{\epsilon}$.

\subsection{Interpolation notation}
\label{sec:interp_and_cc}
Let $\mathbf{Z} = \{\mathbf{z}_1, \ldots, \mathbf{z}_N\}$ be an ordered set of input latents at time $t=T$ (dropping the $t$ subscript for brevity) and let $\mathbf{w} = \{w_1, \ldots, w_N\}$ be a corresponding ordered set of weights with $w_n\!>\!0$ and $\sum_n w_n = 1$. Furthermore, let $\mathbf{z}' = f(\mathbf{Z}, \mathbf{w})$ denote an interpolation operation mixing the $\mathbf{z}_n$ using weights $w_n$. A desired property of any $f$ is that if any $w_n = 1$, then $f(\mathbf{Z}, \mathbf{w}) = \mathbf{z}_n$, such that inputs are reproduced exactly. Finally, for compactness, let $f(\mathbf{Z}) = f(\mathbf{Z}, \{\frac{1}{N}, \ldots, \frac{1}{N}\})$ denote a centroid computation. 

\subsection{Baseline interpolation options}
\label{sec:baseline_interp_options}

\begin{figure*}[t!]
\centering
  \includegraphics[width=1.0\linewidth]{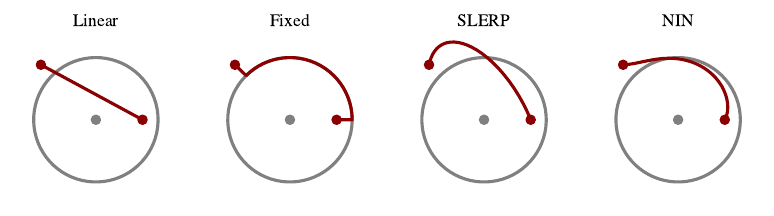}
  \vspace{-1.0cm}
  \caption{Interpolation paths produced by the methods from Section~\ref{sec:baseline_interp_options} for a 2D toy example. The gray circle represents latents with norm $\sqrt{L}$, which is where randomly sampled latents are typically located.}
  \label{fig:interpolation_paths}
\end{figure*}

\mysubheading{Linear interpolation}. The most direct baseline option is basic linear interpolation or \emph{convex combination} according to
\begin{equation}
  \label{eq:convexconv_intro}
  \mathbf{z}' = f_\textrm{LIN}(\mathbf{Z}, \mathbf{w}) \triangleq \sum_n w_n \mathbf{z}_n \, .
\end{equation}
As noted in prior work~\cite{samuel2024norm} and illustrated in Figures~\ref{fig:intro_example}-\ref{fig:interpolation_paths}, this often produces output with a significantly lower norm than the inputs, leading to washed-out images with a severe lack of detail.

\mysubheading{Fixed normalization}. The norm of randomly sampled $\mathbf{z} \sim \mathcal{N}(0, I)$ is sharply distributed around $\sqrt{L}$, where $L$ is the dimensionality of $\mathbf{z}$~\cite{samuel2024norm}. A simple attempt to fix the low norms produced by linear interpolation would be to normalize the interpolated latent to the typical norm $\sqrt{L}$ using
\begin{equation}
  \label{eq:unitnorm_def}
  \mathbf{z}' = f_\text{FIX}(\mathbf{Z}, \mathbf{w}) = 
    \frac{\sqrt{L}}
         { \big\| \sum_n w_n \mathbf{z}_n \big\| }
    \sum_n w_n \mathbf{z}_n \, .
\end{equation}
However, as illustrated in Figure~\ref{fig:interpolation_paths}, this breaks the desired input reproduction property mentioned in Section~\ref{sec:interp_and_cc}, since $f_{\textrm{FIX}}(\mathbf{Z}, \{1, 0, 0, \ldots\}) = \sqrt{L} \|\mathbf{z}_1\|^{-1} \mathbf{z}_1$ which is not equal to $\mathbf{z}_1$ in the general case. Therefore, $f_{\textrm{FIX}}$ may not be suitable as a general interpolation method. 

\mysubheading{SLERP}. Spherical linear interpolation (SLERP)~\cite{shoemake1985animating} is a way of interpolating between two points on a unit sphere according to 
\begin{equation}
  \label{eq:slerp_def}
  \textrm{slerp}([\mathbf{p}_1, \mathbf{p}_2], t) = 
    \frac{\sin{\big((1-t) \theta\big)}}{\sin{\theta}} \mathbf{p}_1 + 
    \frac{\sin{(t \theta)}}{\sin{\theta}} \mathbf{p}_2  
\end{equation}
where $\theta = \cos^{-1}(\mathbf{p}_1^\textrm{T} \mathbf{p}_2)$ is the angle between $\mathbf{p}_1$ and $\mathbf{p}_2$. If the inputs are not on the unit sphere, this is often handled by a normalization when computing $\theta$, letting $\theta = \cos^{-1}(\mathbf{p}_1^\textrm{T} \mathbf{p}_2 / \|\mathbf{p}_1\| \|\mathbf{p}_2\|)$ \cite{wang2023infodiffusion,yang2023impus}. However, this departs from the original SLERP formulation and can lead to unexpected results. In the example in Figure~\ref{fig:interpolation_paths}, the path initially moves away from the circle representing the typical norm value near the original inputs. This leads to latent norms outside the input norm range, which is undesirable since generated image quality tends to deteriorate when the norm departs from the nominal value~\cite{samuel2024norm}. Furthermore, in order to generalize SLERP to multiple inputs, iterative methods are required~\cite{buss2001spherical}. For these reasons, we will not consider SLERP further in this paper.

\mysubheading{Normalization to interpolated norms}. In~\cite{landolsi2024tiny}, it was suggested to instead set the norm of an interpolated latent to the linearly interpolated norms of the inputs, i.e. letting
\begin{equation}
  \label{eq:nin_def}
  \mathbf{z}' = f_{\textrm{NIN}}(\mathbf{Z}, \mathbf{w}) = \frac{\sum_n{w_n}\|\mathbf{z}_n\|}{\big\|\sum_n w_n \mathbf{z}_n\big\|} \sum_n{w_n \mathbf{z}_n}   
\end{equation}
This operation fulfills the input reproduction property mentioned in Section~\ref{sec:interp_and_cc} and does not suffer from the norm overshoots that can happen using SLERP (see Figure~\ref{fig:interpolation_paths}).

\section{Degenerate interpolation output}
\label{sec:degenerate}
Using $f_\textrm{FIX}$ from Eq.~\ref{eq:unitnorm_def} or $f_\textrm{NIN}$ from Eq.~\ref{eq:nin_def}, one might expect interpolated latents to be free from issues caused by lacking normalization. However, as illustrated in Figure~\ref{fig:intro_example}, as the number of inputs included in an interpolation operation grows, the output can be degenerate even though the latent norm should now be well-behaved. In this section, we analyze the cause of this phenomenon.

\subsection{Initial investigation}
\label{sec:initial_investigation}
Let us first consider an ideal case of $N$ i.i.d.\ latents $\mathbf{z}_{n} \sim \N(0, I)$ and their normalized average $\mathbf{z}' = f_\textrm{FIX}(\mathbf{Z}, \mathbf{w})$. The weighted sum of i.i.d.\ normally distributed variables is also normally distributed with adjusted variance. After normalization, $\mathbf{z}'$ is thus uniformly distributed on a hypersphere with radius $\sqrt{L}$, regardless of $N$. In this ideal case, we do not see any degeneration effect as $N$ grows. Hence, we can conclude that our inverted latents $\mathbf{z}_n$ of real images are not i.i.d.\ $\N(0, I)$.

\begin{figure*}[t!]
\centering
  \includegraphics[width=0.15\linewidth]{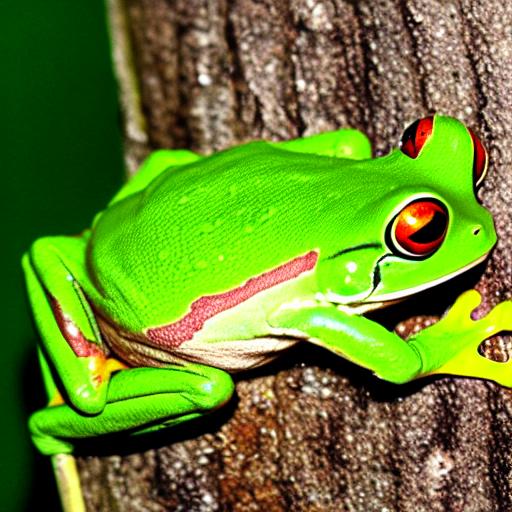}
  \includegraphics[width=0.15\linewidth]{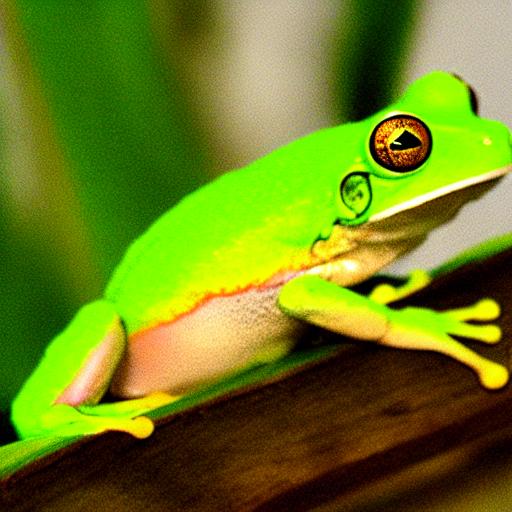}
  \includegraphics[width=0.15\linewidth]{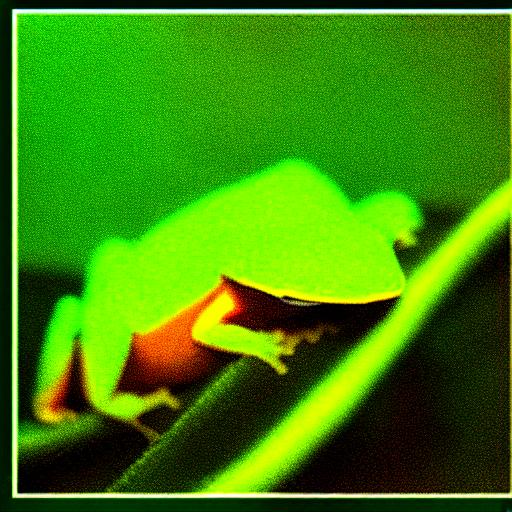}
  \includegraphics[width=0.15\linewidth]{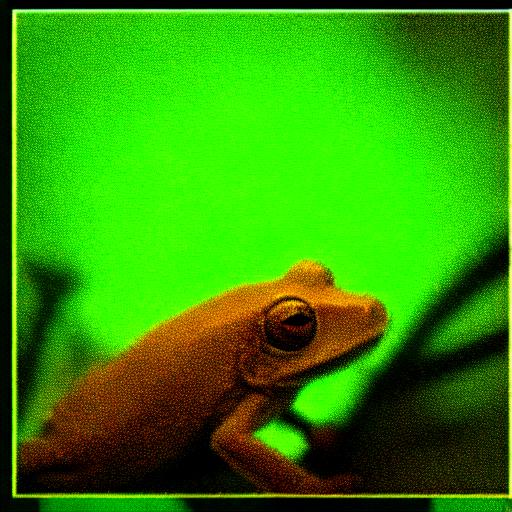} 
  \includegraphics[width=0.15\linewidth]{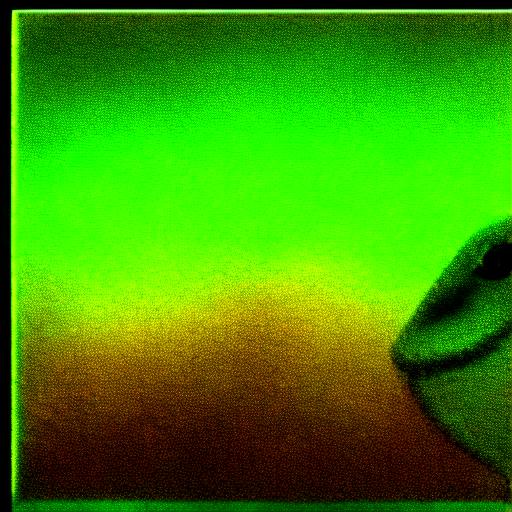}
  \includegraphics[width=0.15\linewidth]{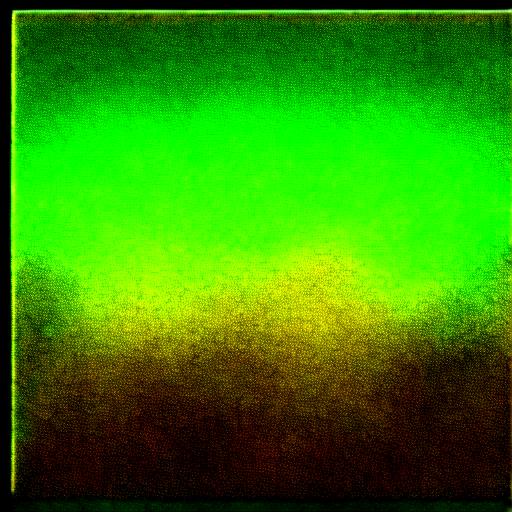} \\
 
  \caption{Images generated from centroids created using the recipe in Section~\ref{sec:initial_investigation}. ImageNet class ``tree frog'', $N = 2, 8, 32, 48, 64, 96$, fixed normalization.}

  \label{fig:degenerate_although_iid}
\end{figure*}

In our examples, the $\mathbf{z}_n$ are drawn from the same ImageNet class, and that class might not align perfectly with a text concept in the diffusion model. Therefore, inverted latents might be clustered in the latent space, breaking the normality assumption. To check if the issue is caused by such a misalignment, we could instead construct examples $\mathbf{z}_n$ using the following procedure:
\begin{enumerate}
  \item Draw a random initial $\mathbf{\epsilon} \sim \N(0, I)$, feed it through the diffusion model to create an image $\mathbf{x}$ using the prompt \emph{a photo of a [class name]}.
  \item Perturb $\mathbf{x}$ by adding a small amount of Gaussian noise, in order to create a new image that is not a pixel-perfect actual diffusion model output.
  \item Run a diffusion inversion procedure on the perturbed $\mathbf{x}$ to create an inverted noisy latent $\mathbf{z}$.
\end{enumerate}
Running this $N$ times produces $N$ i.i.d.\ latents $\mathbf{z}_{n}$ where there should be no clustering in the latent space caused by misalignment between input data classes and the diffusion model. Figure~\ref{fig:degenerate_although_iid} shows an example of images generated from centroids computed from such latents. We see that the degeneration issue remains, showing that the issue is not caused by the mentioned potential misalignment.

Hence, we can conclude that inverted latents do not in general follow the statistics of random samples drawn from a normal distribution. In fact, we have noted that the channel-wise mean values of inverted latents often have a small bias that gets further amplified by the latent normalization.

This effect is studied in more detail in the following subsections.

\subsection{Bias amplification}
To see how a small latent bias is amplified, consider a simplified case with latents $\mathbf{z}_n = \mathbf{d} + \mathbf{e}_n$ consisting of i.i.d.\ noise terms $\mathbf{e}_n \sim \mathcal{N}(0, I)$ perturbed by a small common deterministic term $\mathbf{d}$. We now consider what happens to the normalized mean value $\mathbf{z}' = f_\textrm{FIX}(\mathbf{Z})$ compared to the ideal unperturbed $\mathbf{e}' = f_\textrm{FIX}\mathbf(\mathbf{E})$, where $\mathbf{E} = \{\mathbf{e}_1, \ldots, \mathbf{e}_N\}$. Let $\alpha$ be the amplification factor in $f_\textrm{FIX}(\mathbf{Z})$, i.e.
\begin{equation}
  \alpha 
    = \frac{\sqrt{L}}{ \left\| \sum_n{ \frac{1}{N} \mathbf{z}_n} \right\|}
    = \frac{\sqrt{L}}{ \left\| \mathbf{d} + \frac{1}{N}\textstyle\sum_n{ \mathbf{e}_n} \right\| }
    \, .
\end{equation}
For $\mathbf{z}'$, we can write
\begin{equation}
\mathbf{z}' 
  = \alpha \sum_n{\frac{1}{N} \mathbf{z_n}} 
  = \alpha \mathbf{d} + \alpha \frac{1}{N} \sum_n{\mathbf{e}_n}.
\end{equation}
To understand the amplification factor $\alpha$, note that since $\mathbf{e}_n$ are i.i.d.\ Gaussians with $\E{\|\mathbf{e}_n\|^2} = L$, then $\E{ \| \frac{1}{N}\textstyle\sum_n{ \mathbf{e}_n} \|^2} = \frac{L}{N}$. Since $L$ is large, $\|\mathbf{e}_n\|$ is sharply distributed, and $\| \frac{1}{N}\textstyle\sum_n{ \mathbf{e}_n} \|$ even more so. We can therefore approximate this norm with a fixed value, letting
\begin{equation}
\left\| \frac{1}{N}\textstyle\sum_n{ \mathbf{e}_n} \right\| \approx \sqrt{\frac{L}{N}}.
\end{equation}
If $\mathbf{d}$ is small, its contribution to $\alpha$ is negligible. Consider for example the case where $\mathbf{d} = b$ is a fixed small constant. Then $\|\mathbf{d}\| = b \sqrt{L}$, and $\mathbf{d}$ is negligible in the $\alpha$ denominator if $b^2 \ll \frac{1}{N}$, which is a reasonable assumption in the situations studied in this paper. We can then approximate $\alpha$ as 
\begin{equation}
  \alpha 
    \approx \frac{\sqrt{L}}{ \left\| \frac{1}{N} \sum_n{\mathbf{e}_n} \right\|}
    \approx \sqrt{N} \, ,
\end{equation}
leading to the final approximation
\begin{equation}
  \mathbf{z}' \approx \sqrt{N} \mathbf{d} + \mathbf{e}' \, .
\end{equation}
In other words, any small common bias in the latents will be amplified by approximately $\sqrt{N}$ in the normalization. Although this analysis used $f_\textrm{FIX}$, similar behavior can be expected also using $f_\textrm{NIN}$, since all $\|\mathbf{z}_n\| \approx \sqrt{L}$.

\begin{figure*}[t!]
  \centering
  \includegraphics[width=0.47\linewidth]{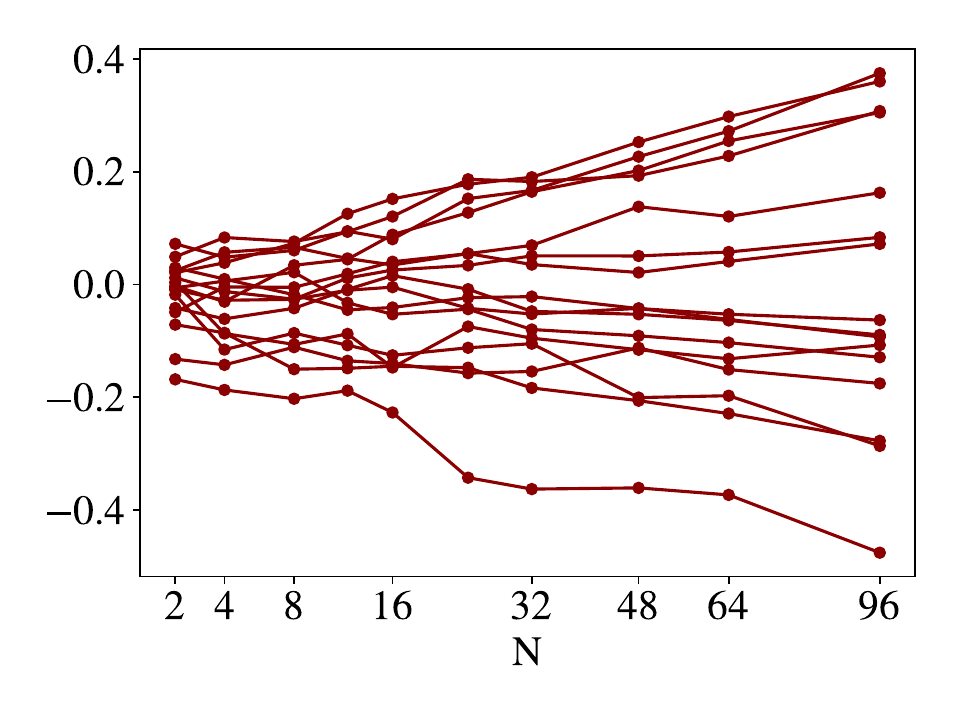}
  \includegraphics[width=0.47\linewidth]{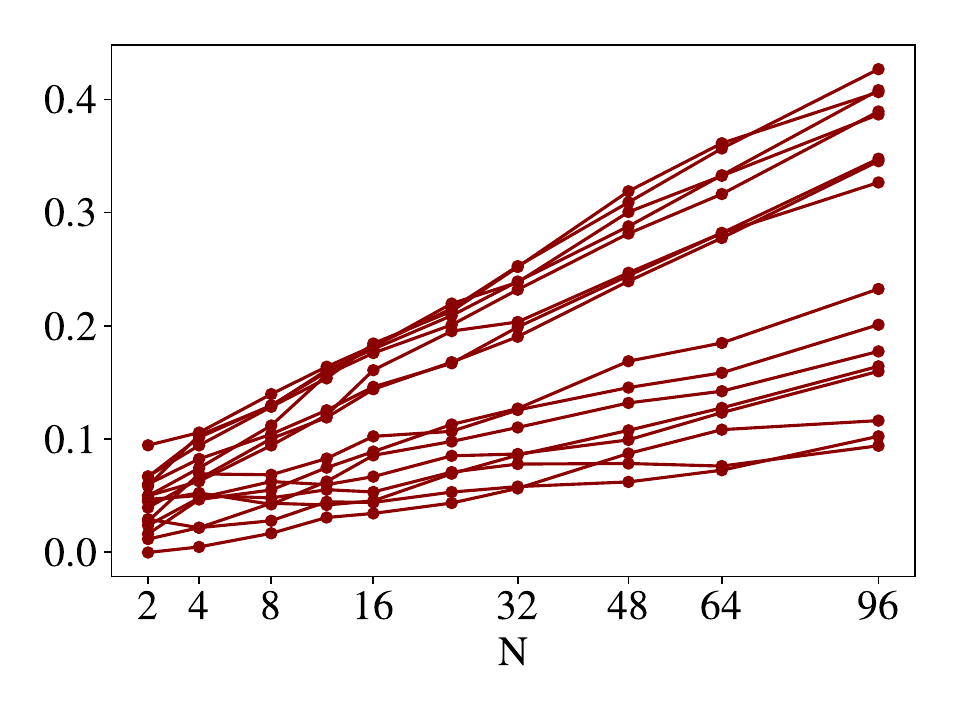}
  \vspace{-0.5cm}
 
  \caption{Channel-wise mean of the first two channels of $f_\textrm{FIX}$ centroids computed from $N$ inverted input images, plotted for 8 selected ImageNet classes. Left: SD 1.5, right: SD 3.5. The x-axes have a square-root scale to illustrate the linear dependence on $\sqrt{N}$.}

  \label{fig:bias_amplification}
\end{figure*}

In Figure~\ref{fig:bias_amplification}, we show an example where the measured channel-wise mean for a few examples are plotted against $N$. There are a few outliers, but the general trend is that the bias grows roughly linearly in $\sqrt{N}$, as predicted by our theory.

\subsection{The origin of latent bias}
\label{sec:origin_of_latent_bias}
\begin{figure*}[t!]
\centering
  \includegraphics[width=0.15\linewidth]{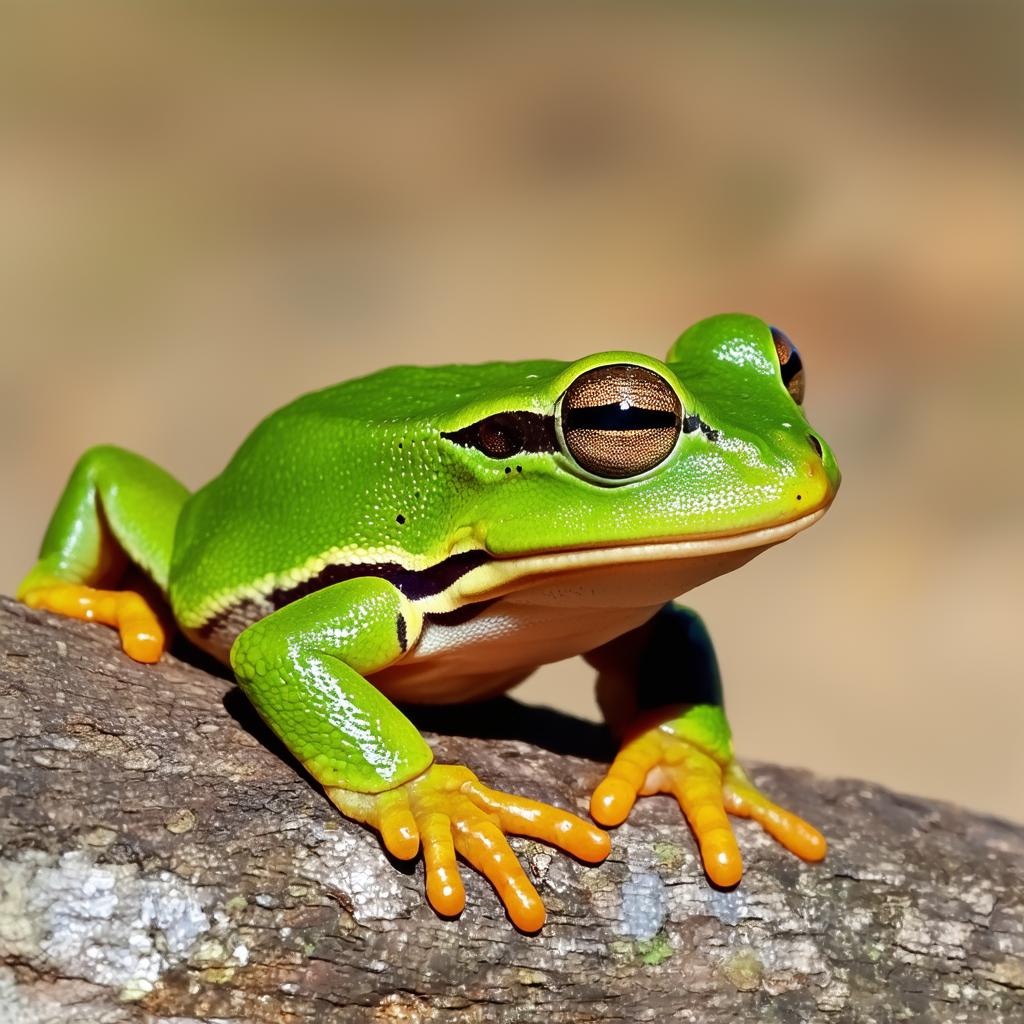}
  \includegraphics[width=0.15\linewidth]{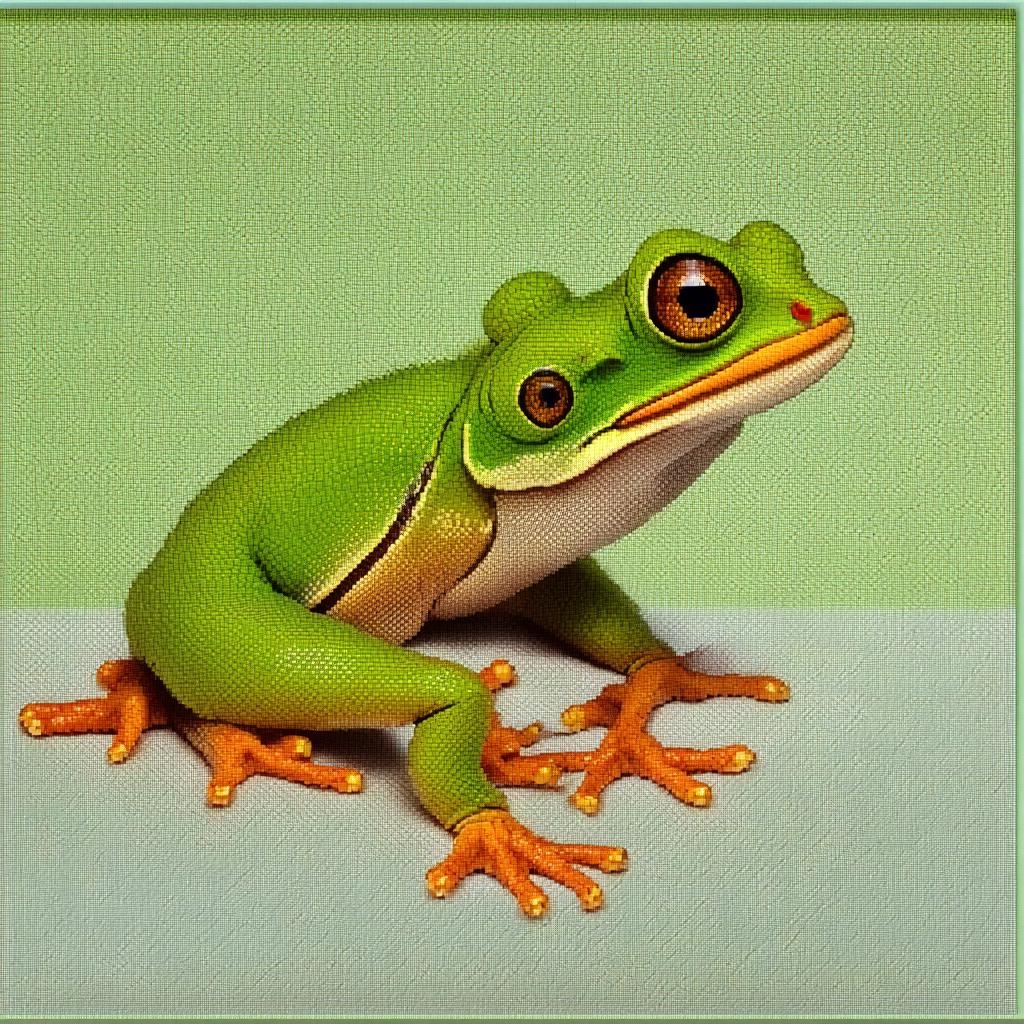}
  \includegraphics[width=0.15\linewidth]{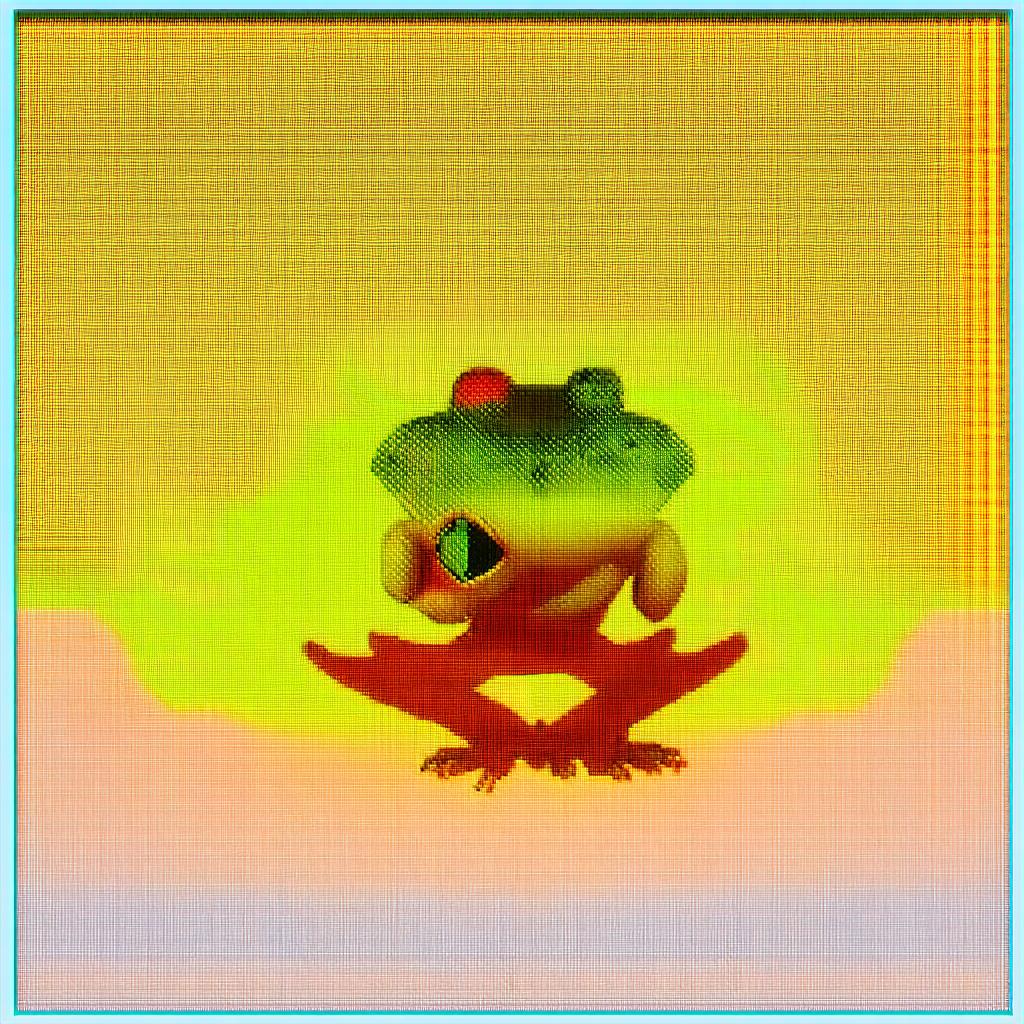}
  \includegraphics[width=0.15\linewidth]{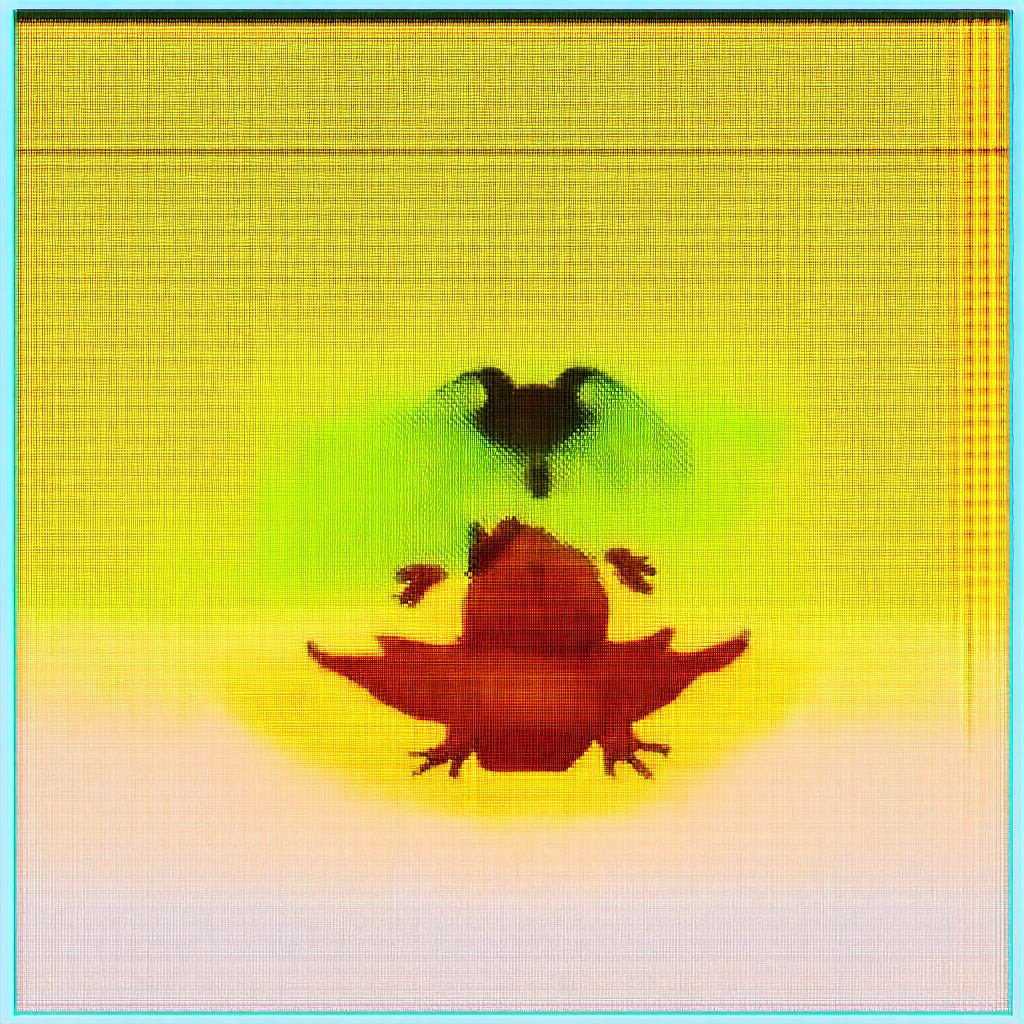} 
  \includegraphics[width=0.15\linewidth]{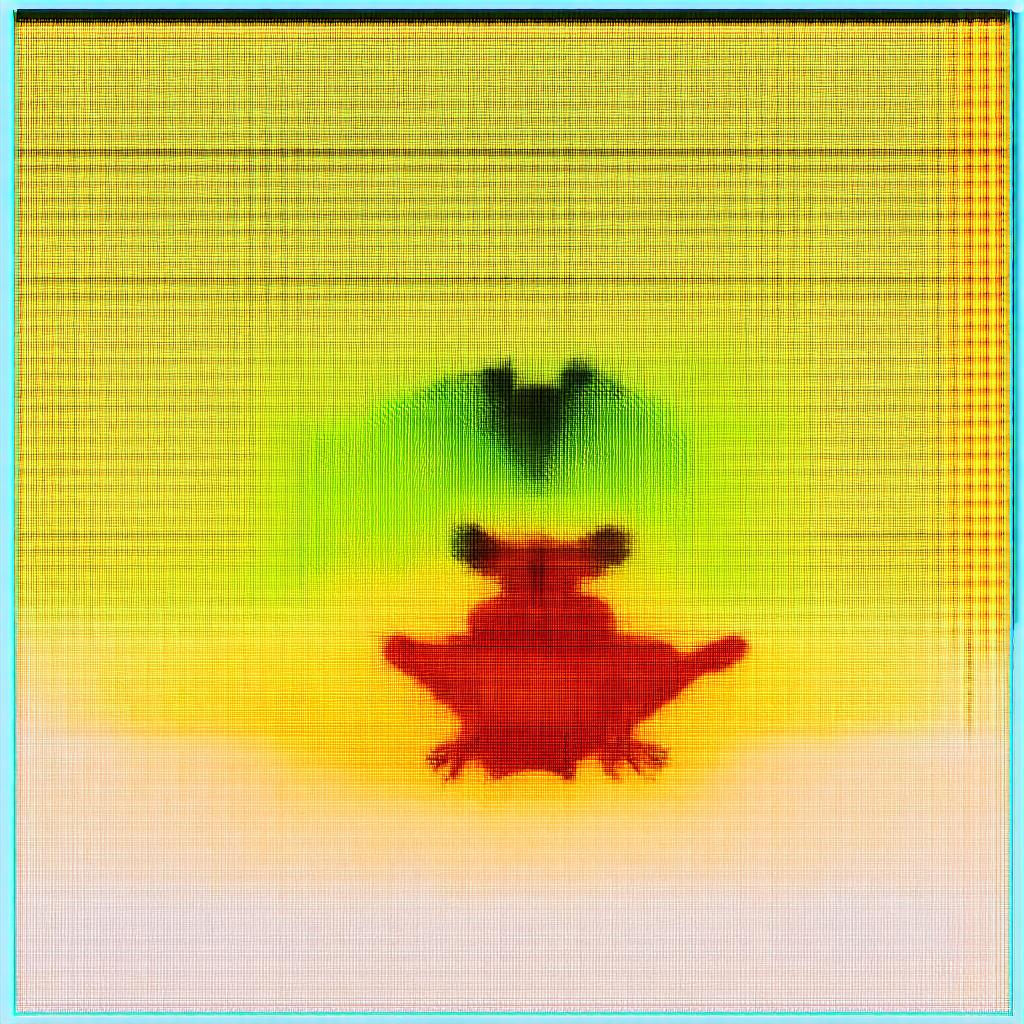}
  \includegraphics[width=0.15\linewidth]{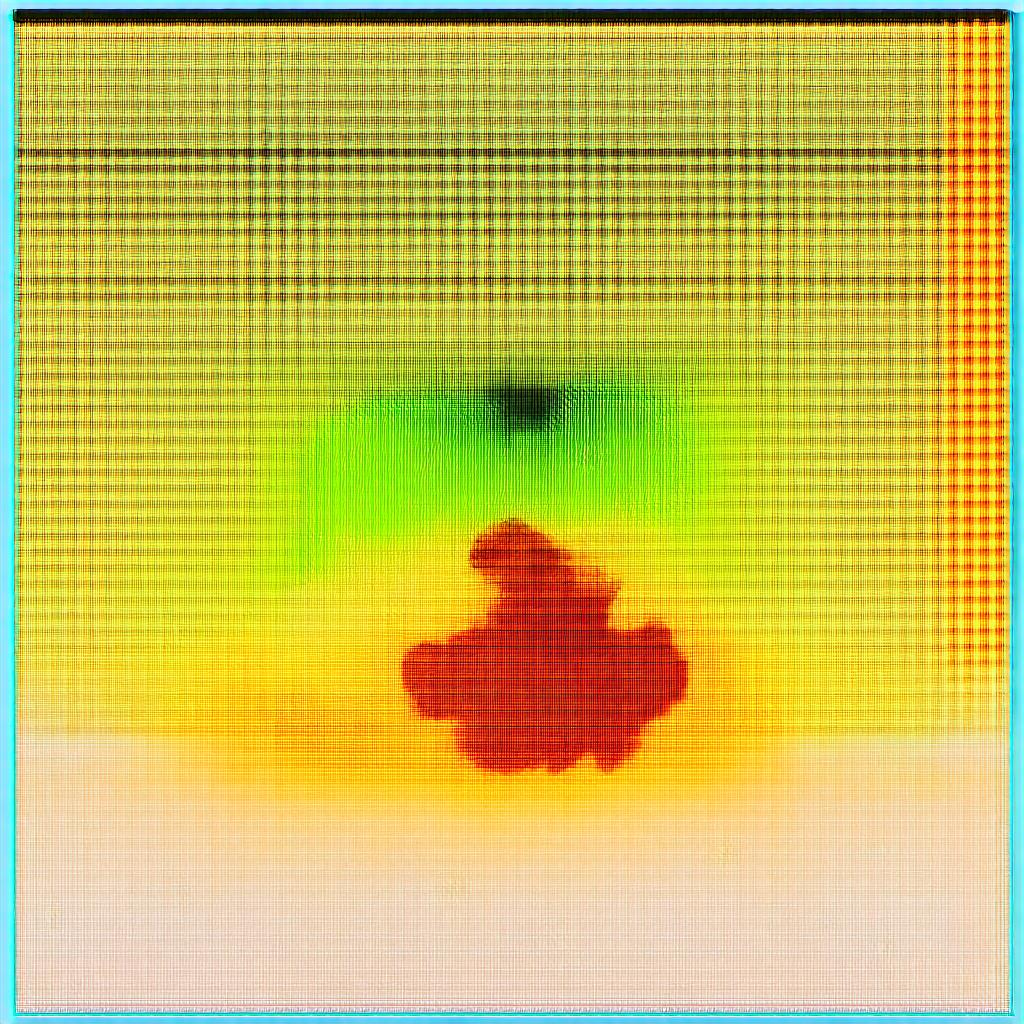} \\
 
  \caption{Images generated from centroids created using the recipe from Section~\ref{sec:initial_investigation}, to ensure i.i.d.\ examples, using SD 3.5 to avoid the non-zero terminal SNR issue. $N = 2, 8, 32, 48, 64, 96$ (increasing to the right), fixed normalization.}

  \label{fig:degenerate_although_sd35}
\end{figure*}

It has previously been shown~\cite{lin2023common} that common implementations of diffusion model schedulers are flawed, in the sense that the latent $\mathbf{z}_t$ does not reach zero terminal SNR at $t=T$. For e.g. the DDIM scheduler in the Hugging Face Diffusers implementation of SD 1.5, at $t=T$ the latent $\mathbf{z}_T = \sqrt{\alpha_T} \mathbf{z}_0 + \sqrt{1-\alpha_T} \mathbf{\epsilon}$ with $\alpha_T = 0.0047$. Even though $\alpha_T$ may appear negligible, $\sqrt{\alpha_T} \approx 0.07$, which is a significant number. This could certainly be a source of a deterministic trace signal in $\mathbf{z}_T$. In SD 3.5, this non-zero terminal SNR issue has been fixed. However, as illustrated in Figure~\ref{fig:degenerate_although_sd35}, the degeneration issue remains. We hypothesize that trace amounts of latent bias could have several origins, including using imperfect schedulers and training imperfections in the noise estimation model.

\subsection{The effect of latent bias}
\begin{figure*}[t!]
  \centering
  \includegraphics[width=0.19\linewidth]{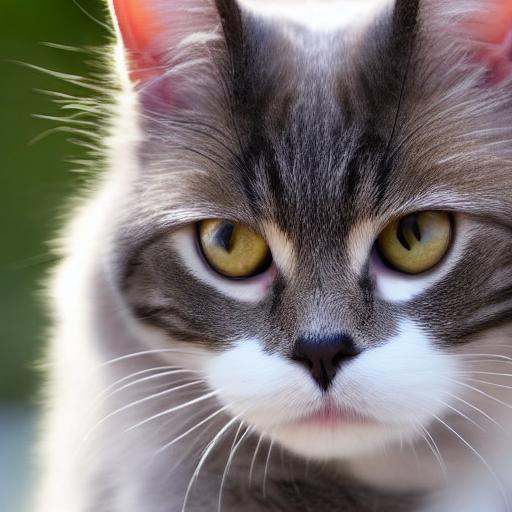}
  \includegraphics[width=0.19\linewidth]{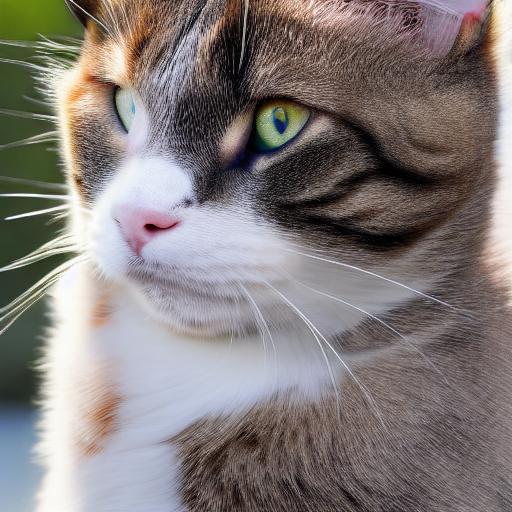}
  \includegraphics[width=0.19\linewidth]{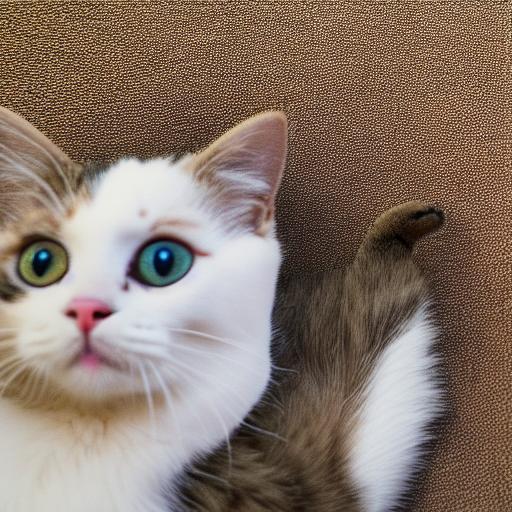}
  \includegraphics[width=0.19\linewidth]{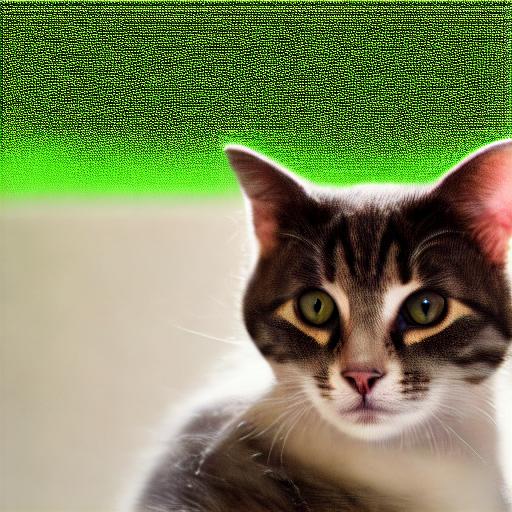}
  \includegraphics[width=0.19\linewidth]{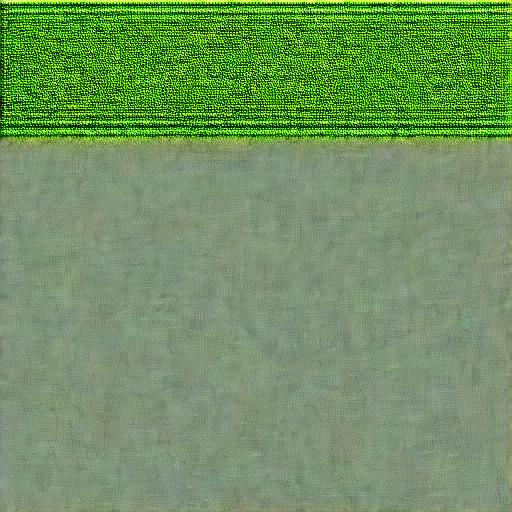} \\
  \vspace{0.2cm}
  \includegraphics[width=0.19\linewidth]{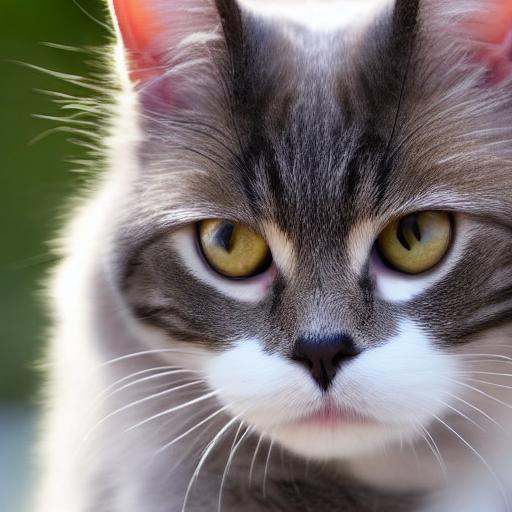}
  \includegraphics[width=0.19\linewidth]{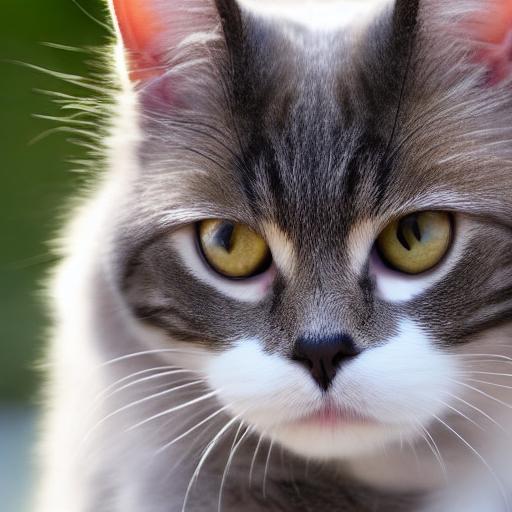}
  \includegraphics[width=0.19\linewidth]{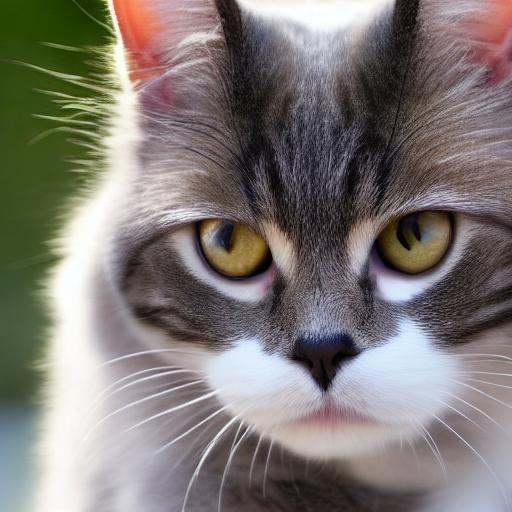}
  \includegraphics[width=0.19\linewidth]{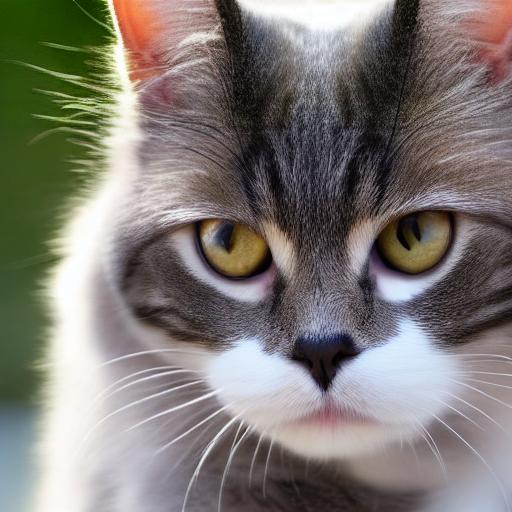}
  \includegraphics[width=0.19\linewidth]{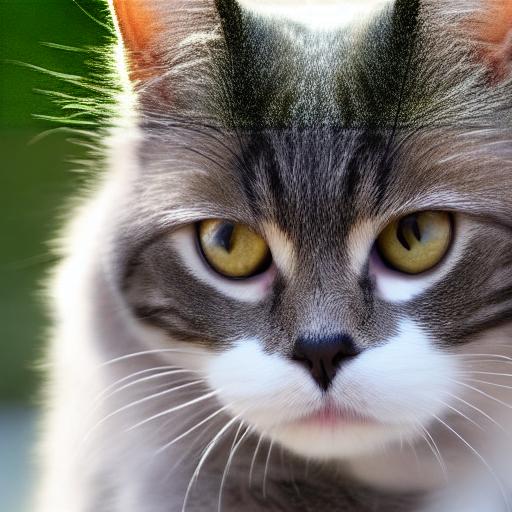} \\
 
  \caption{Illustrating the effect of latent mean value offsets. Top row: offsets at timestep $t=T$. Bottom row: offsets at timestep $t=0$. Columns: offsets $b = 0.0, 0.1, 0.2, 0.4, 0.8$.}
  \label{fig:bias_effect}
\end{figure*}

To study the effect of unexpected latent statistics, we show a qualitative example with a set of images produced from the same latent noise, but where a part of the latent was perturbed by a constant offset. Specifically, in the top quarter of the image, channel $0$ of the latent was offset by $-b$ and channel $1$ by $b$, such that the global mean remained unchanged. The modified latent was then fed to the diffusion model, and the generated images are shown in the top row of Figure~\ref{fig:bias_effect}.

As a comparison, the bottom row of Figure~\ref{fig:bias_effect} shows the effect of instead adding the same offsets to the latent at timestep 0, i.e. after the diffusion process but before decoding it into an image. In this case, only the image part corresponding to the modified latent part changes, and the effect is hardly noticeable until the offset is quite large. This indicates that the degeneration issue is not caused by latent codes getting shifted outside of their valid domain, but rather by the noise prediction model being sensitive to unexpected input statistics.

\subsection{Alternative inversion procedures}
Since the degenerate outputs are likely caused by inverted latents $\mathbf{z}_n$ having unexpected statistics, one class of remedies could be to attempt improved inversion methods. Diffusion model inversion is a research direction on its own, and we will only touch briefly upon this direction here.

First, note that DDIM inversion is an approximation of the forward process that works best without classifier-free guidance and with a large number of diffusion timesteps. We confirmed that the degeneracy issue persists also under such conditions. We also considered \emph{null-text inversion} \cite{mokady2023null} as an example of an improved inversion procedure supporting classifier-free guidance. Also in this case, the degeneracy issue persists. More details and qualitative examples can be found in Appendix~\ref{sec:alternative_inv_proc}.

\section{Mean-adjusted interpolation}
\label{sec:remedies_normalized_ccs}
To summarize from last section, the degenerate outputs are likely caused by small biases from imperfections in the noise estimation model that are amplified by the necessary norm adjustment.

In order to find a suitable remedy, we first acknowledge that inverted latents $\mathbf{z}$ may not be correctly modeled as pure noise. Instead, we suggest to model them as a sum of a deterministic part $\mathbf{d}$ and a noise part $\mathbf{e}$, letting $\mathbf{z} = \mathbf{d} + \mathbf{e}$. Given some method for decomposing $\mathbf{z}$ into $\mathbf{d}$ and $\mathbf{e}$ terms, we can treat $\mathbf{d}$ and $\mathbf{e}$ differently. The noise part $\mathbf{e}$ is the main term, where norm adjustment is absolutely critical. Therefore, it makes sense to interpolate $\mathbf{e}$ using one of the norm-adjustment schemes in Section~\ref{sec:interp_and_cc}, while regular linear interpolation might suffice for $\mathbf{d}$ to avoid amplifying the bias. The final interpolated $\mathbf{z}'$ could then be the sum of the interpolated deterministic and noise parts. In other words, let $\mathbf{d}' = f_\textrm{LIN}(\mathbf{D}, \mathbf{w})$, $\mathbf{e}' = f_*(\mathbf{E}, \mathbf{w})$, and $\mathbf{z}' = \mathbf{d}' + \mathbf{e}'$, where $f_*$ could be any norm-adjusted interpolation. We will consider using $f_\textrm{FIX}$ or $f_\textrm{NIN}$ as $f_*$.

Note that the desired input reproduction property from Section~\ref{sec:interp_and_cc} is fulfilled if using $f_\textrm{NIN}$, since if $\mathbf{w} = \{1, 0, \ldots\}$, then $\mathbf{z}' = \mathbf{d}_1 + \|\mathbf{\mathbf{e_1}}\| / \|\mathbf{\mathbf{e_1}}\| \mathbf{e}_1 = \mathbf{z}_{1}$. Also note that this holds regardless of the method used for estimating $\mathbf{d}$ and $\mathbf{e}$ from $\mathbf{z}$. However, this property does not hold if we use $f_\textrm{FIX}$ instead of $f_\textrm{NIN}$.

What remains to determine is a method for estimating an approximate split of $\mathbf{z}$ into terms $\mathbf{d}$ and $\mathbf{e}$. Some options could be:
\begin{enumerate}
\item Approximate $\mathbf{d} = 0$. This represents a baseline choice and is equivalent to simply using $f_\textrm{NIN}$ or $f_\textrm{FIX}$ from Section~\ref{sec:baseline_interp_options} directly.
\item Approximate $\mathbf{d}$ as the mean value of $\mathbf{z}$ over all channels and spatial dimensions.
\item Approximate $\mathbf{d}$ using the mean value of each feature channel in $\mathbf{z}$ separately, i.e. let $\mathbf{d}$ be a constant signal over spatial dimensions but with a distinct value per feature channel.
\item Approximate $\mathbf{d}$ using a low-pass-filtered version of $\mathbf{z}$.
\end{enumerate}
We aim for a simple normalization scheme that can easily be integrated into other methods and opt for comparing options (2) and (3), keeping option (1) as a baseline choice. More advanced options are left as future research.

\section{Experimental results}
\label{sec:results}
In this section, we examine experimentally whether the suggested normalization procedures reduce degeneracies. All evaluation is done on images from ImageNet, due to its wide availability. For more experimental details, see Appendix~\ref{sec:additional_exp_details}.

\mysubheading{Evaluation metrics.}
To measure image quality, we use two common measures; the FID metric~\cite{heusel2017gans} and CLIP~\cite{radford2021learning} embedding distance. We acknowledge that the FID metric has been criticized for not always aligning well with human assessment and for being sensitive to the choice of resampling operations and number of examples, and that alternatives have been suggested~\cite{chong2020effectively,jayasumana2024rethinking,khanna2023diffusionsat}. However, as we are interested in quantifying grave degradations rather than precisely comparing the quality of competing high-aesthetic outputs, we opt for using the original metric due to its wide availability. For the CLIP distance, the CLIP embedding of images produced from latent centroids are compared to the mean image embedding of all training examples for the class using the cos distance.

\begin{figure*}[t!]
  \centering
  \includegraphics[width=0.49\linewidth]{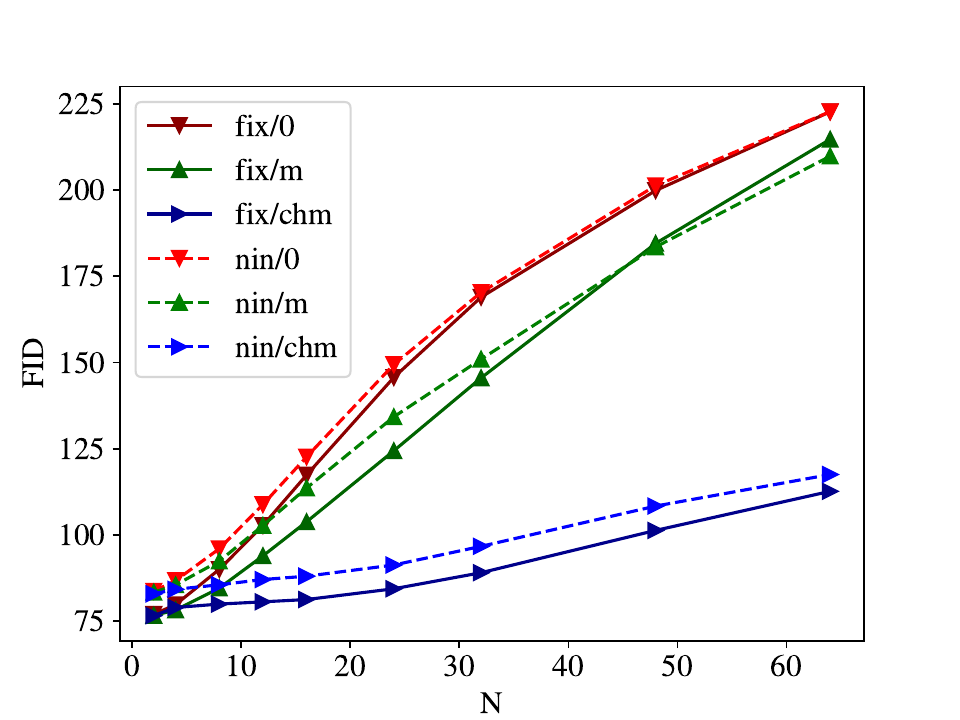}
  \includegraphics[width=0.49\linewidth]{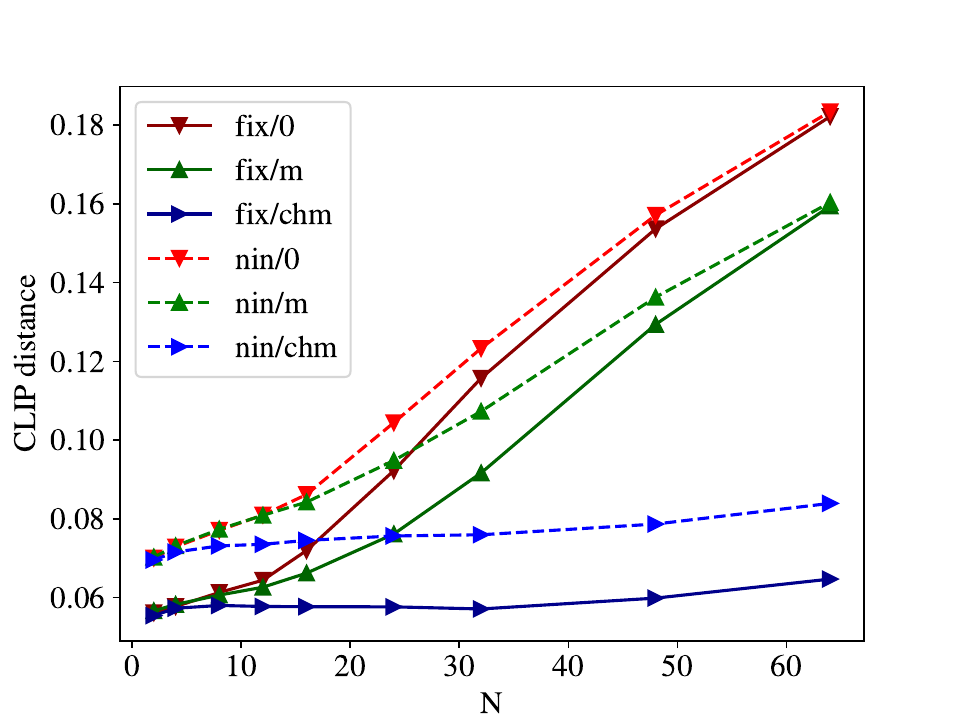}
 
  \caption{Image quality of images produced from centroids of $N$ noisy diffusion model latents, measured using the FID metric and cos distance in CLIP embedding space.}

  \label{fig:main_results}
\end{figure*}

\mysubheading{Mean adjustment results.}
Quality metrics for the compared methods are shown in Figure~\ref{fig:main_results}. Here, we let \texttt{fix} and \texttt{nin} denote norm adjustment according to $f_\textrm{FIX}$ and $f_\textrm{NIN}$ from Section~\ref{sec:baseline_interp_options}, while the suffixes \texttt{/0}, \texttt{/m} and \texttt{/chm} denote the choice of mean adjustment (none, global mean, or channel-wise mean) according to Section~\ref{sec:remedies_normalized_ccs}. The figure shows that the quality drops dramatically as $N$ grows using baseline methods. The mean-adjusted options are slightly better, while the channel-wise mean adjustment options provide a significant remedy. These overall trends are similar between the FID and CLIP measures. Some qualitative examples of centroid images produced using these methods can be found in Figure~\ref{fig:qualitative_ex_monarch}-\ref{fig:qualitative_ex_manyclasses}, with one more example in Appendix~\ref{sec:more_qual_ex}. In Figure~\ref{fig:qualitative_ex_manyclasses}, all examples are produced using the same $N$. Not all examples are degenerate using the baseline option at this $N$, but also for the non-degenerate examples, the apparent visual quality is better with the mean-adjusted method. We do note that the results are not always perfect. There is a measurable drop in quality also for the \texttt{chm} methods as $N$ grows sufficiently large. Qualitatively, this often manifests in over-saturated colors and loss of detail, as visible in the high $N$ examples in Figure~\ref{fig:intro_example}.

Figure~\ref{fig:main_results} also shows that using $f_\textrm{FIX}$ produces slightly better quality metrics than $f_\textrm{NIN}$ most of the time, both with and without any mean adjustment in place. This was surprising to us. We hypothesize that the reason for this behavior is that the computed latent centroids are often far enough away from the input $\mathbf{z}_n$ to render the norms of the inputs non-representative as suitable normalization targets for the centroid. Using the nominal norm $\sqrt{L}$ seems to be a slightly better choice for such latents. However, this difference is significantly smaller than the difference caused by the degeneracy issue, and the apparent visual quality is often similar (as in Figure~\ref{fig:qualitative_ex_monarch}). We therefore consider both \texttt{fix/chm} and \texttt{nin/chm} to be reasonable options.

\begin{figure*}[t!]
\centering
  \includegraphics[width=0.14\linewidth]{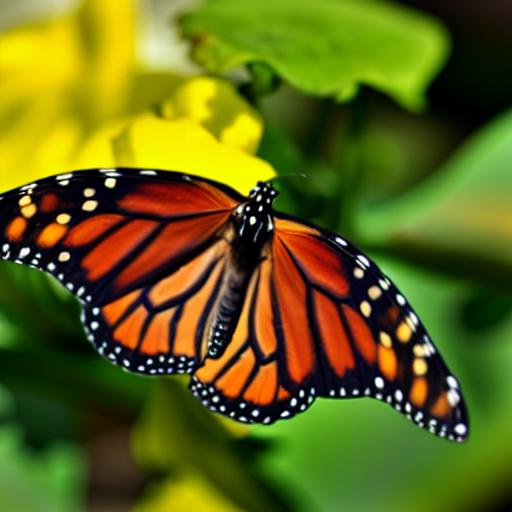}
  \includegraphics[width=0.14\linewidth]{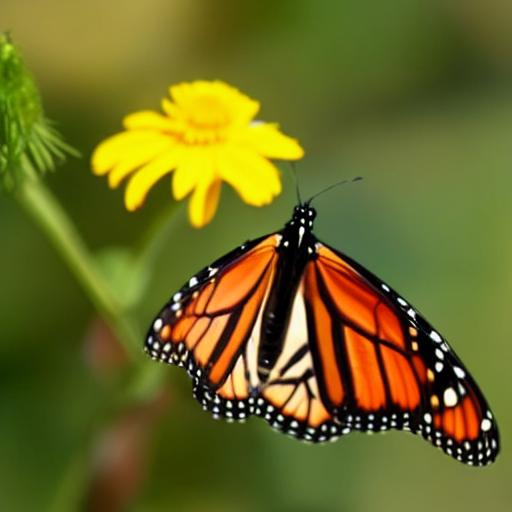}
  \includegraphics[width=0.14\linewidth]{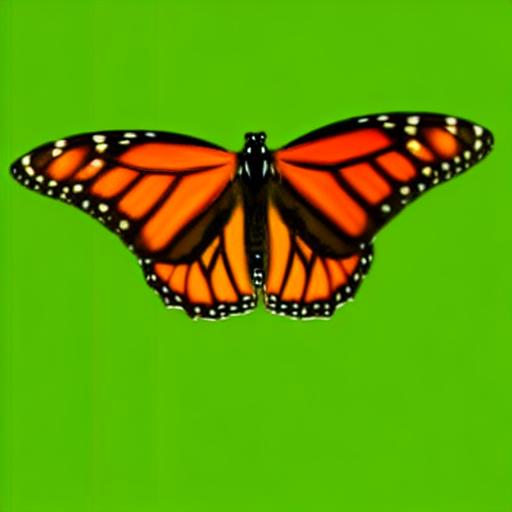}
  \includegraphics[width=0.14\linewidth]{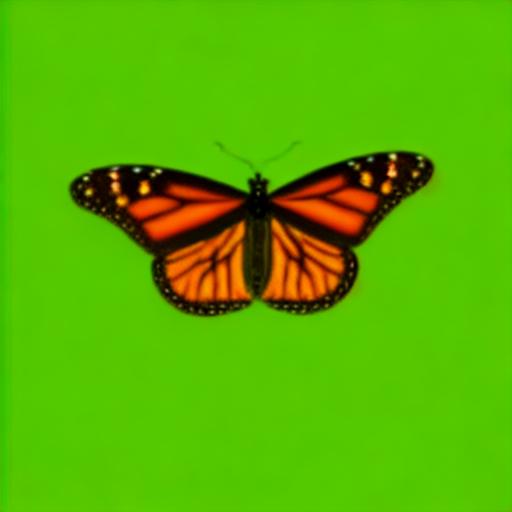} 
  \includegraphics[width=0.14\linewidth]{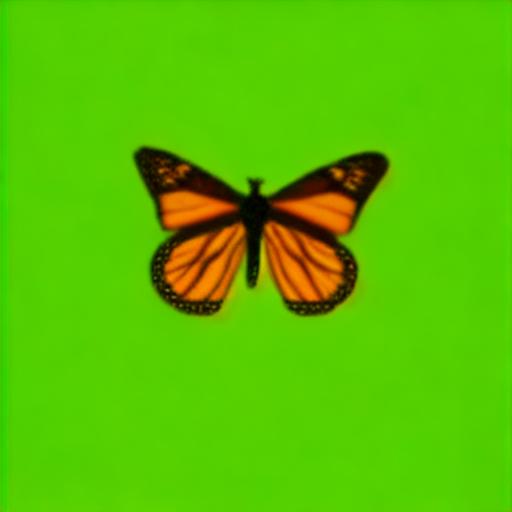}
  \includegraphics[width=0.14\linewidth]{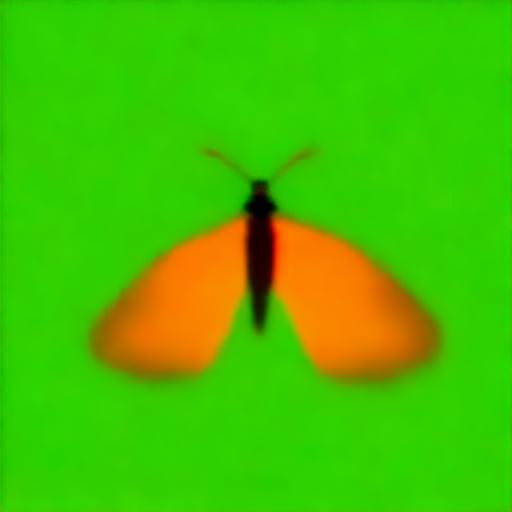} \\
  \vspace{0.2cm}
  \includegraphics[width=0.14\linewidth]{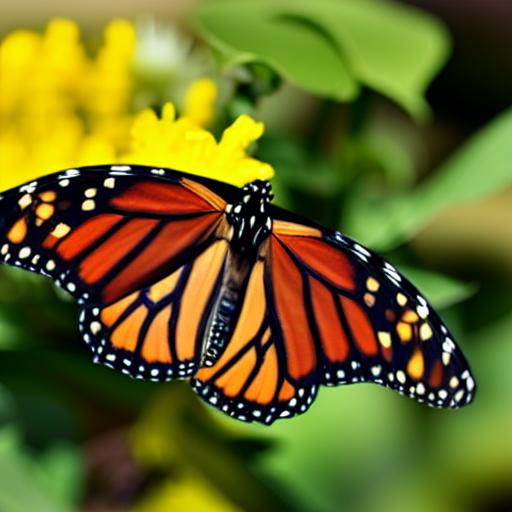}
  \includegraphics[width=0.14\linewidth]{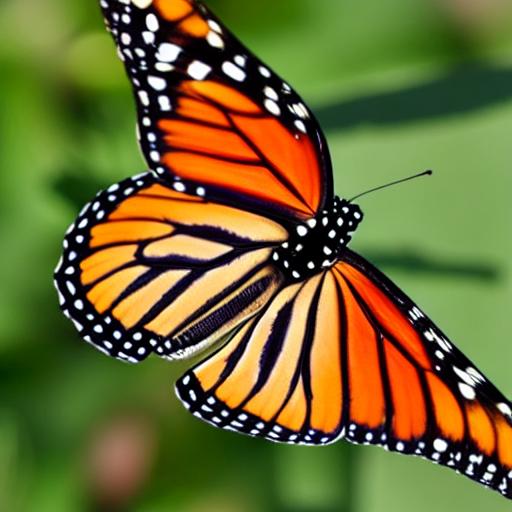}
  \includegraphics[width=0.14\linewidth]{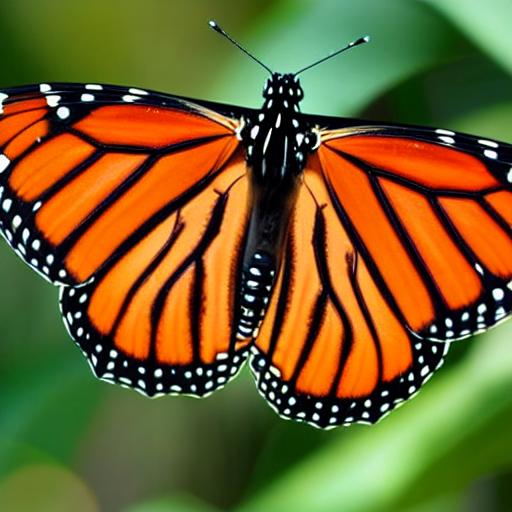}
  \includegraphics[width=0.14\linewidth]{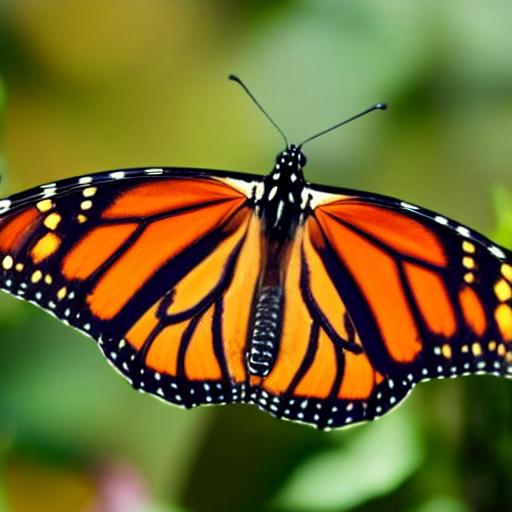}
  \includegraphics[width=0.14\linewidth]{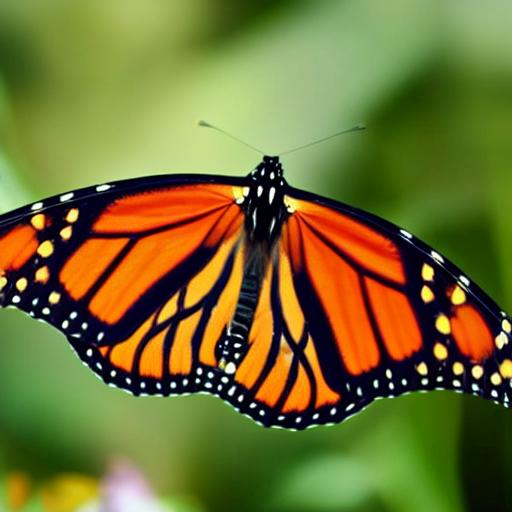}
  \includegraphics[width=0.14\linewidth]{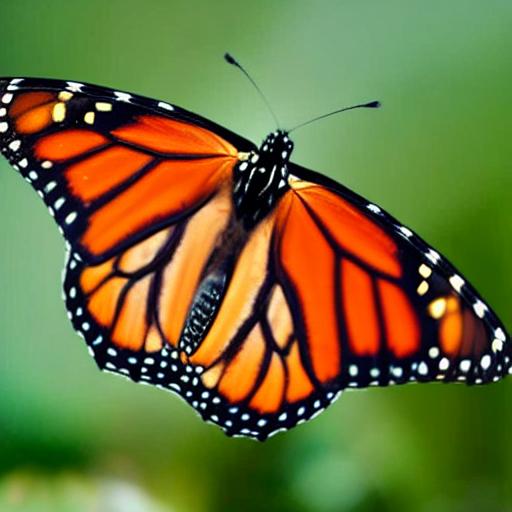} \\
  \vspace{0.2cm}
  \includegraphics[width=0.14\linewidth]{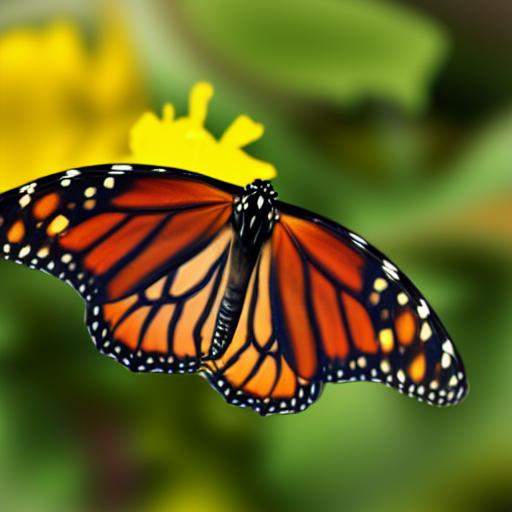}
  \includegraphics[width=0.14\linewidth]{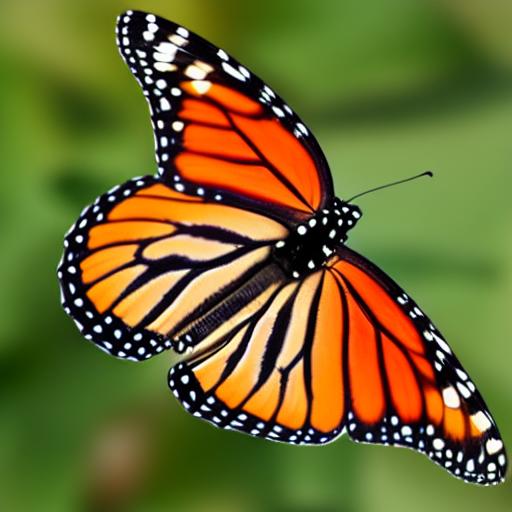}
  \includegraphics[width=0.14\linewidth]{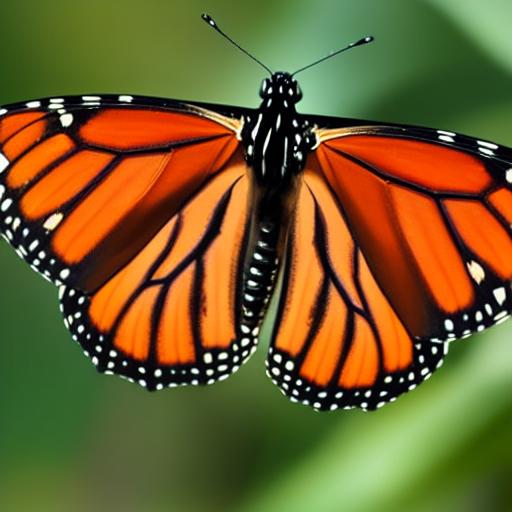}
  \includegraphics[width=0.14\linewidth]{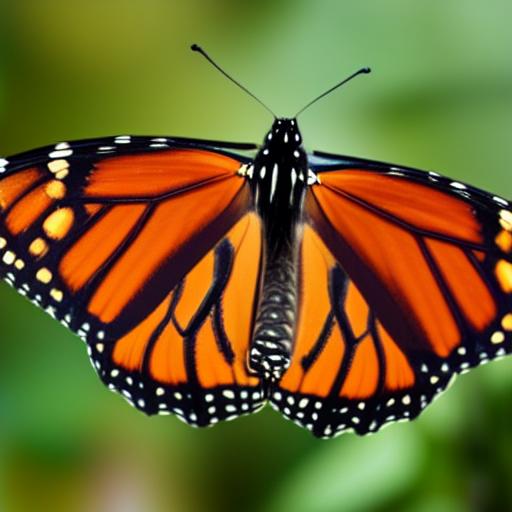} 
  \includegraphics[width=0.14\linewidth]{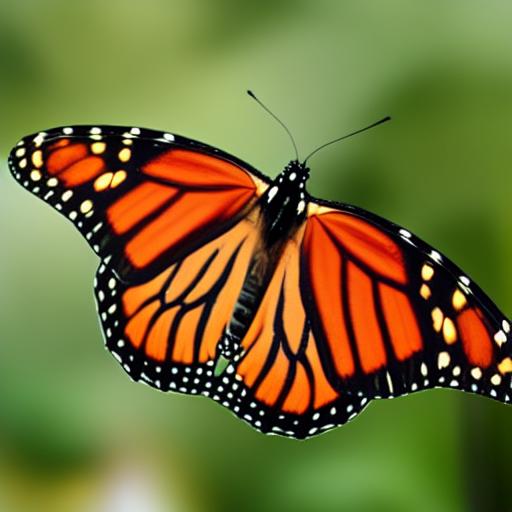}
  \includegraphics[width=0.14\linewidth]{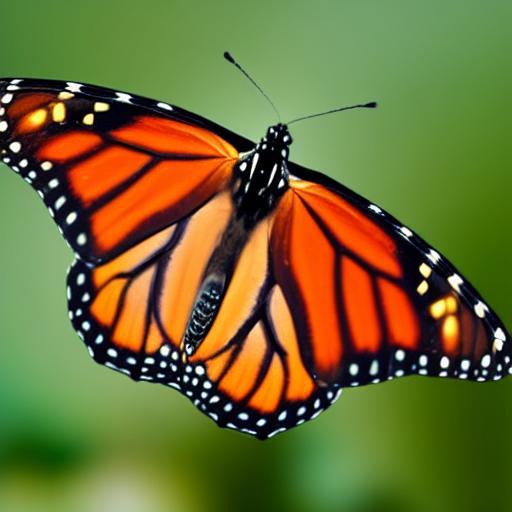} \\
 
  \caption{Images generated from centroids of latents obtained from $N$ input images of ImageNet class ``monarch butterfly'' for $N = 2, 8, 32, 48, 64, 96$, SD 1.5. Top row: Fixed normalization. Middle row: \texttt{fix/chm}. Bottom row: \texttt{nin/chm}.}

  \label{fig:qualitative_ex_monarch}
\end{figure*}

\begin{figure*}[t!]
\centering
  \includegraphics[width=0.17\linewidth]{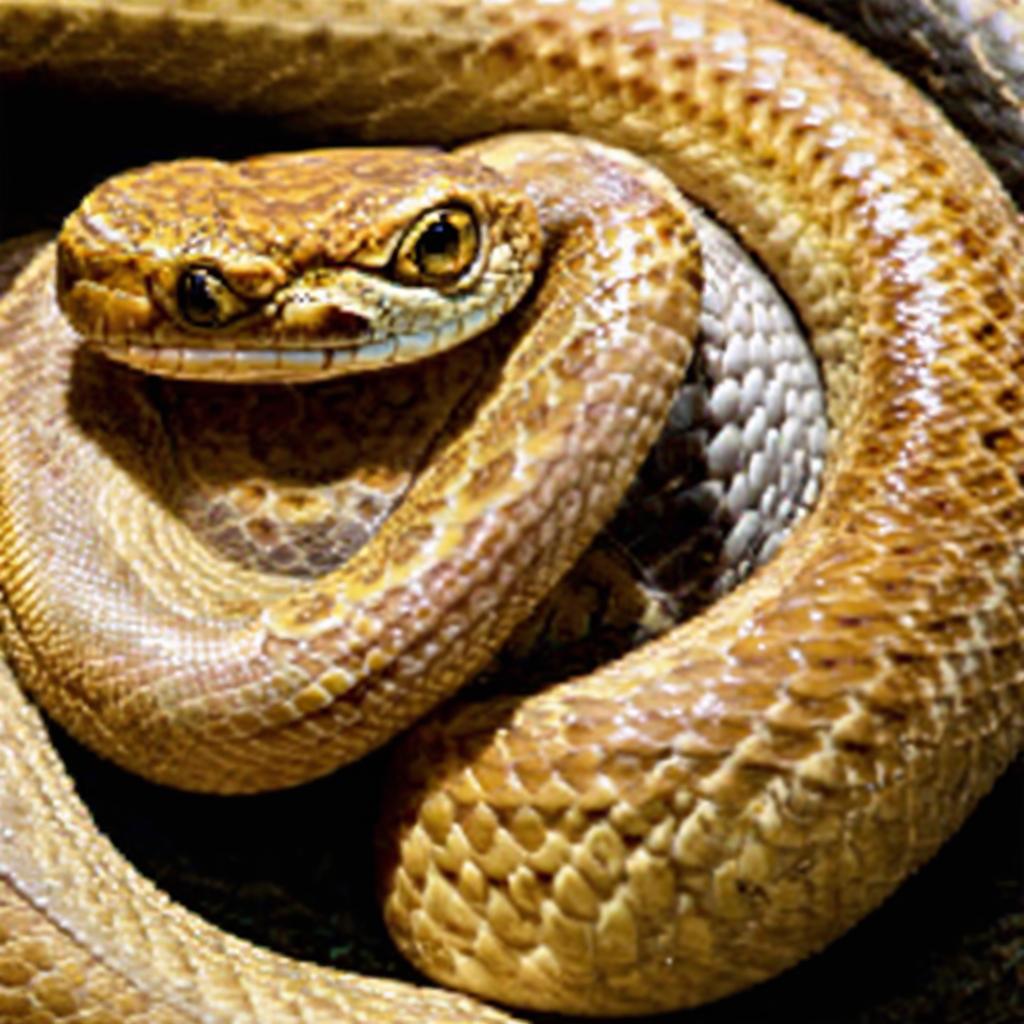}
  \includegraphics[width=0.17\linewidth]{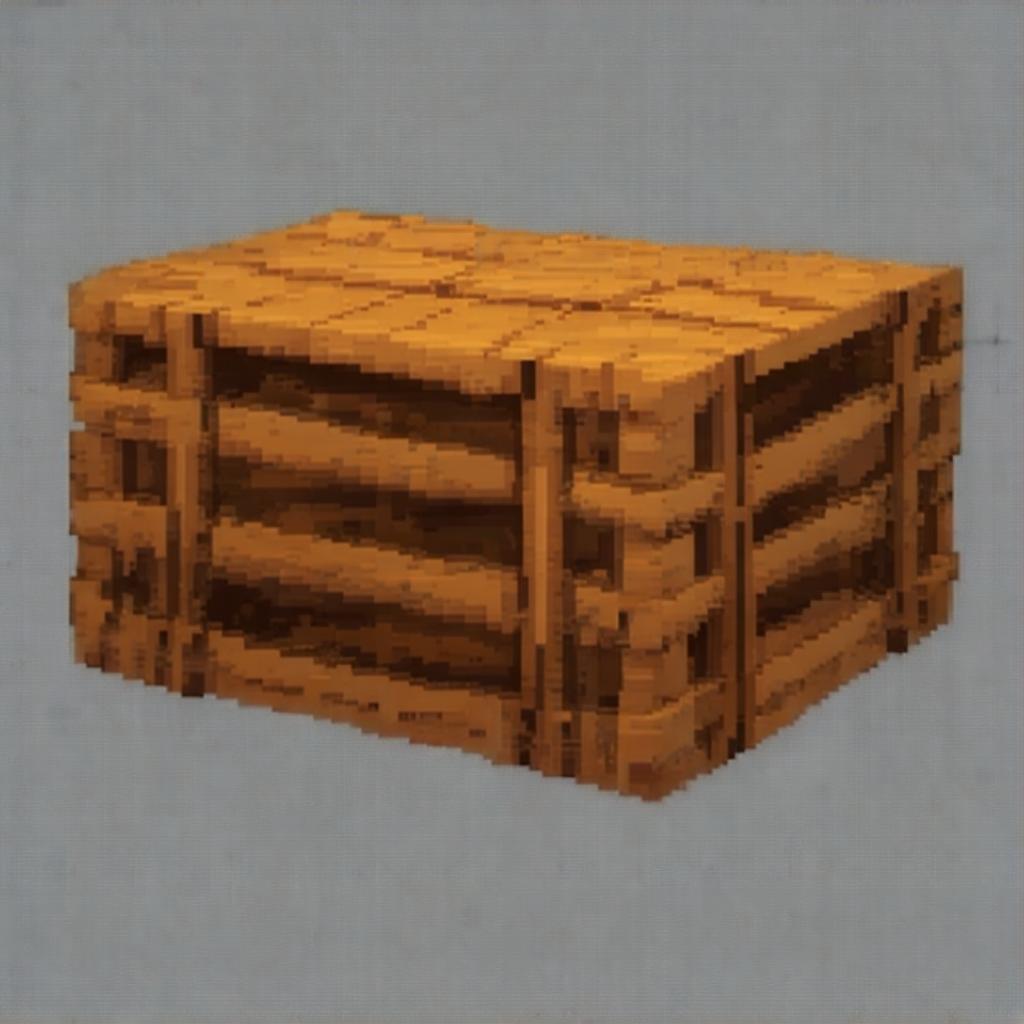}
  \includegraphics[width=0.17\linewidth]{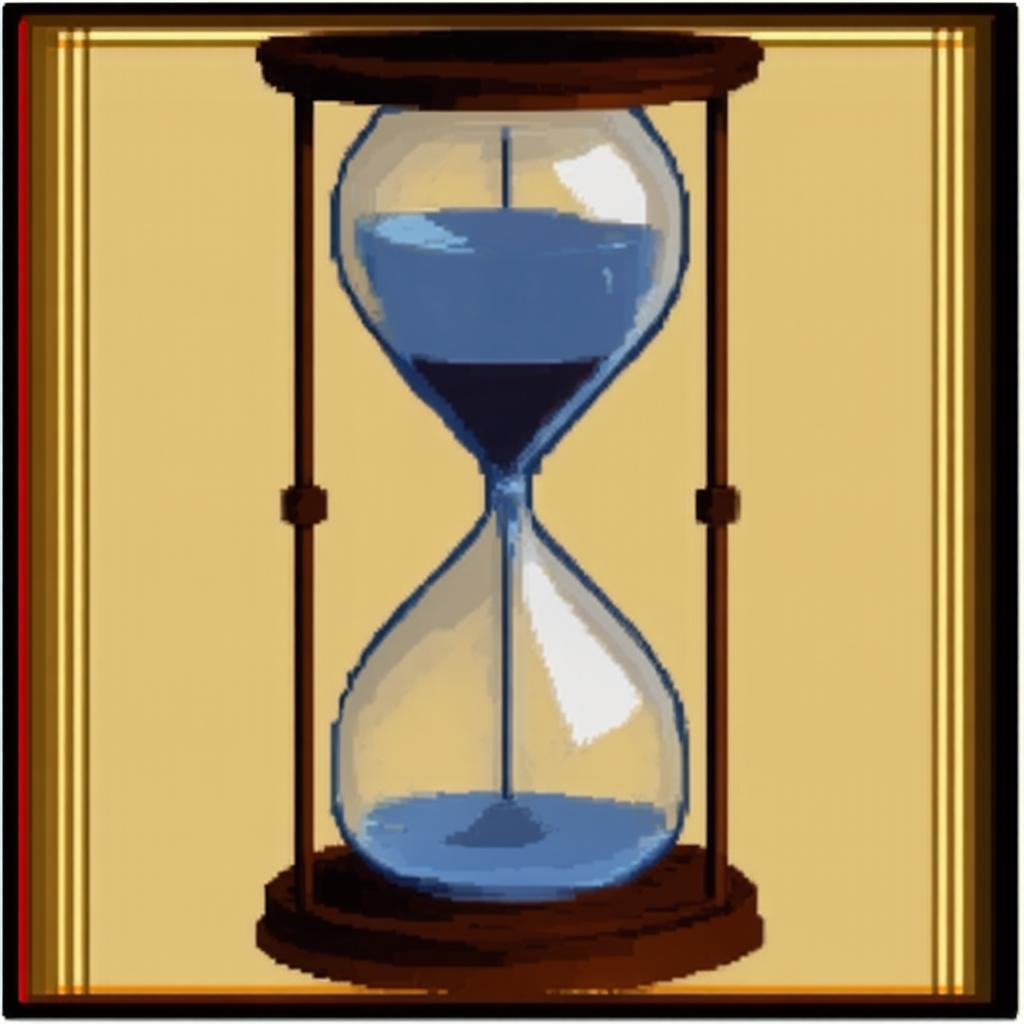}
  \includegraphics[width=0.17\linewidth]{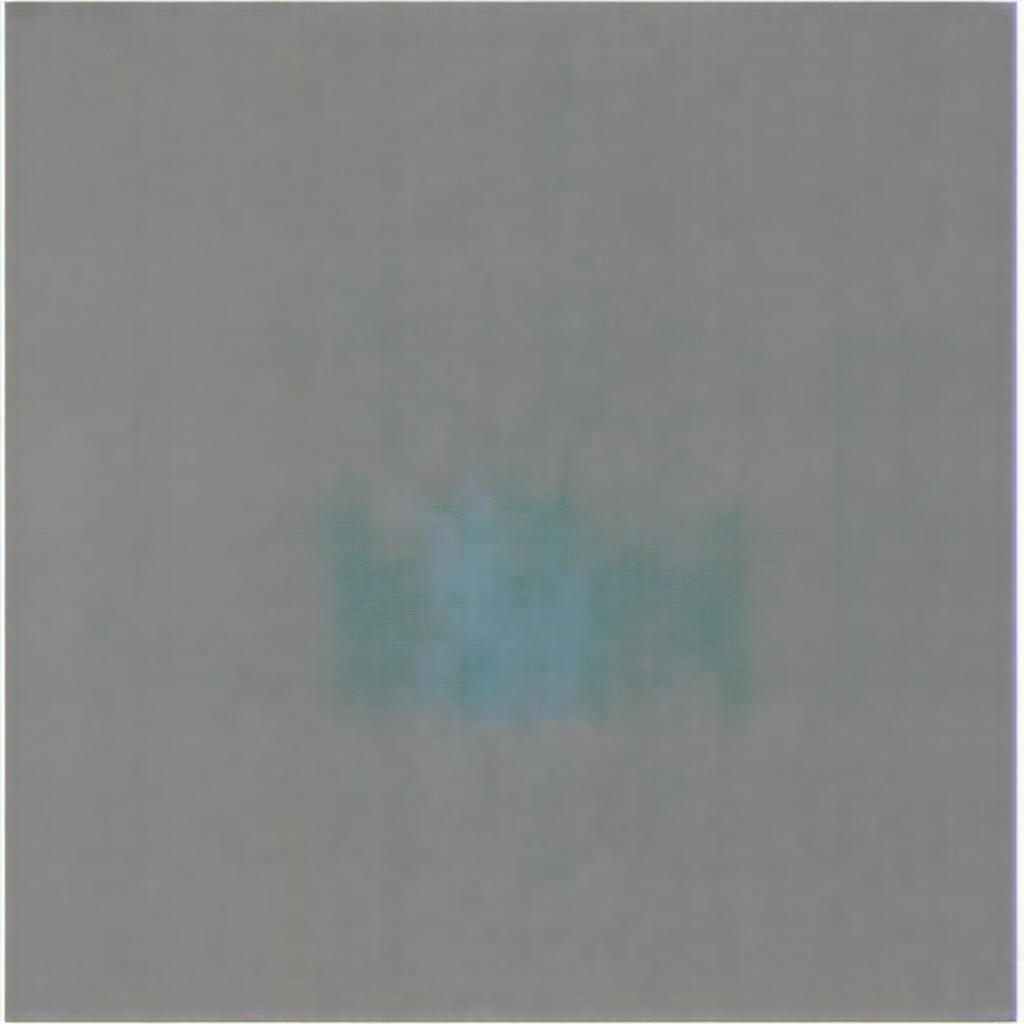}
  \includegraphics[width=0.17\linewidth]{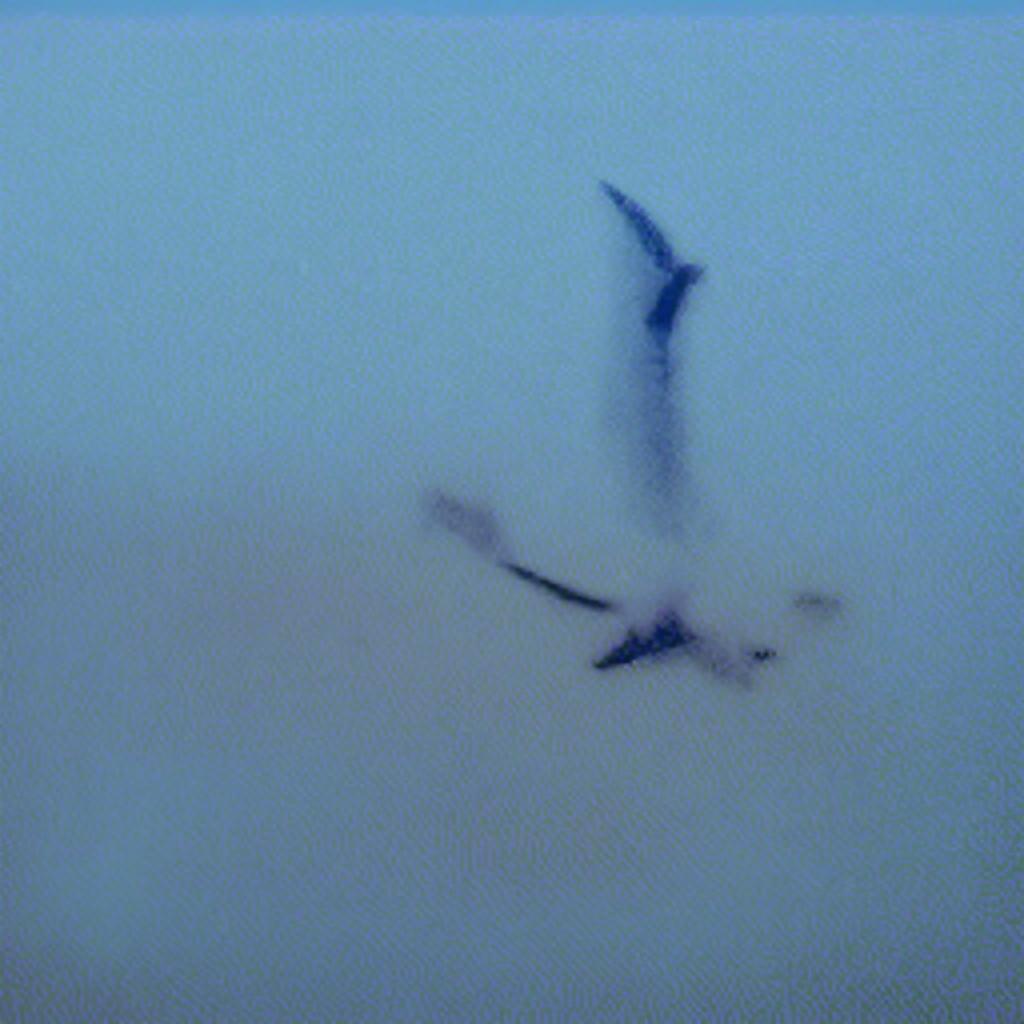} \\
  \vspace{0.2cm}
  \includegraphics[width=0.17\linewidth]{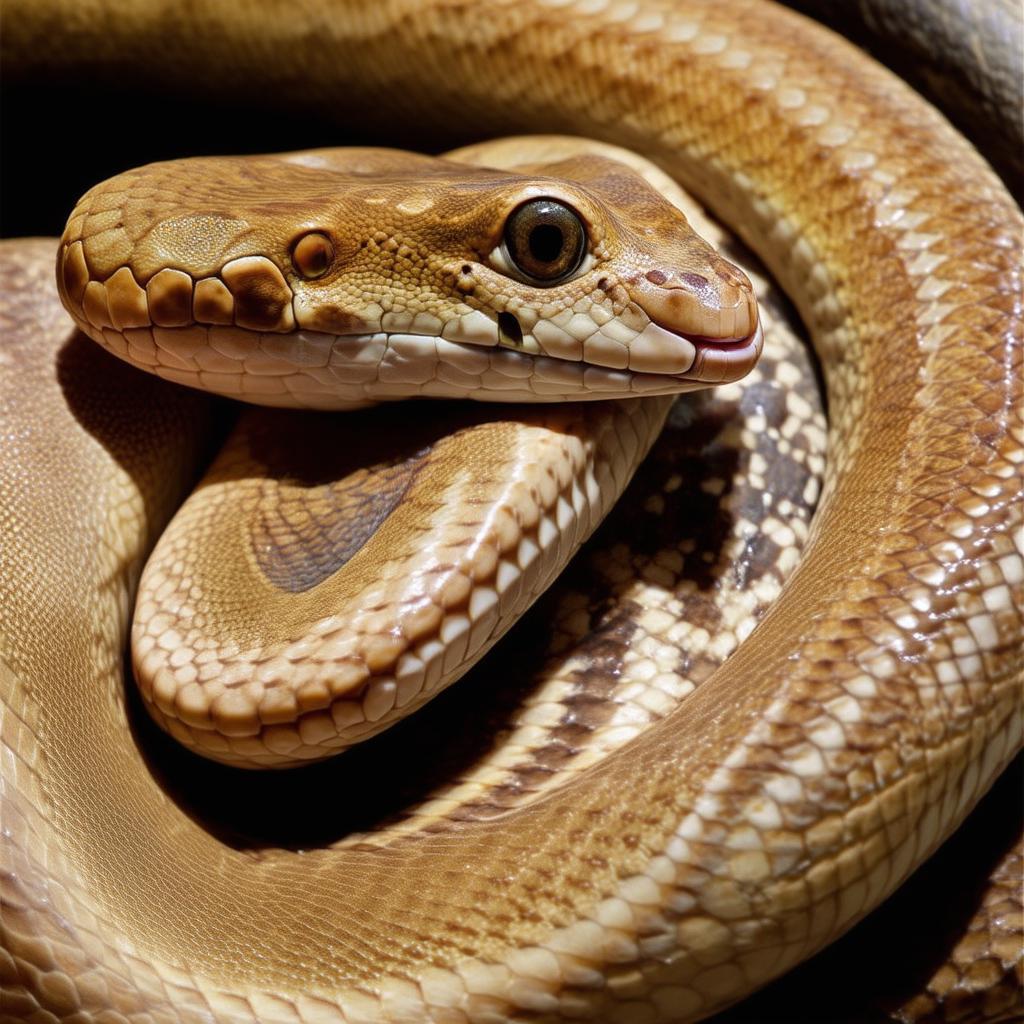}
  \includegraphics[width=0.17\linewidth]{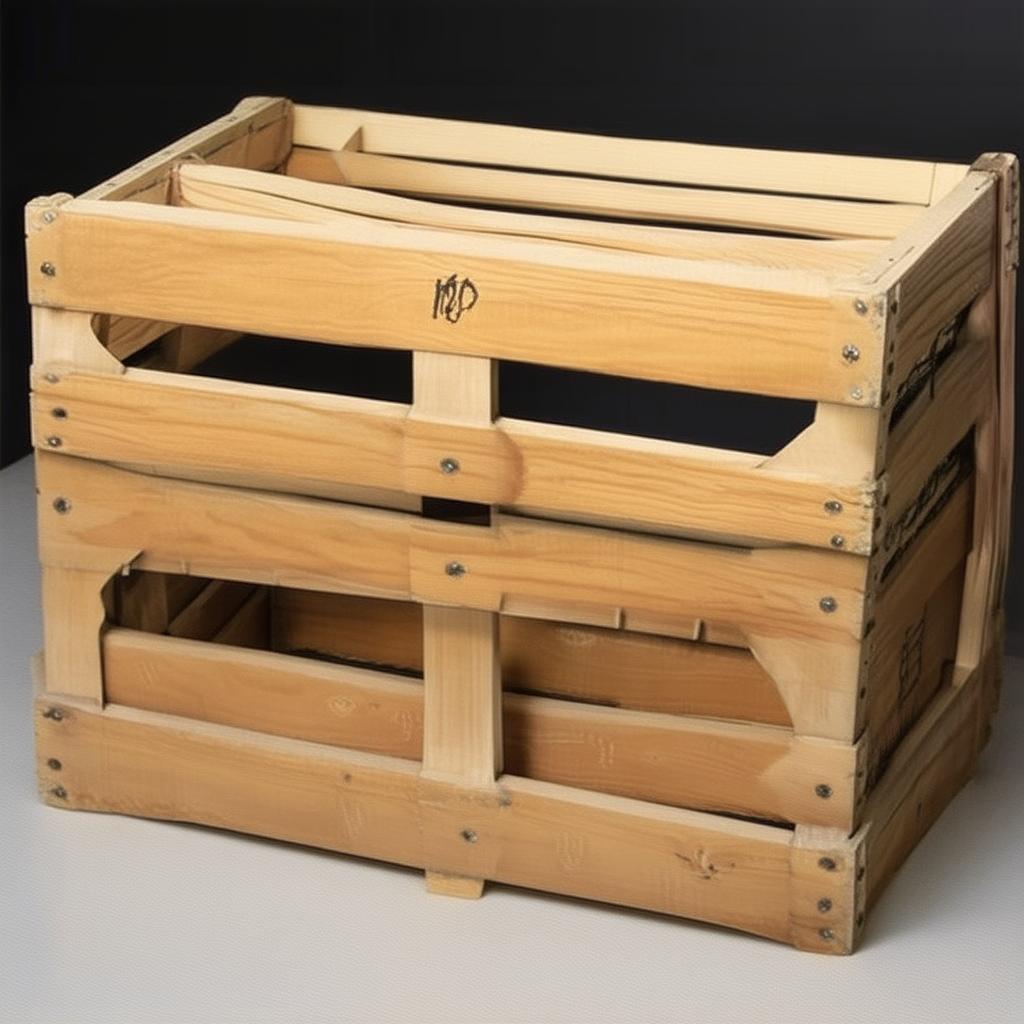}
  \includegraphics[width=0.17\linewidth]{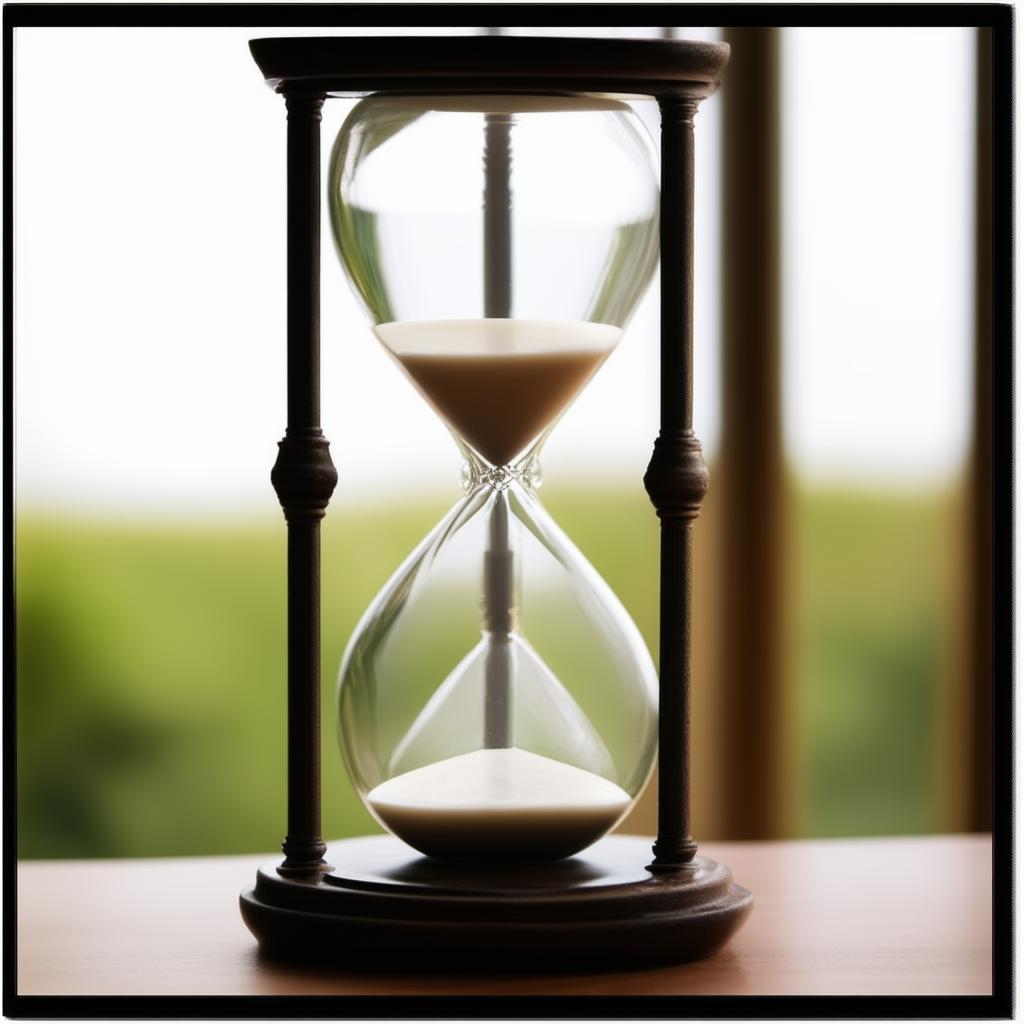}
  \includegraphics[width=0.17\linewidth]{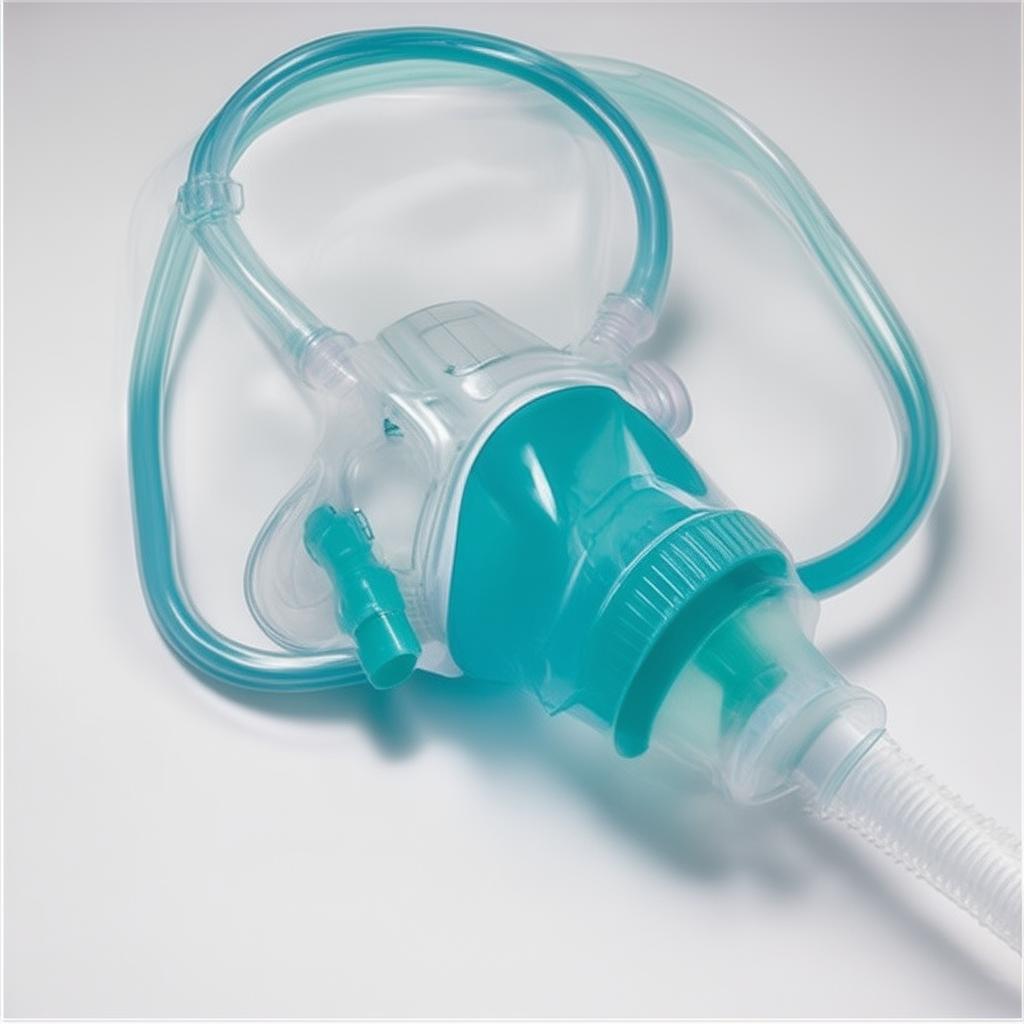}
  \includegraphics[width=0.17\linewidth]{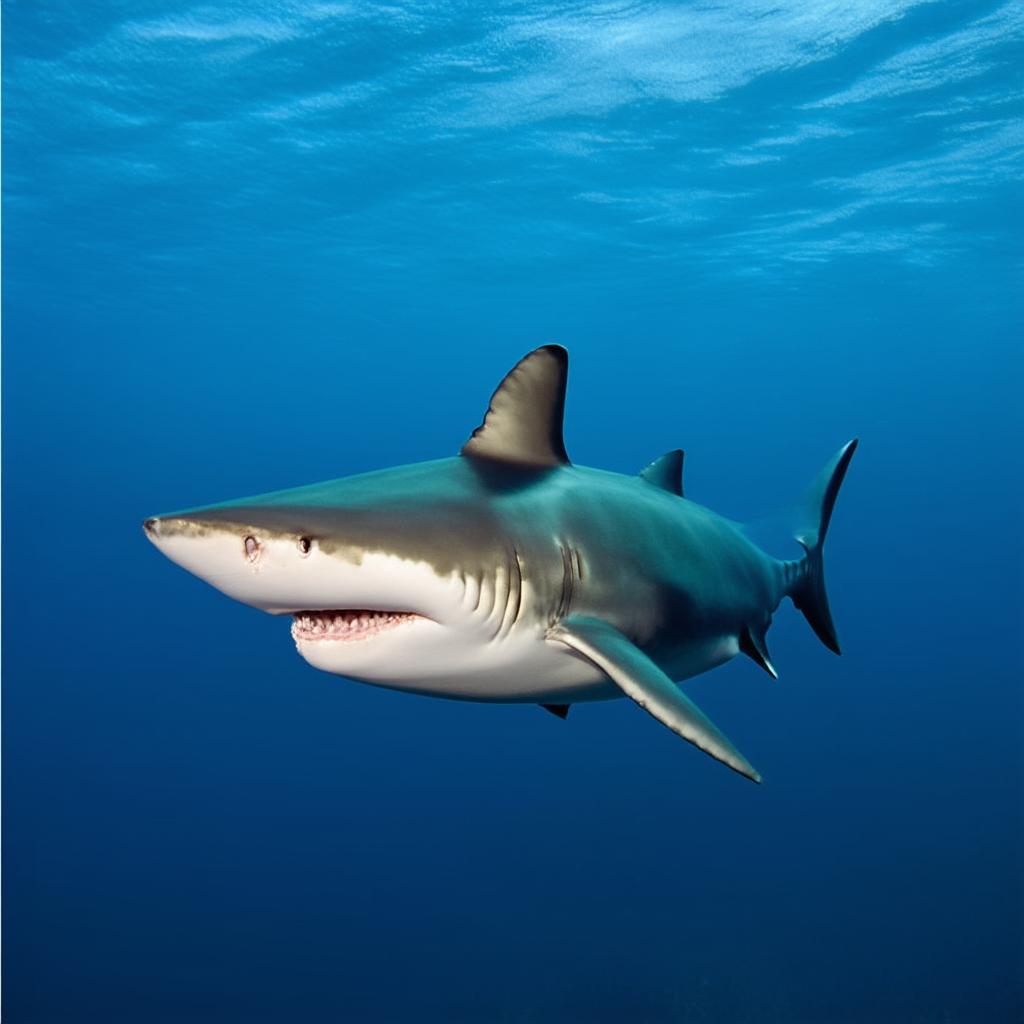}
 
  \caption{Images generated from centroids of latents obtained from $N = 12$ input images of ImageNet classes ``boa constrictor'', ``crate'', ``hourglass'', ``oxygen mask'', ``shark'' using SD 3.5. Top row: Fixed normalization. Bottom row: \texttt{nin/chm}.}

  \label{fig:qualitative_ex_manyclasses}
\end{figure*}

\section{Discussion}
We have identified and dissected a problem with interpolating diffusion model latents, and suggested a modified normalization scheme that offers a significant remedy. We consider the work foundational in nature, aiming at a better understanding of the intricacies of diffusion models and their latent spaces. As such, the work is not tied to a particular application. However, the most direct applications would be methods for image morphing using diffusion models. Fixing the degeneracies opens up for generalizing such methods to many inputs, producing latent space manifolds that can be traversed by adjusting the interpolation weights. Other potential applications are in deep data augmentation for label-constrained learning problems. Exploring such applications is a direction for future research.

One surprising discovery was that $f_\textrm{FIX}$ leads to slightly better quality metrics than $f_\textrm{NIN}$. Recall that the motivation behind $f_\textrm{NIN}$ was to ensure interpolation paths without discontinuities close to the original $\mathbf{z}_n$. It is possible to construct other interpolation paths that are continuous close to the original $\mathbf{z}_n$ while approaching the nominal value $\sqrt{L}$ when we get sufficiently far away from the original inputs. One example in this direction is the norm-aware optimization suggested by Samuel et al.~\cite{samuel2024norm}, but that method relies on a cumbersome iterative procedure. Examining and evaluating more convenient options that can be formulated in closed form is also a subject for future research.

% \ackname as suggested in the instructions doesn't work
\mysubheading{Acknowledgements.}
The research work was supported by the Wallenberg AI, Autonomous Systems and Software Program (WASP) funded by the Knut and Alice Wallenberg Foundation, and the computations by the National Academic Infrastructure for Supercomputing in Sweden (NAISS), partially funded by the Swedish Research Council through grant agreement no.\ 2022-06725.

\mysubheading{Disclosure of Interests.}
The authors have no competing interests to declare that are relevant to the content of this article.

\bibliographystyle{splncs04}
\bibliography{main}

\appendix
\clearpage
\setcounter{page}{1}

\section{Additional experimental details}
\label{sec:additional_exp_details}
All experiments were implemented in PyTorch with the HuggingFace Diffusers library. Except where otherwise stated, examples and results were produced using Stable Diffusion 1.5. Initial preliminary experiments showed similar effects across versions 1.4, 1.5, 2.1 and 3.5.

All experiments used the default classifier-free guidance values of 7.5 for SD 1.5 and 4.5 for SD 3.5, except where stated that no guidance was used. Experiments using SD 3.5 used the \emph{medium} model size.

The null-text inversion experiments were run using 32-bit float precision, due to the use of optimization using backpropagation. The high-iteration DDIM-inversion experiments with SD 1.5 also used 32-bit precision in order to get good numerical conditions. The high-iteration DDIM inversion experiments with SD 3.5 used 16-bit precision due to GPU memory constraints (we wanted all experiments to be runnable using 24 Gb of VRAM). All other experiments were run with 16-bit float precision in the image generation but 32-bit precision in the interpolation operations in order to ensure that the interpolation was not affected negatively by the limited precision. Finally, we note that the observed bias (that is amplified by the normalization), is significantly larger than the 16-bit floating-point epsilon.

The final results in Section~\ref{sec:results} were produced by comparing 1000 computed centroids from up to 64 examples of one class from the training split of ImageNet, averaged over 10 randomly drawn classes. This experiment required around 300 GPU-hours using NVidia A40 GPUs. The additional compute required for the suggested normalization was negligible compared to running the diffusion model. 

\section{Alternative inversion procedures}
\label{sec:alternative_inv_proc}
\begin{figure*}[t!]
\centering
  \includegraphics[width=0.14\linewidth]{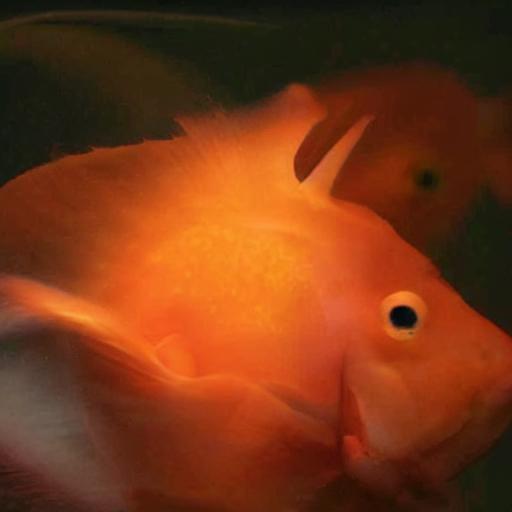}
  \includegraphics[width=0.14\linewidth]{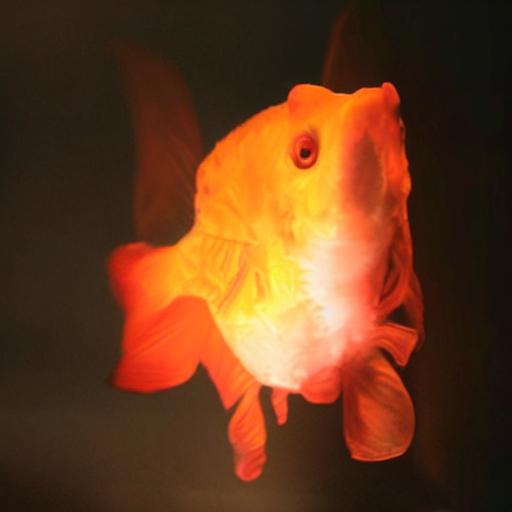}
  \includegraphics[width=0.14\linewidth]{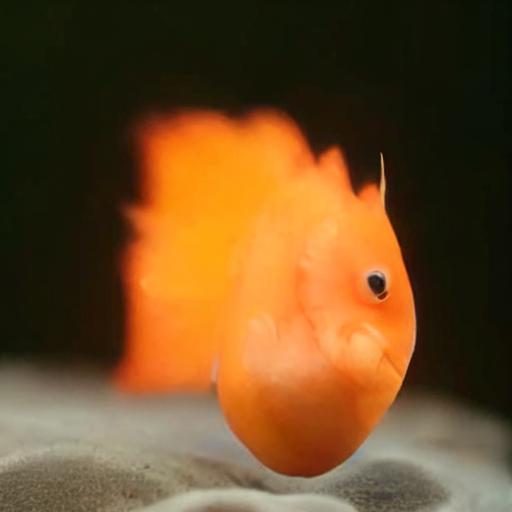}
  \includegraphics[width=0.14\linewidth]{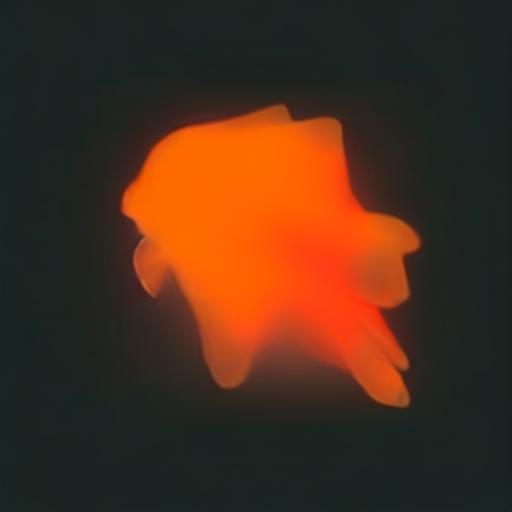} 
  \includegraphics[width=0.14\linewidth]{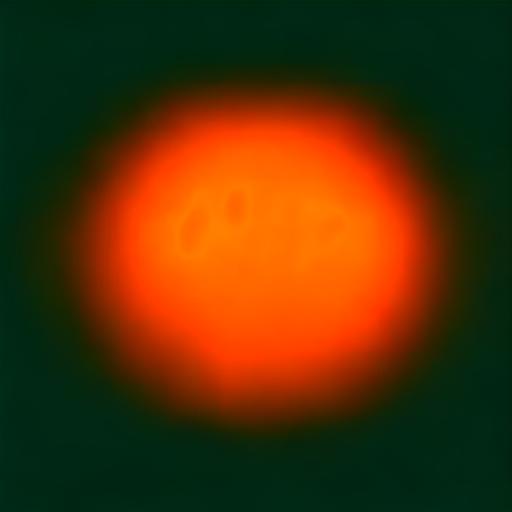}
  \includegraphics[width=0.14\linewidth]{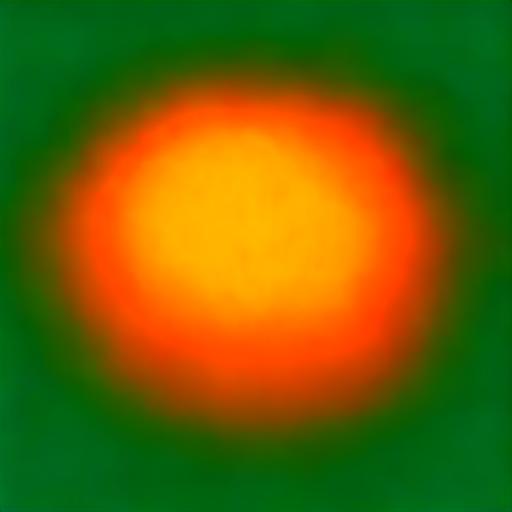} \\
  \vspace{0.2cm}
  \includegraphics[width=0.14\linewidth]{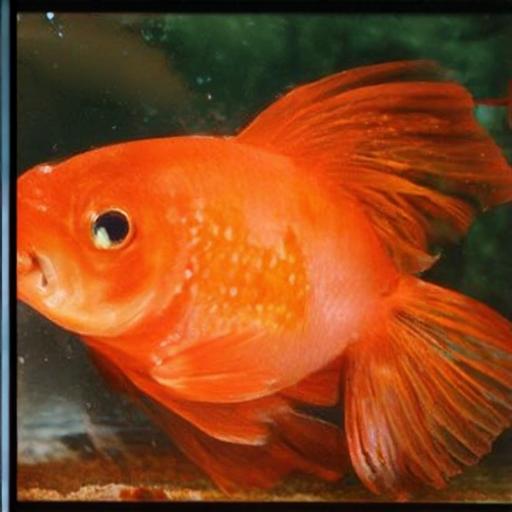}
  \includegraphics[width=0.14\linewidth]{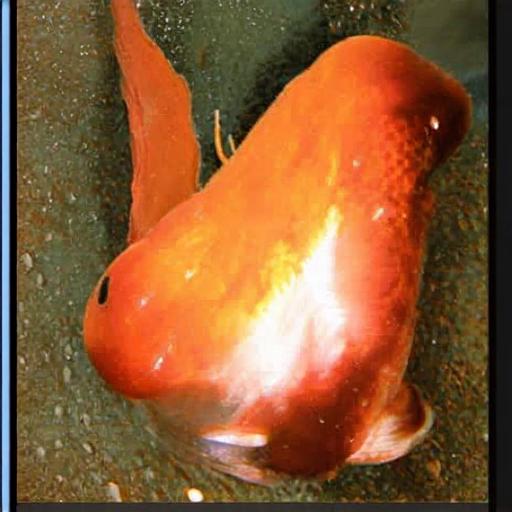}
  \includegraphics[width=0.14\linewidth]{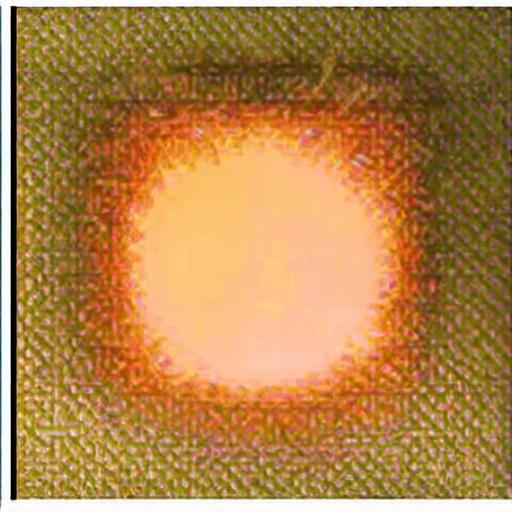}
  \includegraphics[width=0.14\linewidth]{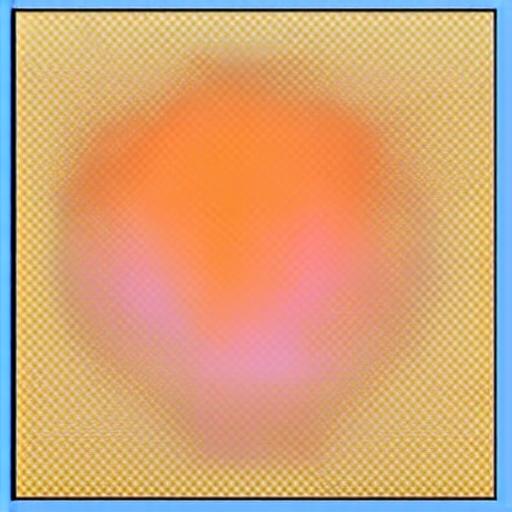} 
  \includegraphics[width=0.14\linewidth]{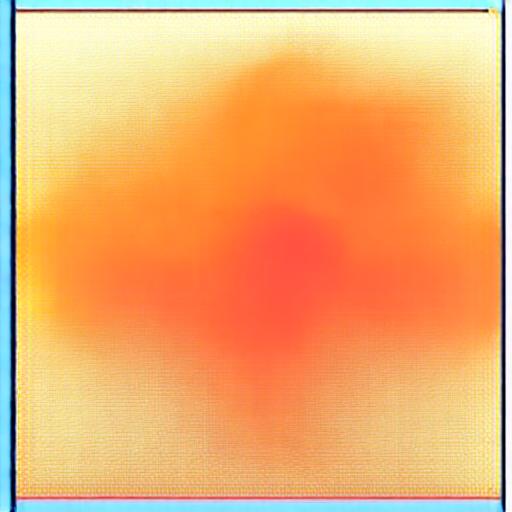}
  \includegraphics[width=0.14\linewidth]{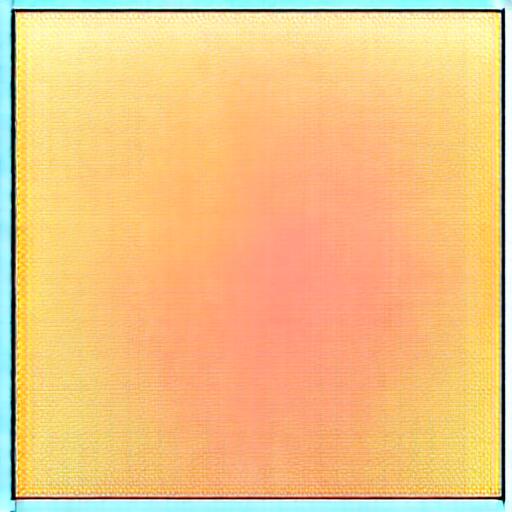}
 
  \caption{Images generated from centroids using 500 diffusion timesteps without classifier-free guidance, showing that the degeneration issue persists also under more ideal DDIM inversion conditions. $N = 2, 4, 8, 16, 32, 64$ (increasing to the right), fixed normalization, ImageNet class ``goldfish''. Top row: SD 1.5, bottom row: SD 3.5.}
  \label{fig:degenerate_although_bestddiminv}
\end{figure*}

To see if the degenerate behavior persists across more inversion procedures, we run two qualitative experiments. The first experiment used DDIM inversion with 500 iterations and without classifier-free guidance. This experiment was run using both SD 1.5 and 3.5, where the 3.5 results are also free from the non-terminal SNR issue mentioned in Section~\ref{sec:origin_of_latent_bias}. A qualitative example is shown in Figure~\ref{fig:degenerate_although_bestddiminv}. As expected, the image quality is worse than when using guidance, but the point here is to show that results are still degenerate for large $N$.

\begin{figure*}[t!]
\centering
  \includegraphics[width=0.14\linewidth]{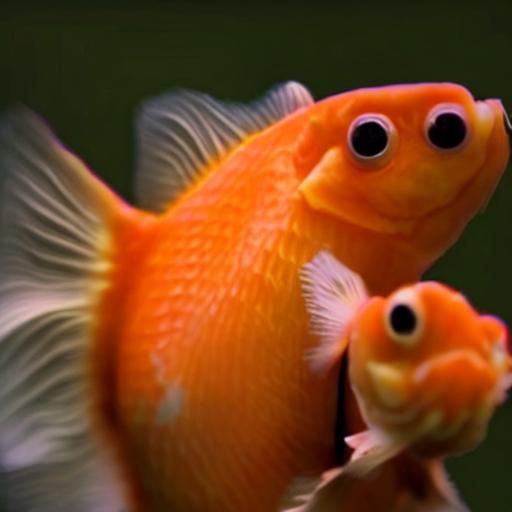}
  \includegraphics[width=0.14\linewidth]{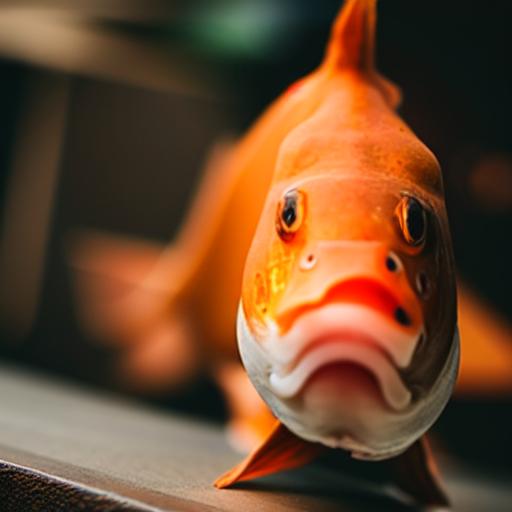}
  \includegraphics[width=0.14\linewidth]{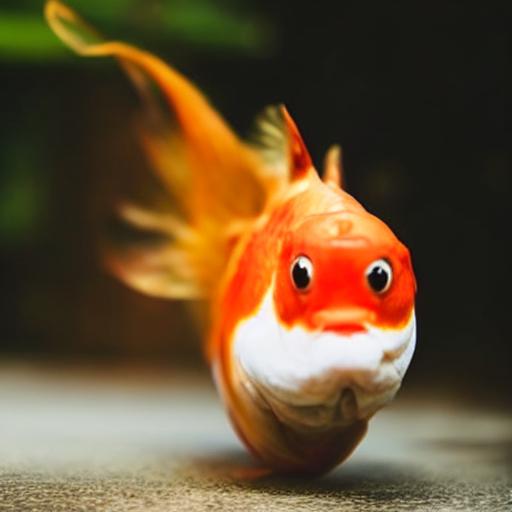}
  \includegraphics[width=0.14\linewidth]{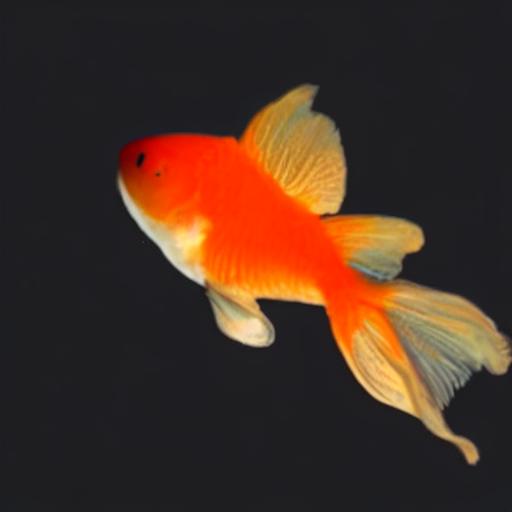} 
  \includegraphics[width=0.14\linewidth]{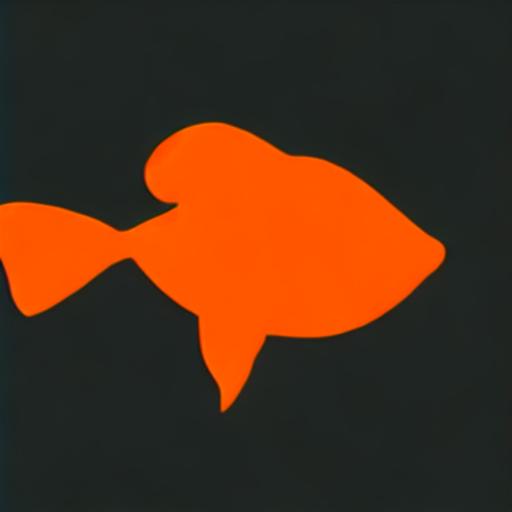}
  \includegraphics[width=0.14\linewidth]{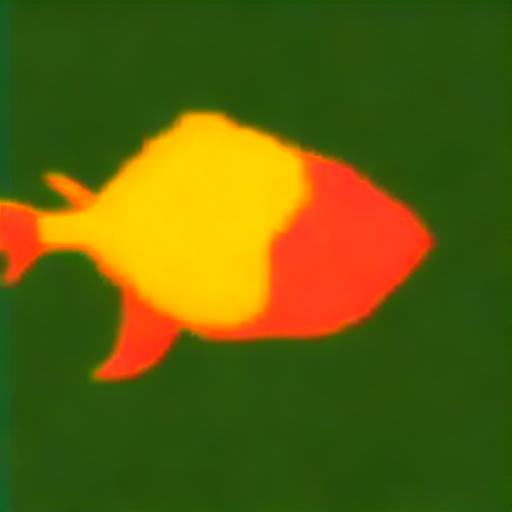}

  \caption{Images generated from centroids computed using null-text inversion, showing that the degeneration issue still persists. $N = 2, 4, 8, 16, 32, 64$ (increasing to the right), fixed normalization, ImageNet class ``goldfish''.}

  \label{fig:degenerate_although_nti}
\end{figure*}

The second experiment used null-text inversion~\cite{mokady2023null}. Latents were interpolated using $f_\textrm{FIX}$ and unconditional embeddings using $f_\textrm{LIN}$, following prior work using linear interpolation for conditioning inputs~\cite{yang2023impus,zhang2024diffmorpher}. A qualitative example is shown in Figure~\ref{fig:degenerate_although_nti}, where the degeneration issue clearly still exists. 

\section{Additional qualitative example}
\label{sec:more_qual_ex}
\begin{figure*}[t!]
\centering
  \includegraphics[width=0.16\linewidth]{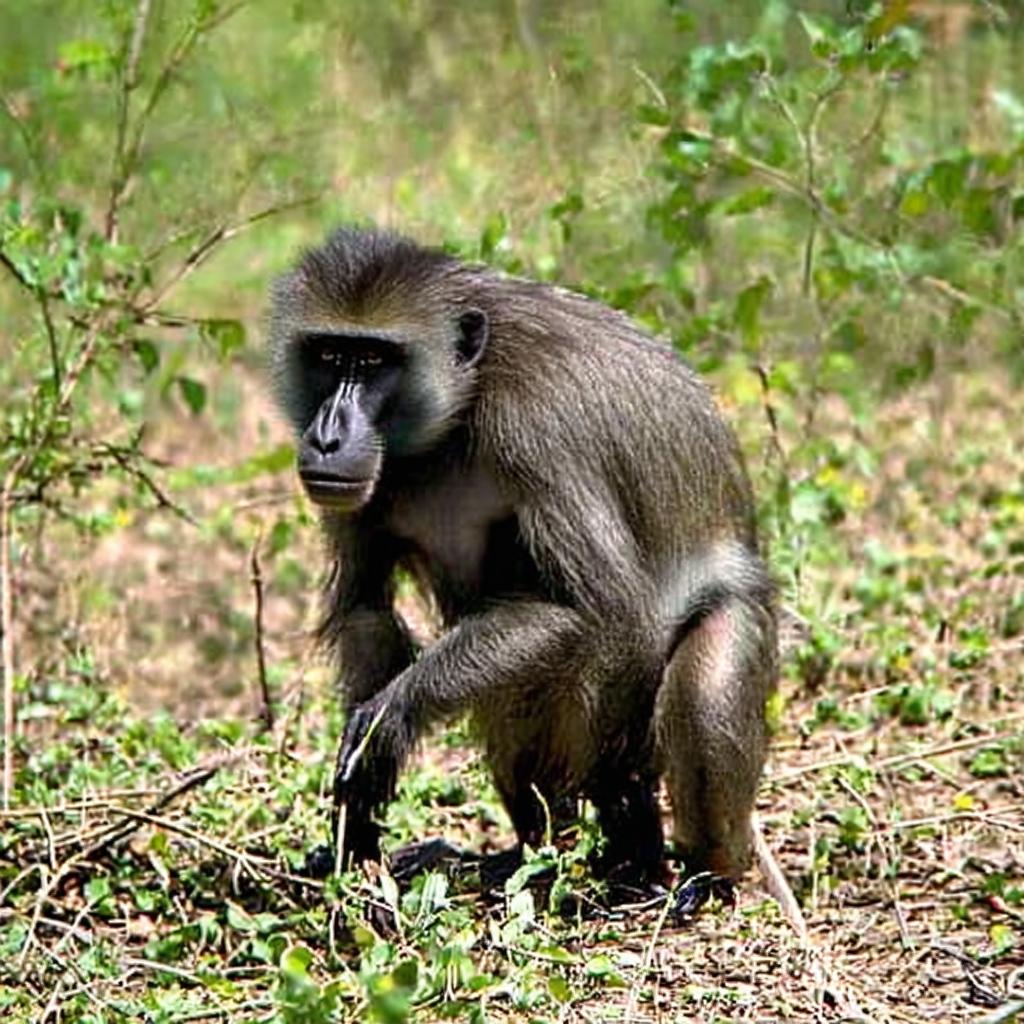}
  \includegraphics[width=0.16\linewidth]{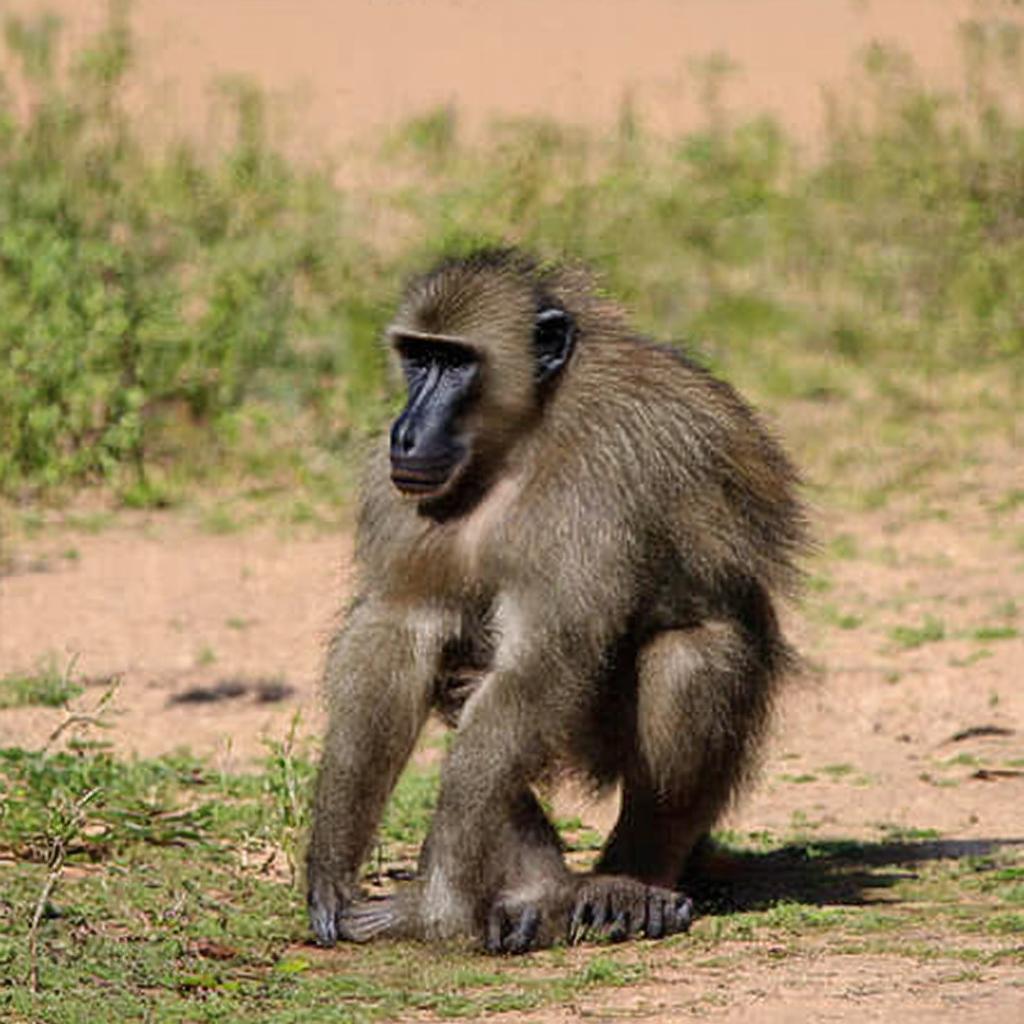}
  \includegraphics[width=0.16\linewidth]{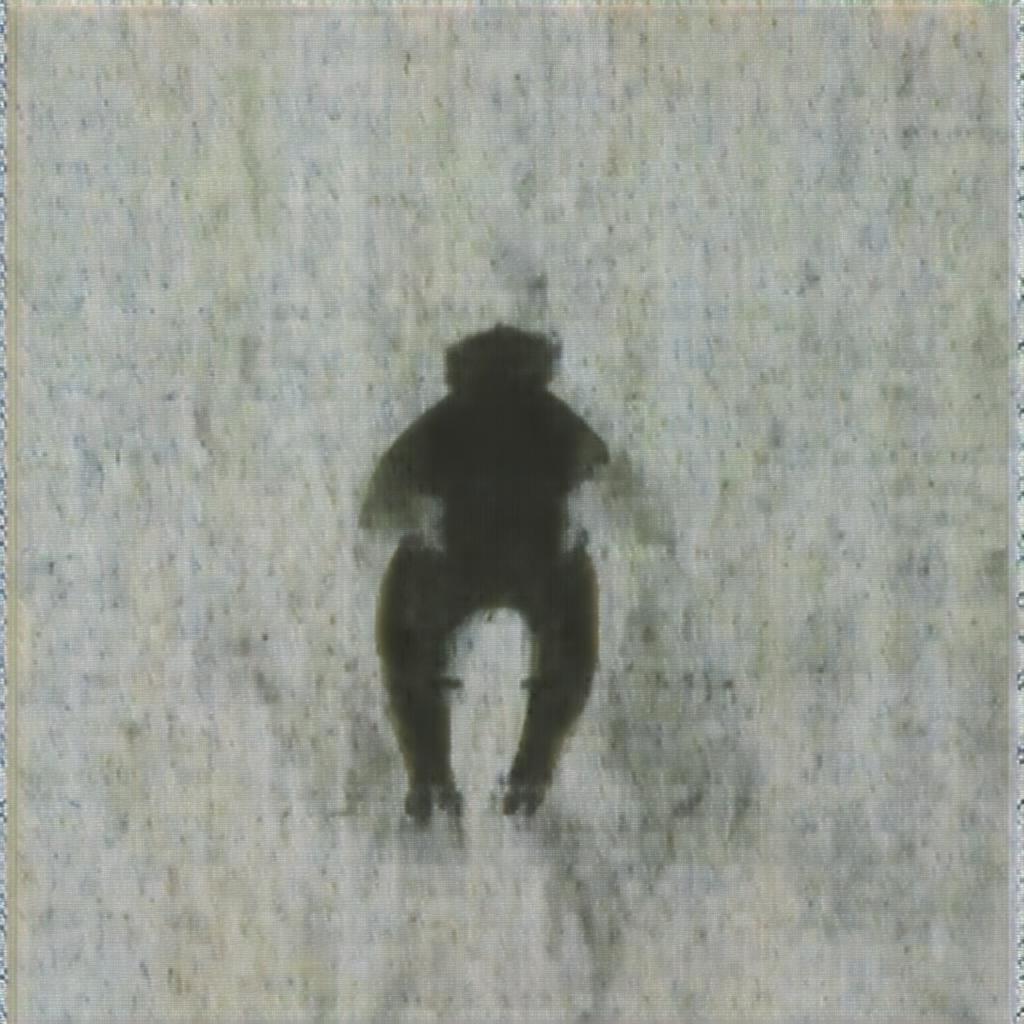}
  \includegraphics[width=0.16\linewidth]{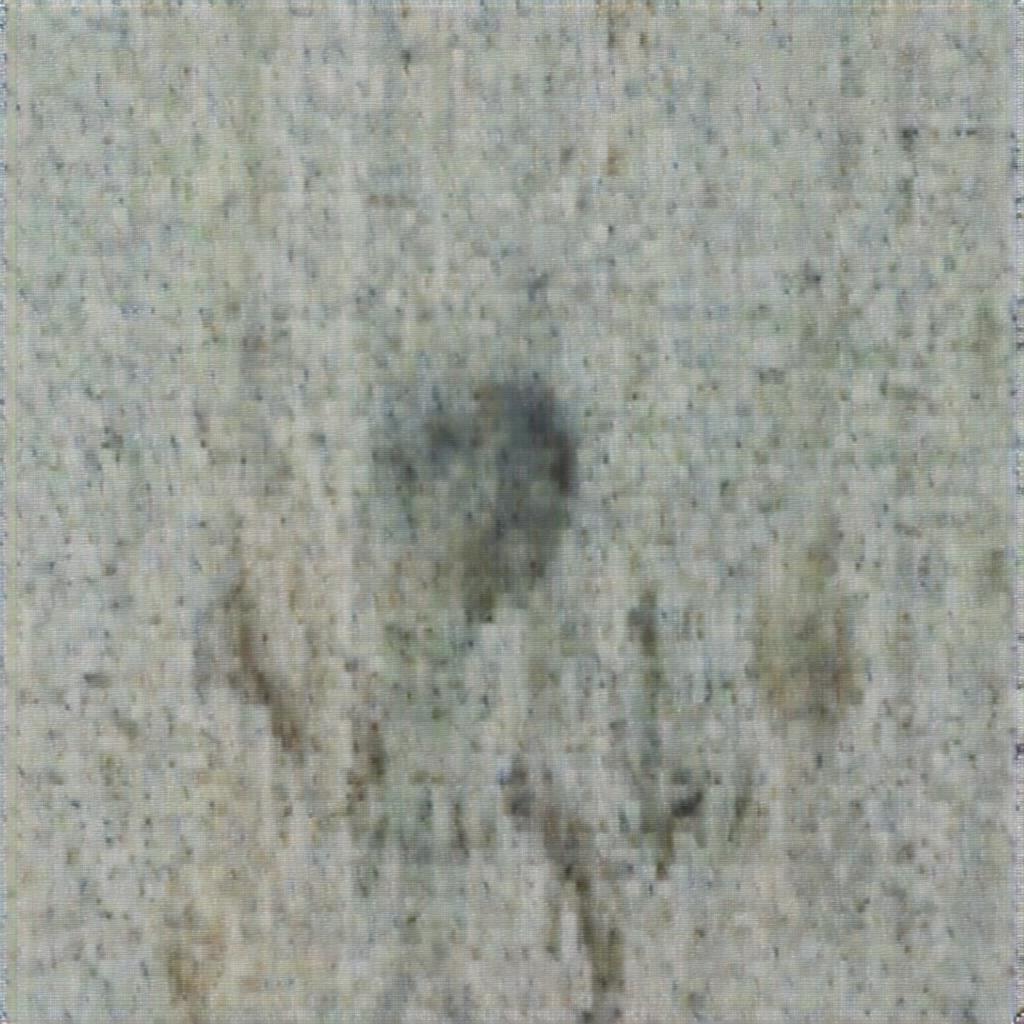} 
  \includegraphics[width=0.16\linewidth]{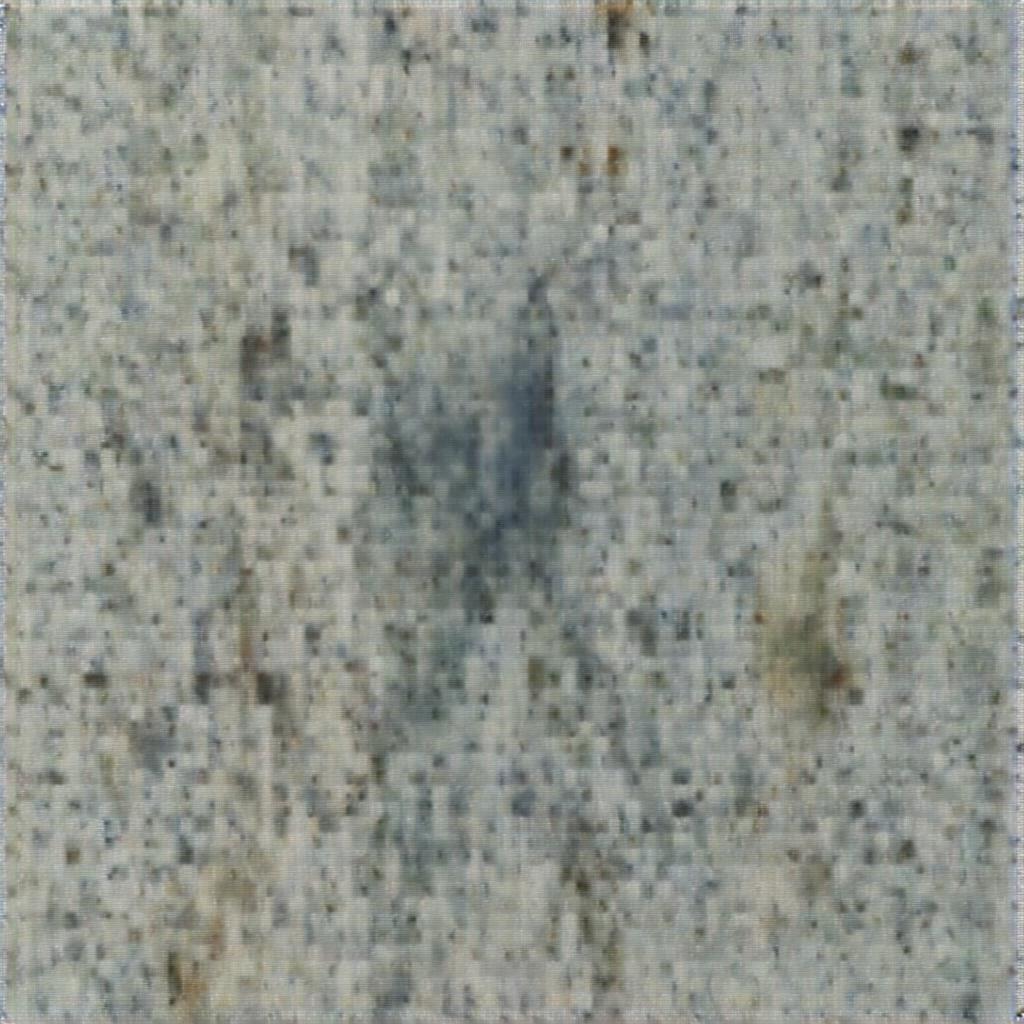}
  \includegraphics[width=0.16\linewidth]{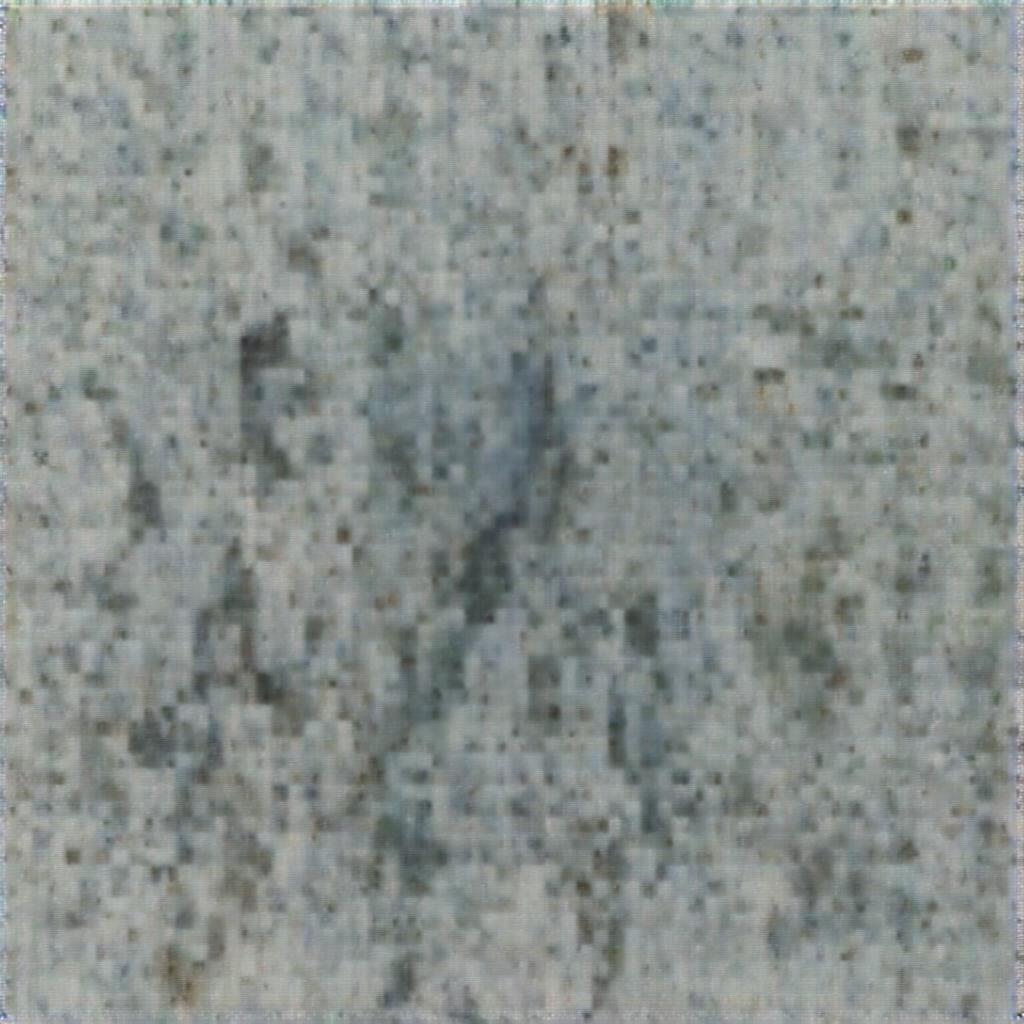} \\
  \vspace{0.2cm}
  \includegraphics[width=0.16\linewidth]{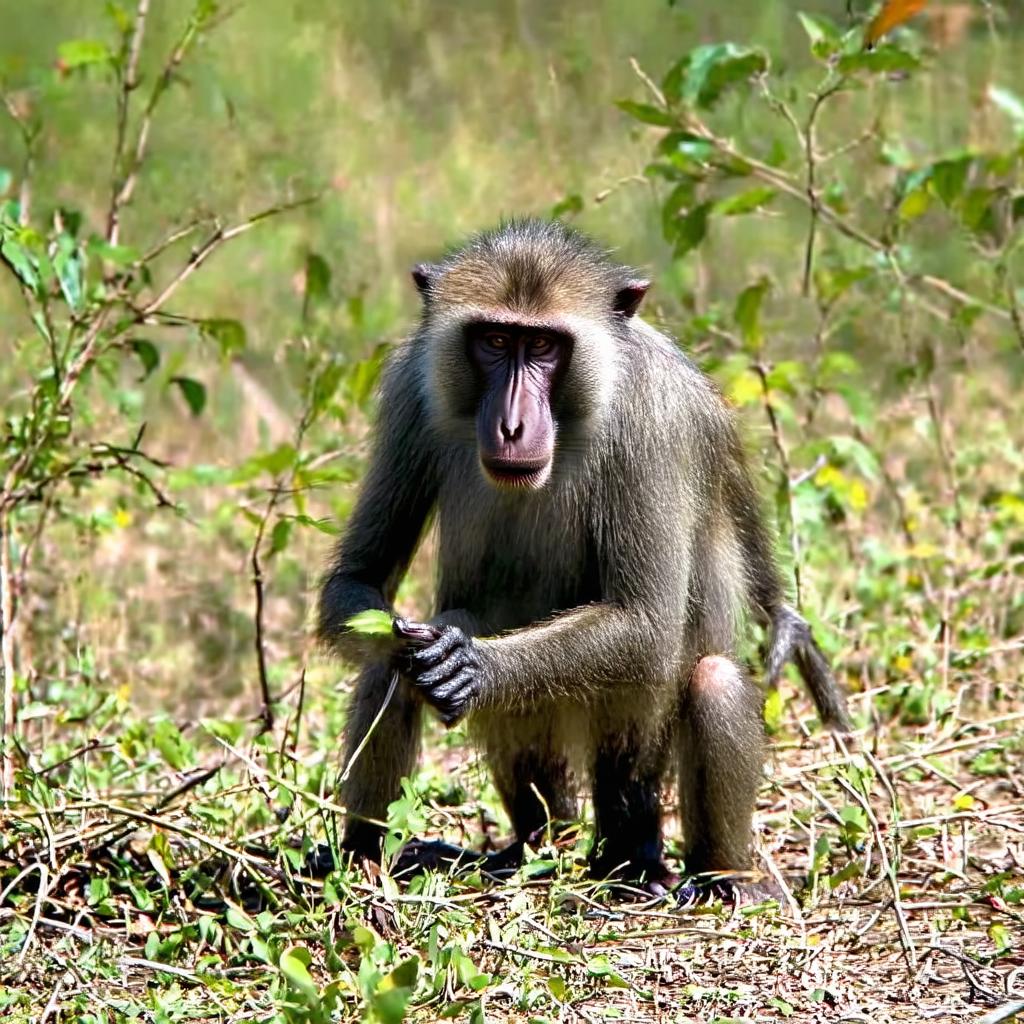}
  \includegraphics[width=0.16\linewidth]{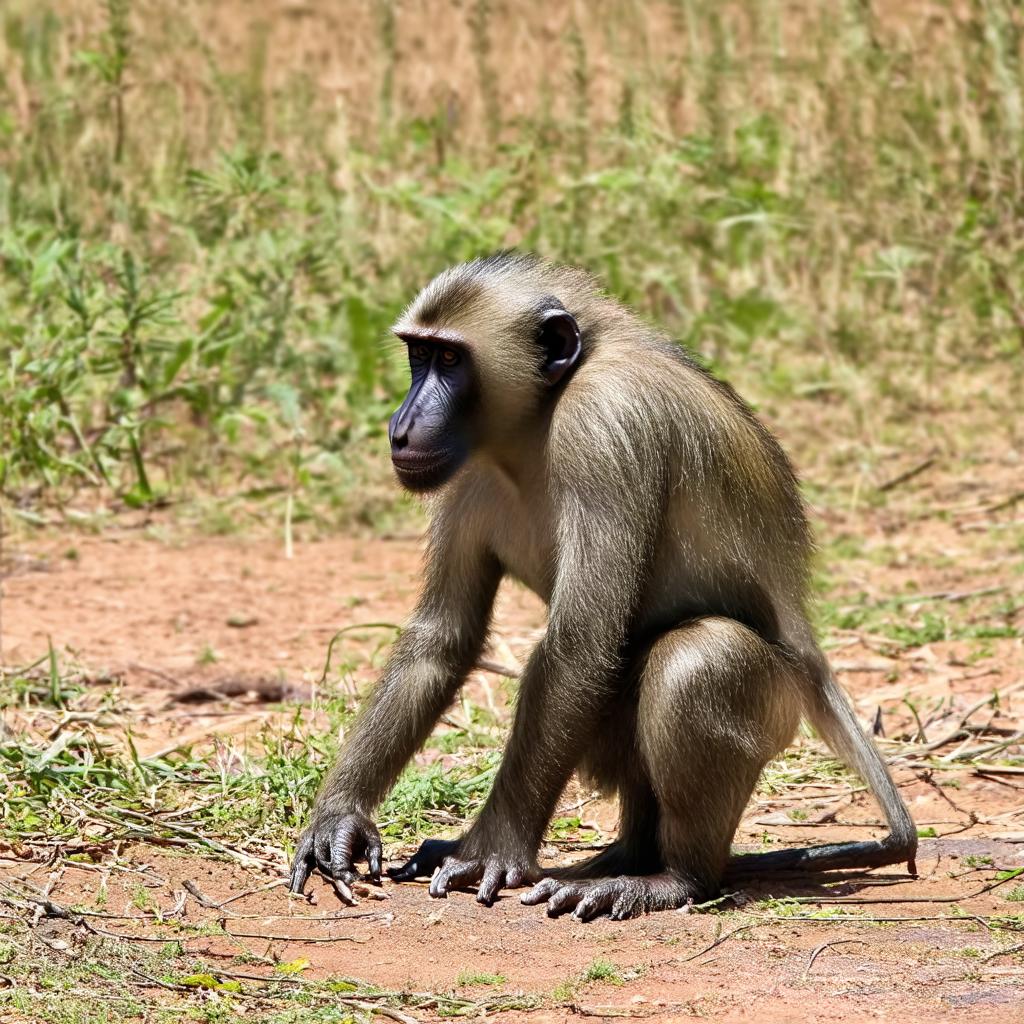}
  \includegraphics[width=0.16\linewidth]{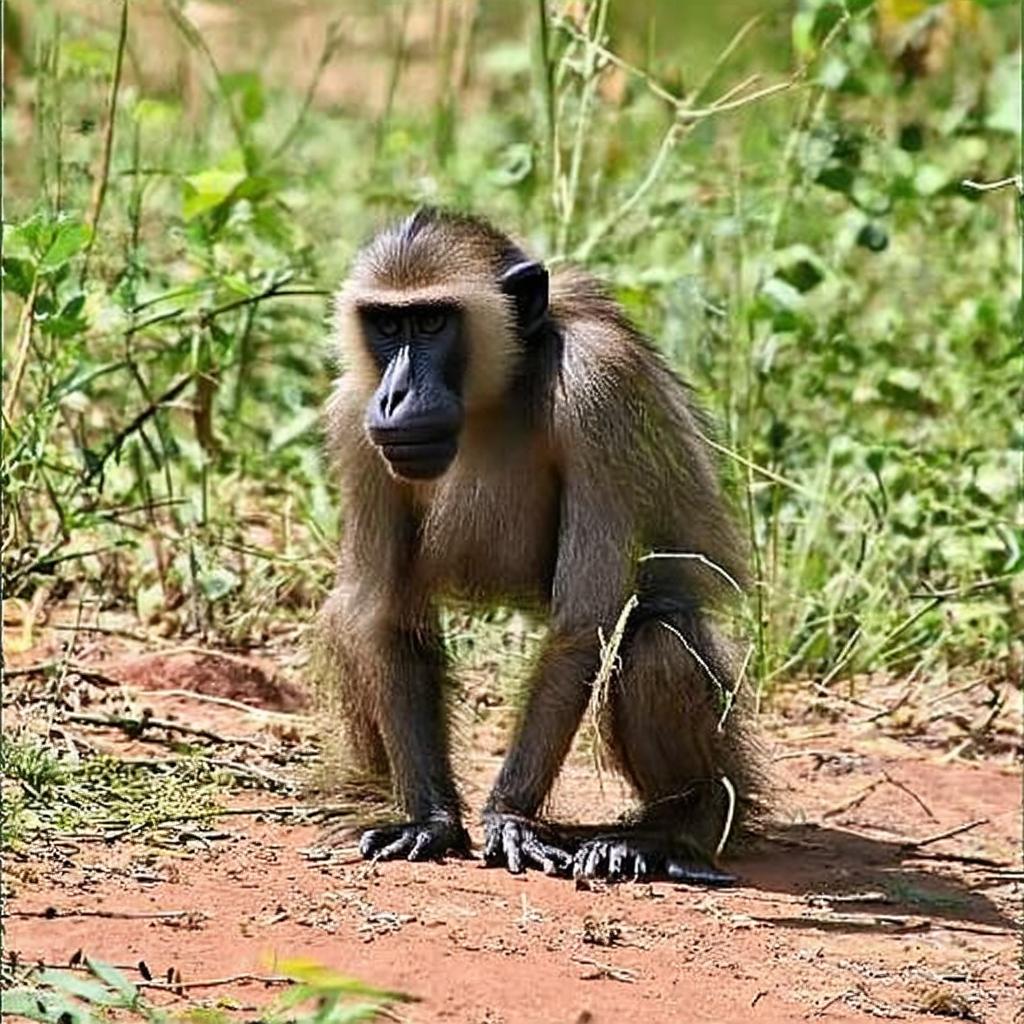}
  \includegraphics[width=0.16\linewidth]{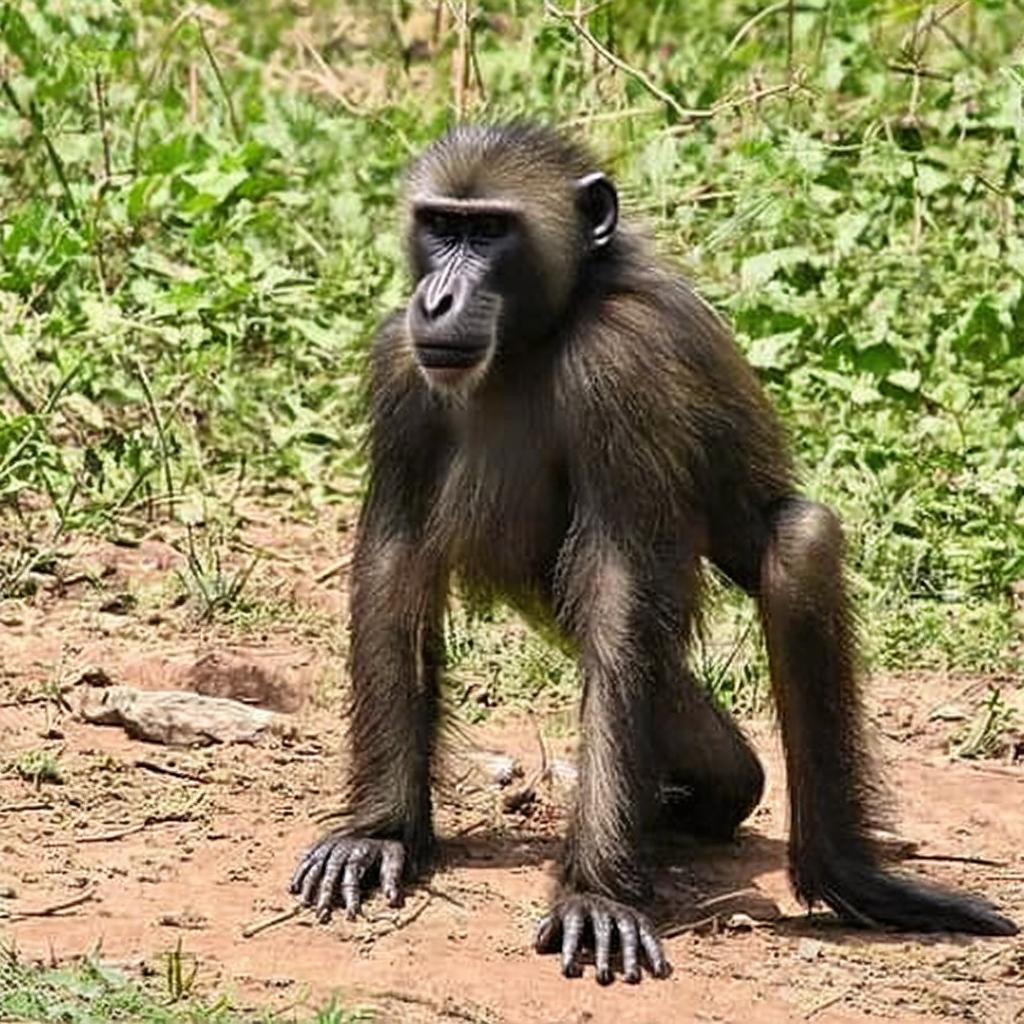}
  \includegraphics[width=0.16\linewidth]{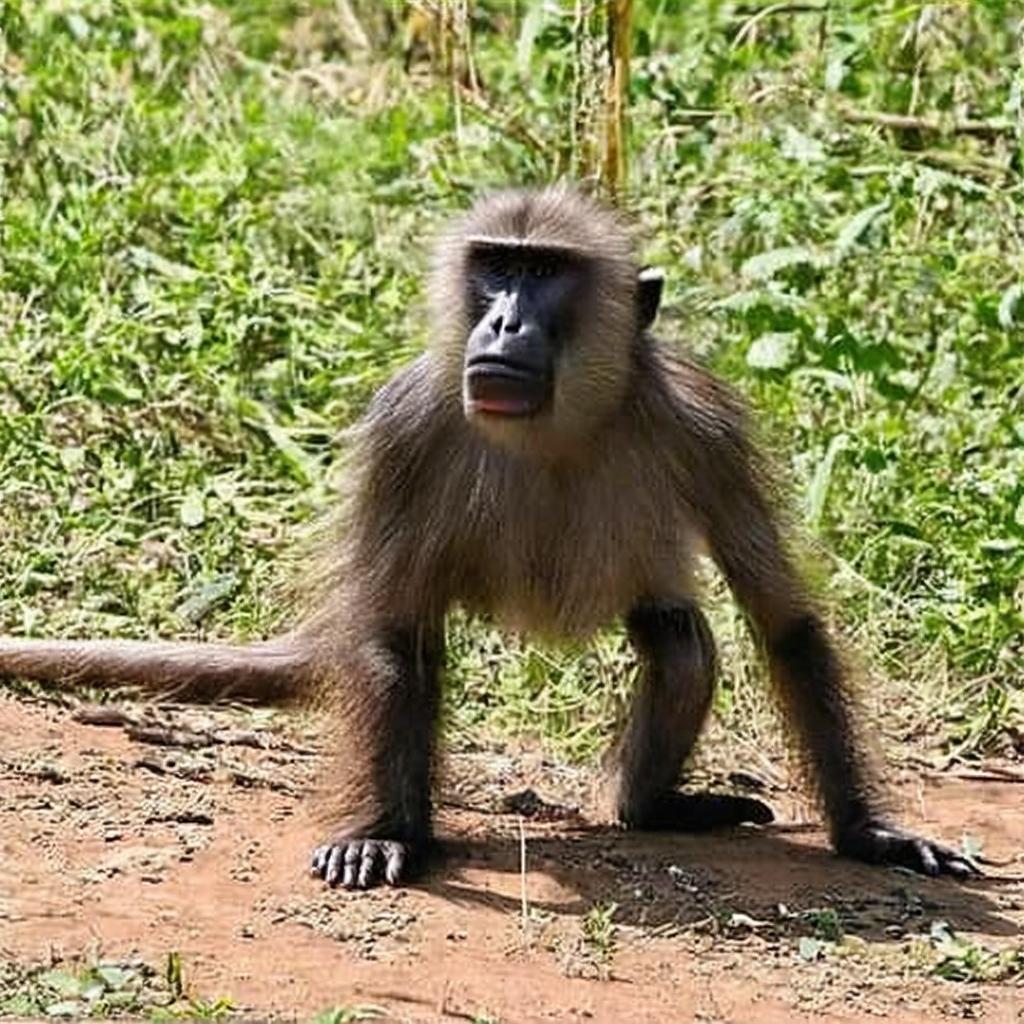}
  \includegraphics[width=0.16\linewidth]{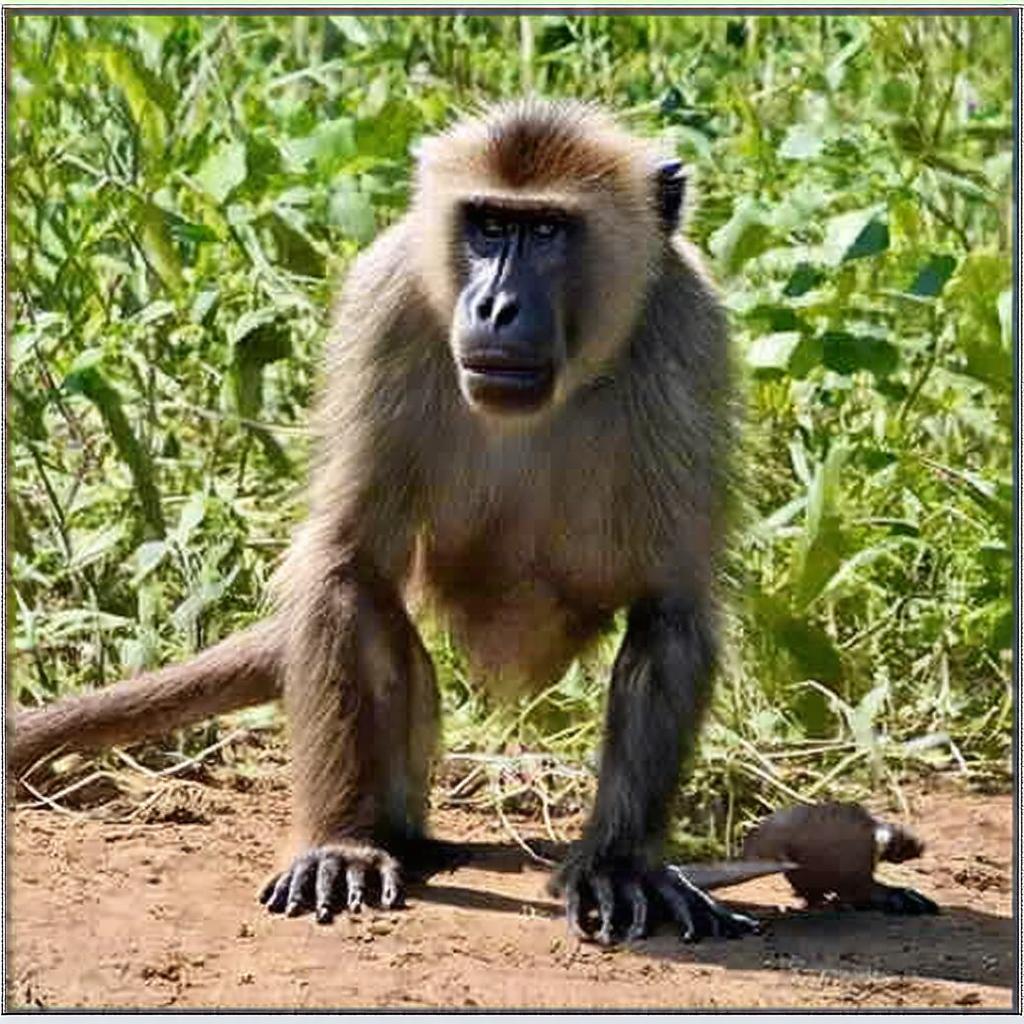} \\
  \vspace{0.2cm}
  \includegraphics[width=0.16\linewidth]{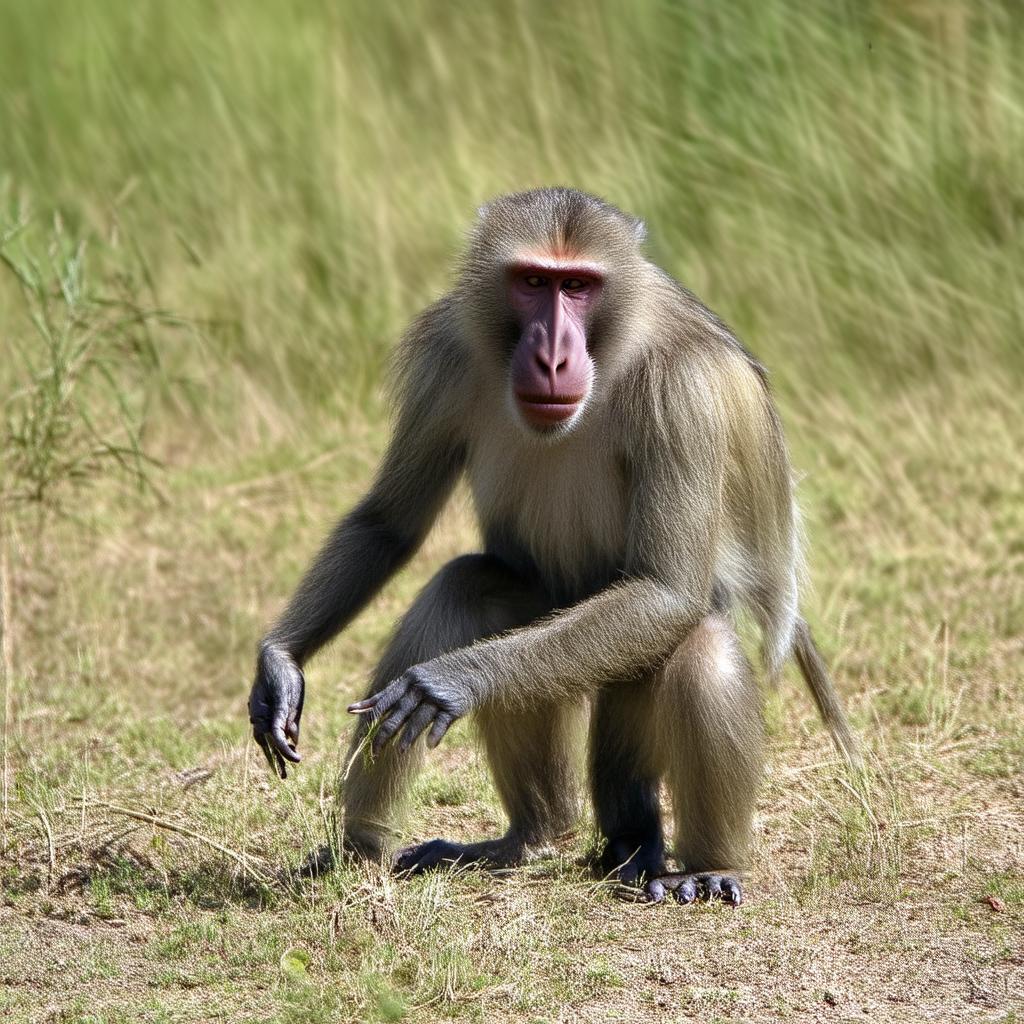}
  \includegraphics[width=0.16\linewidth]{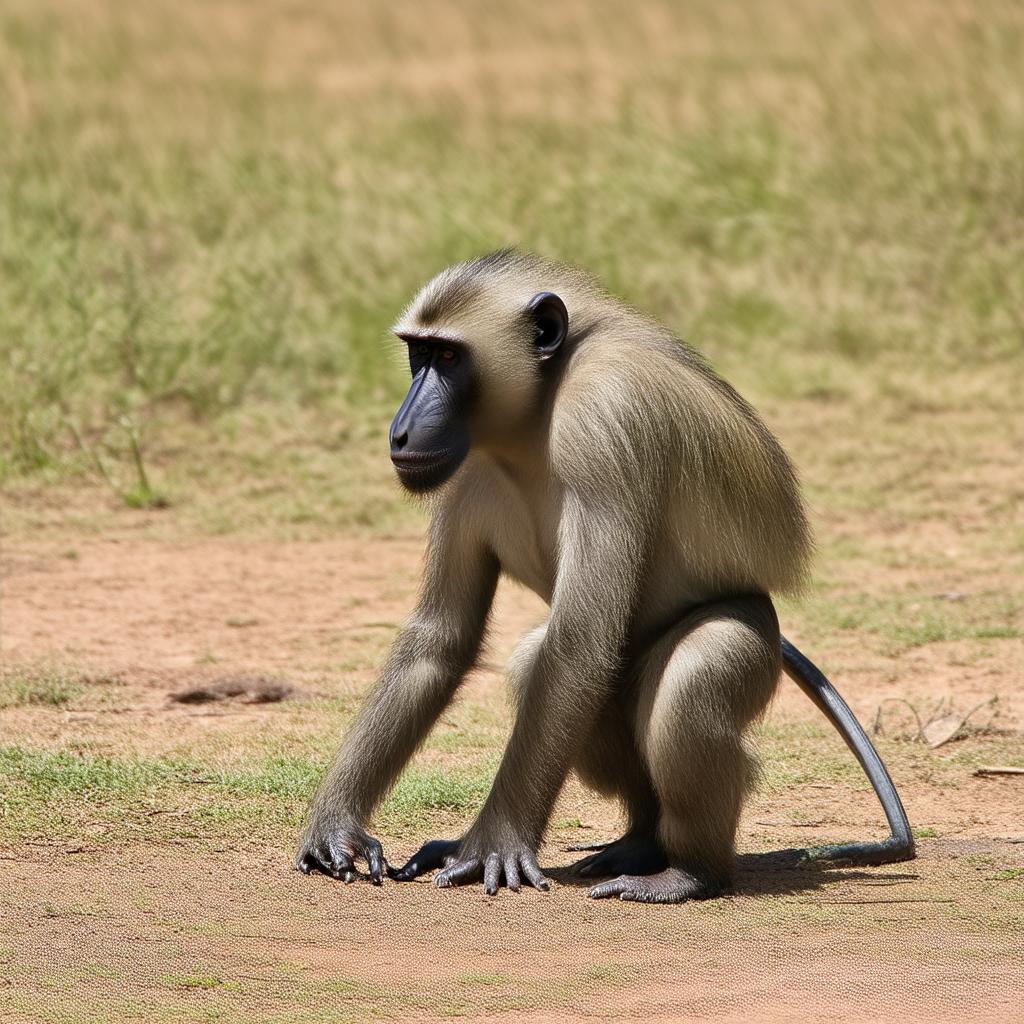}
  \includegraphics[width=0.16\linewidth]{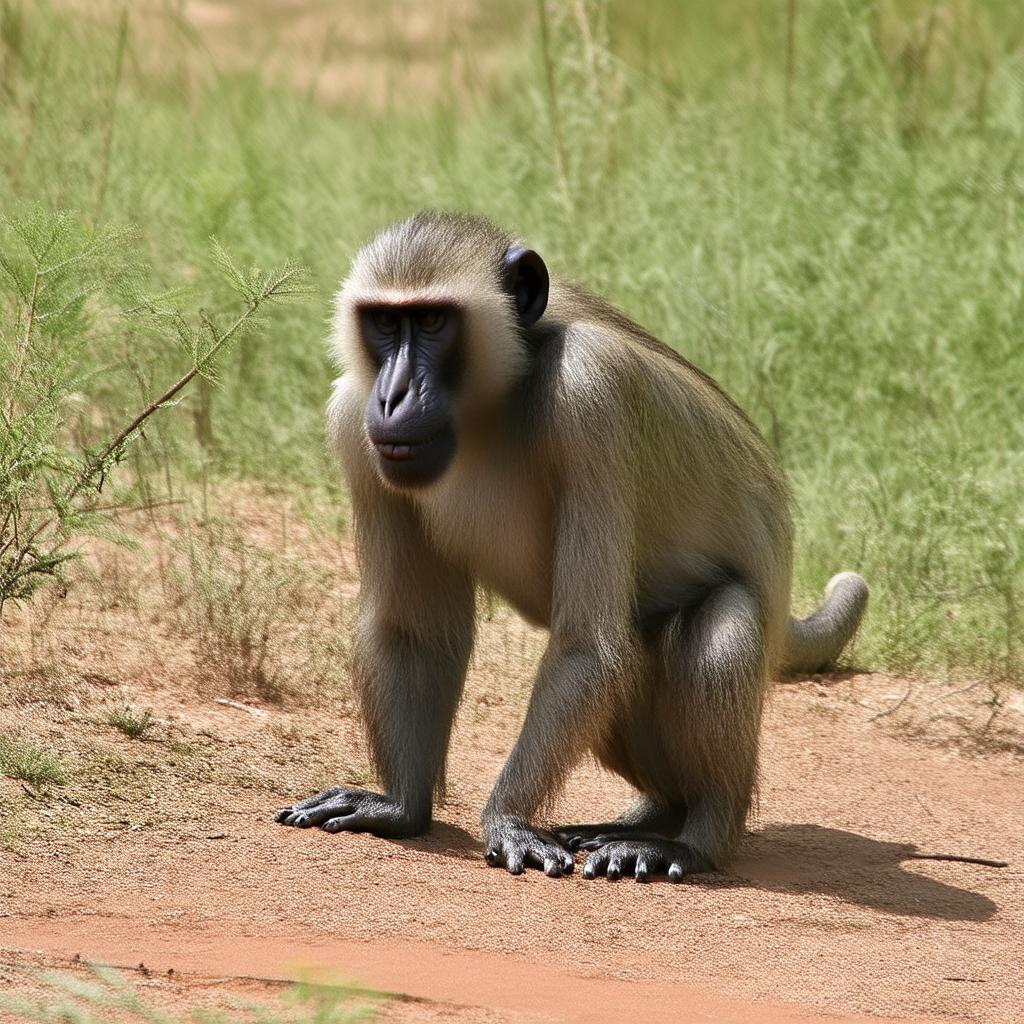}
  \includegraphics[width=0.16\linewidth]{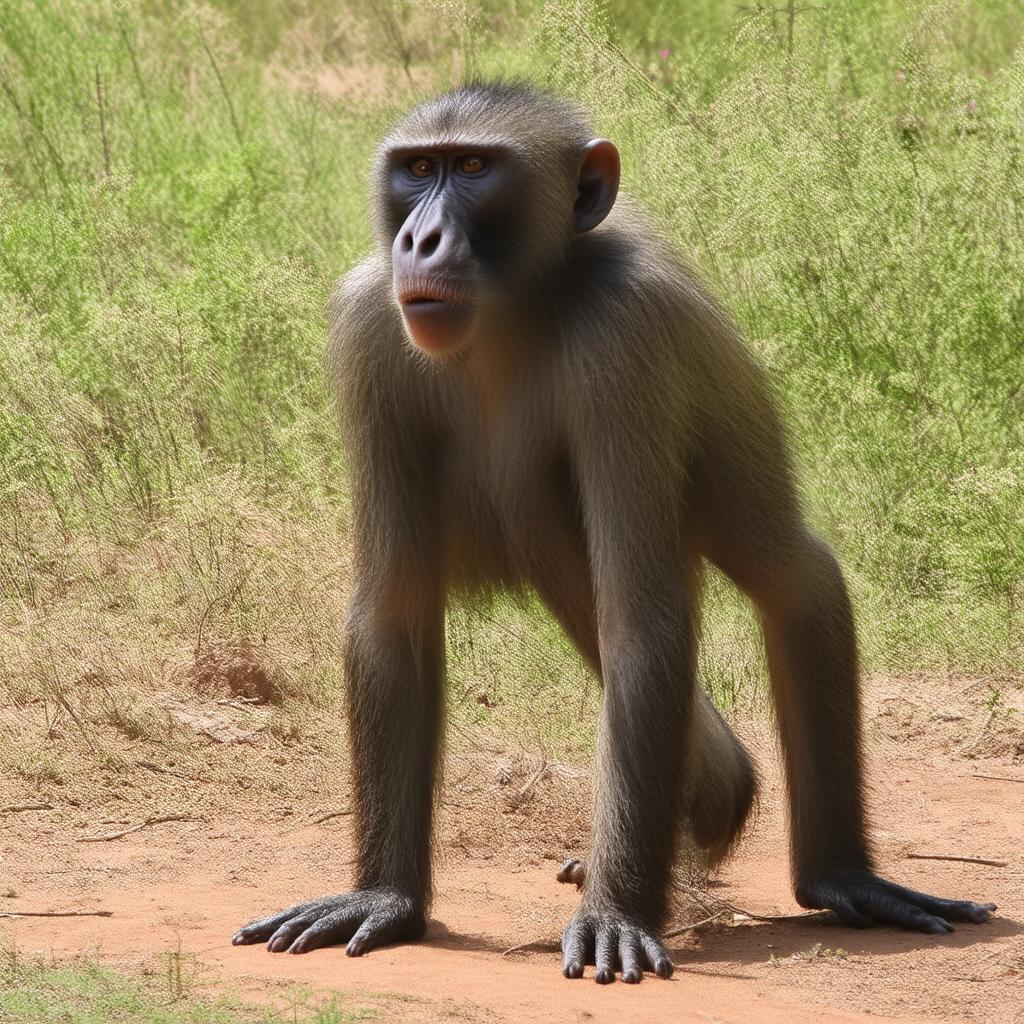} 
  \includegraphics[width=0.16\linewidth]{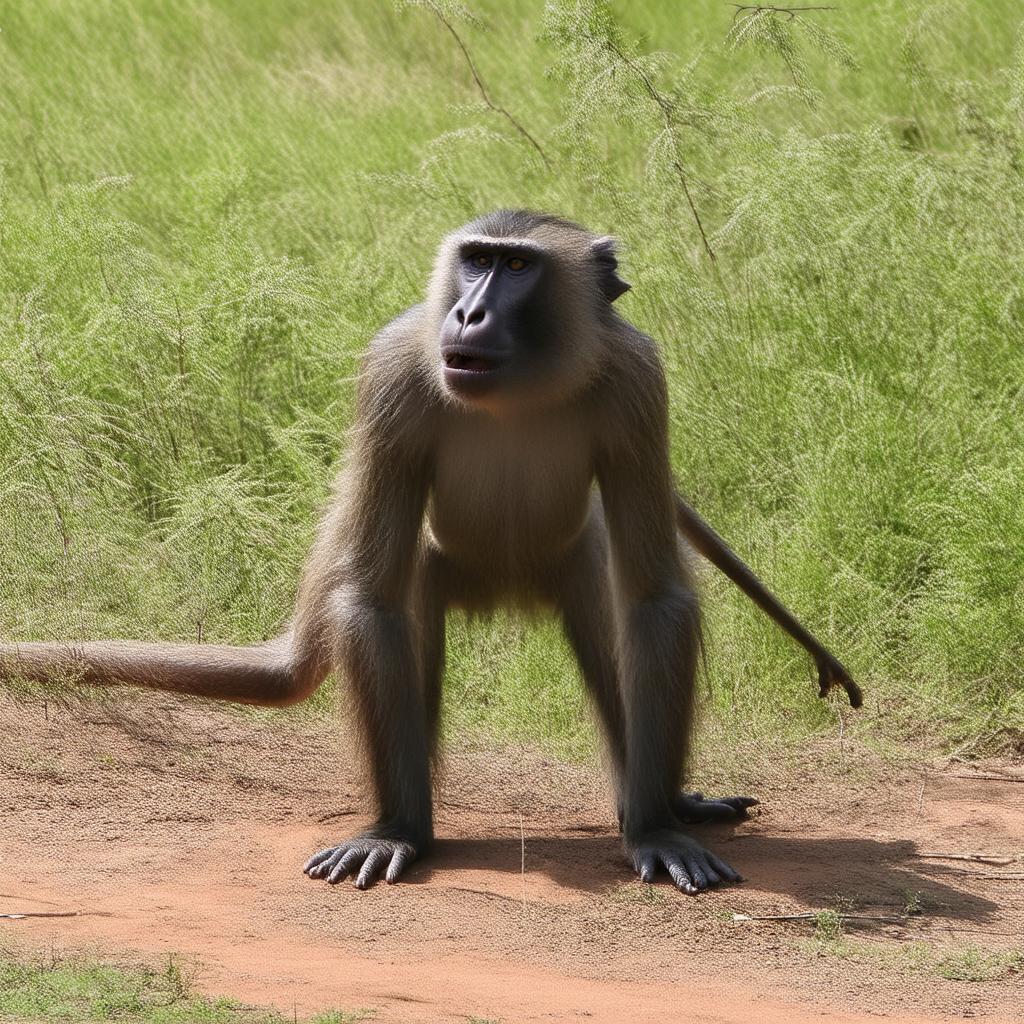}
  \includegraphics[width=0.16\linewidth]{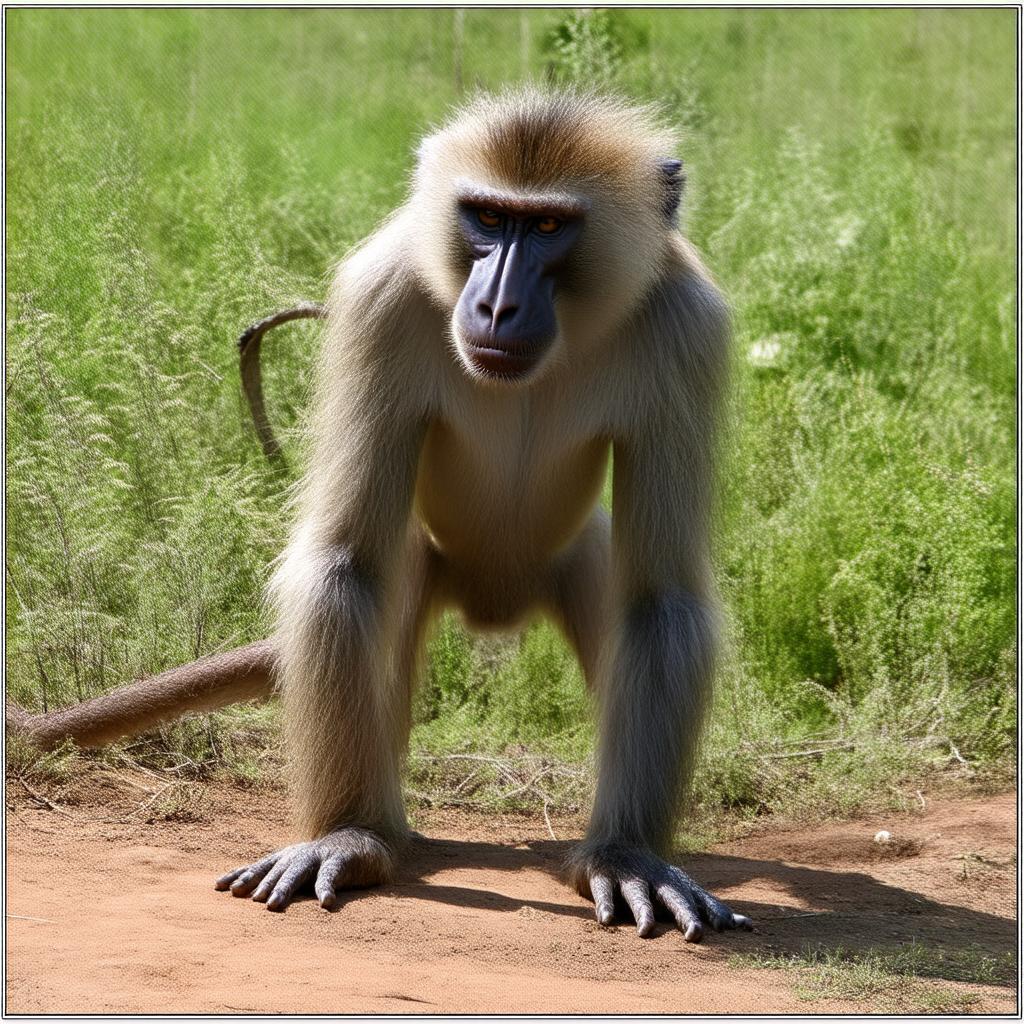} \\
 
  \caption{Images generated from centroids of latents obtained from $N$ input images of ImageNet class ``baboon'' using SD 3.5. $N = 2, 8, 32, 48, 64, 96$. Top row: Fixed normalization. Middle row: \texttt{fix/chm}. Bottom row: \texttt{nin/chm}.}

  \label{fig:additional_qualitative_ex}
\end{figure*}

Figure~\ref{fig:additional_qualitative_ex} shows an additional example with images generated from centroids computed using increasing $N$, this time using SD 3.5. Similarly to Figure~\ref{fig:qualitative_ex_monarch}, the degeneracy is greatly reduced using channel-wise mean adjustment, and the difference between the \texttt{nin} and \texttt{fix} is marginal in comparison.

\end{document}